\documentclass[11pt,fleqn]{book} 

\usepackage[top=3cm,bottom=3cm,left=3cm,right=3cm,headsep=10pt,a4paper]{geometry} 

\usepackage{graphicx} 
\graphicspath{{fig/}} 

\usepackage{lipsum} 

\usepackage{tikz} 

\usepackage[english]{babel} 

\usepackage{enumitem} 
\setlist{nolistsep} 

\usepackage{booktabs} 

\usepackage{xcolor} 
\definecolor{ocre}{RGB}{243,102,25} 

\usepackage{avant} 
\usepackage{mathptmx} 

\usepackage{microtype} 
\usepackage[utf8]{inputenc} 
\usepackage[T1]{fontenc} 

\usepackage[style=authoryear-comp,uniquelist=false,citestyle=authoryear,sorting=nyt,sortcites=true,autopunct=true,babel=hyphen,hyperref,abbreviate=false,backref=true,backend=biber,maxcitenames=2,maxbibnames=20]{biblatex}
\defbibheading{bibempty}{}

\usepackage{calc} 
\usepackage{makeidx} 
\makeindex 

\usepackage{titletoc} 

\contentsmargin{0cm} 

\titlecontents{part}[0cm]
{\addvspace{20pt}\centering\large\bfseries}
{}
{}
{}

\titlecontents{chapter}[1.25cm] 
{\addvspace{12pt}\large\sffamily\bfseries} 
{\color{ocre!60}\contentslabel[\Large\thecontentslabel]{1.25cm}\color{ocre}} 
{\color{ocre}}  
{\color{ocre!60}\normalsize\;\titlerule*[.5pc]{.}\;\thecontentspage} 

\titlecontents{section}[1.25cm] 
{\addvspace{3pt}\sffamily\bfseries} 
{\contentslabel[\thecontentslabel]{1.25cm}} 
{}
{\hfill\color{black}\thecontentspage} 
[]

\titlecontents{subsection}[1.25cm] 
{\addvspace{1pt}\sffamily\small} 
{\contentslabel[\thecontentslabel]{1.25cm}} 
{}
{\ \titlerule*[.5pc]{.}\;\thecontentspage} 
[]

\titlecontents{figure}[0em]
{\addvspace{-5pt}\sffamily}
{\thecontentslabel\hspace*{1em}}
{}
{\ \titlerule*[.5pc]{.}\;\thecontentspage}
[]

\titlecontents{table}[0em]
{\addvspace{-5pt}\sffamily}
{\thecontentslabel\hspace*{1em}}
{}
{\ \titlerule*[.5pc]{.}\;\thecontentspage}
[]

\titlecontents{lchapter}[0em] 
{\addvspace{15pt}\large\sffamily\bfseries} 
{\color{ocre}\contentslabel[\Large\thecontentslabel]{1.25cm}\color{ocre}} 
{}  
{\color{ocre}\normalsize\sffamily\bfseries\;\titlerule*[.5pc]{.}\;\thecontentspage} 

\titlecontents{lsection}[0em] 
{\sffamily\small} 
{\contentslabel[\thecontentslabel]{1.25cm}} 
{}
{}

\titlecontents{lsubsection}[.5em] 
{\normalfont\footnotesize\sffamily} 
{}
{}
{}

\usepackage{fancyhdr} 

\fancypagestyle{plain}{\fancyhead{}} 

\makeatletter
\renewcommand{\cleardoublepage}{
\clearpage\ifodd\c@page\else
\hbox{}
\vspace*{\fill}
\thispagestyle{empty}
\newpage
\fi}

\usepackage{amsmath,amsfonts,amssymb,amsthm} 

\newtheoremstyle{ocrenumbox}
{0pt}
{0pt}
{\normalfont}
{}
{\small\bf\sffamily\color{ocre}}
{\;}
{0.25em}
{\small\sffamily\color{ocre}\thmname{#1}\nobreakspace\thmnumber{\@ifnotempty{#1}{}\@upn{#2}}
\thmnote{\nobreakspace\the\thm@notefont\sffamily\bfseries\color{black}---\nobreakspace#3.}} 

\newtheoremstyle{blacknumex}
{5pt}
{5pt}
{\normalfont}
{} 
{\small\bf\sffamily}
{\;}
{0.25em}
{\small\sffamily{\tiny\ensuremath{\blacksquare}}\nobreakspace\thmname{#1}\nobreakspace\thmnumber{\@ifnotempty{#1}{}\@upn{#2}}
\thmnote{\nobreakspace\the\thm@notefont\sffamily\bfseries---\nobreakspace#3.}}

\newtheoremstyle{blacknumbox} 
{0pt}
{0pt}
{\normalfont}
{}
{\small\bf\sffamily}
{\;}
{0.25em}
{\small\sffamily\thmname{#1}\nobreakspace\thmnumber{\@ifnotempty{#1}{}\@upn{#2}}
\thmnote{\nobreakspace\the\thm@notefont\sffamily\bfseries---\nobreakspace#3.}}

\newtheoremstyle{ocrenum}
{5pt}
{5pt}
{\normalfont}
{}
{\small\bf\sffamily\color{ocre}}
{\;}
{0.25em}
{\small\sffamily\color{ocre}\thmname{#1}\nobreakspace\thmnumber{\@ifnotempty{#1}{}\@upn{#2}}
\thmnote{\nobreakspace\the\thm@notefont\sffamily\bfseries\color{black}---\nobreakspace#3.}} 
\makeatother

\newcounter{dummy} 
\numberwithin{dummy}{section}
\theoremstyle{ocrenumbox}
\newtheorem{theoremeT}[dummy]{Theorem}

\newtheorem{exerciseT}{Exercise}[chapter]
\theoremstyle{blacknumex}
\newtheorem{exampleT}{Example}[chapter]
\theoremstyle{blacknumbox}

\newtheorem{definitionT}{Definition}[section]
\newtheorem{corollaryT}[dummy]{Corollary}
\theoremstyle{ocrenum}

\RequirePackage[framemethod=default]{mdframed} 

\newmdenv[skipabove=7pt,
skipbelow=7pt,
backgroundcolor=black!5,
linecolor=ocre,
innerleftmargin=5pt,
innerrightmargin=5pt,
innertopmargin=5pt,
leftmargin=0cm,
rightmargin=0cm,
innerbottommargin=5pt]{tBox}

\newmdenv[skipabove=7pt,
skipbelow=7pt,
rightline=false,
leftline=true,
topline=false,
bottomline=false,
backgroundcolor=ocre!10,
linecolor=ocre,
innerleftmargin=5pt,
innerrightmargin=5pt,
innertopmargin=5pt,
innerbottommargin=5pt,
leftmargin=0cm,
rightmargin=0cm,
linewidth=4pt]{eBox}	

\newmdenv[skipabove=7pt,
skipbelow=7pt,
rightline=false,
leftline=true,
topline=false,
bottomline=false,
linecolor=ocre,
innerleftmargin=5pt,
innerrightmargin=5pt,
innertopmargin=0pt,
leftmargin=0cm,
rightmargin=0cm,
linewidth=4pt,
innerbottommargin=0pt]{dBox}	

\newmdenv[skipabove=7pt,
skipbelow=7pt,
rightline=false,
leftline=true,
topline=false,
bottomline=false,
linecolor=gray,
backgroundcolor=black!5,
innerleftmargin=5pt,
innerrightmargin=5pt,
innertopmargin=5pt,
leftmargin=0cm,
rightmargin=0cm,
linewidth=4pt,
innerbottommargin=5pt]{cBox}

\newenvironment{theorem}{\begin{tBox}\begin{theoremeT}}{\end{theoremeT}\end{tBox}}
				  
\newenvironment{definition}{\begin{dBox}\begin{definitionT}}{\end{definitionT}\end{dBox}}	
\newenvironment{example}{\begin{exampleT}}{\hfill{\tiny\ensuremath{\blacksquare}}\end{exampleT}}		
\newenvironment{corollary}{\begin{cBox}\begin{corollaryT}}{\end{corollaryT}\end{cBox}}	

\newenvironment{remark}{\par\vspace{10pt}\small 
\begin{list}{}{
\leftmargin=35pt 
\rightmargin=25pt}\item\ignorespaces 
\makebox[-2.5pt]{\begin{tikzpicture}[overlay]
\node[draw=ocre!60,line width=1pt,circle,fill=ocre!25,font=\sffamily\bfseries,inner sep=2pt,outer sep=0pt] at (-15pt,0pt){\textcolor{ocre}{R}};\end{tikzpicture}} 
\advance\baselineskip -1pt}{\end{list}\vskip5pt} 

\makeatletter
\renewcommand{\@seccntformat}[1]{\llap{\textcolor{ocre}{\csname the#1\endcsname}\hspace{1em}}}                    
\renewcommand{\section}{\@startsection{section}{1}{\z@}
{-4ex \@plus -1ex \@minus -.4ex}
{1ex \@plus.2ex }
{\normalfont\large\sffamily\bfseries}}
\renewcommand{\subsection}{\@startsection {subsection}{2}{\z@}
{-3ex \@plus -0.1ex \@minus -.4ex}
{0.5ex \@plus.2ex }
{\normalfont\sffamily\bfseries}}
\renewcommand{\subsubsection}{\@startsection {subsubsection}{3}{\z@}
{-2ex \@plus -0.1ex \@minus -.2ex}
{.2ex \@plus.2ex }
{\normalfont\small\sffamily\bfseries}}                        
\renewcommand\paragraph{\@startsection{paragraph}{4}{\z@}
{-2ex \@plus-.2ex \@minus .2ex}
{.1ex}
{\normalfont\small\sffamily\bfseries}}

\newcommand{\@mypartnumtocformat}[2]{%
\setlength\fboxsep{0pt}%
\noindent\colorbox{ocre!20}{\strut\parbox[c][.7cm]{\ecart}{\color{ocre!70}\Large\sffamily\bfseries\centering#1}}\hskip\esp\colorbox{ocre!40}{\strut\parbox[c][.7cm]{\linewidth-\ecart-\esp}{\Large\sffamily\centering#2}}}%
\newcommand{\@myparttocformat}[1]{%
\setlength\fboxsep{0pt}%
\noindent\colorbox{ocre!40}{\strut\parbox[c][.7cm]{\linewidth}{\Large\sffamily\centering#1}}}%
\newlength\esp
\newlength\ecart
\def\@part[#1]#2{%
\ifnum \c@secnumdepth >-2\relax%
\refstepcounter{part}%
\addcontentsline{toc}{part}{\texorpdfstring{\protect\@mypartnumtocformat{\thepart}{#1}}{\partname~\thepart\ ---\ #1}}
\else%
\addcontentsline{toc}{part}{\texorpdfstring{\protect\@myparttocformat{#1}}{#1}}%
\fi%
\startcontents%
\markboth{}{}%
{\thispagestyle{empty}%
\begin{tikzpicture}[remember picture,overlay]%
\node at (current page.north west){\begin{tikzpicture}[remember picture,overlay]%
\fill[ocre!20](0cm,0cm) rectangle (\paperwidth,-\paperheight);
\node[anchor=north] at (4cm,-3.25cm){\color{ocre!40}\fontsize{220}{100}\sffamily\bfseries\@Roman\c@part}; 
\node[anchor=south east] at (\paperwidth-1cm,-\paperheight+1cm){\parbox[t][][t]{8.5cm}{
\printcontents{l}{0}{\setcounter{tocdepth}{1}}%
}};
\node[anchor=north east] at (\paperwidth-1.5cm,-3.25cm){\parbox[t][][t]{15cm}{\strut\raggedleft\color{white}\fontsize{30}{30}\sffamily\bfseries#2}};
\end{tikzpicture}};
\end{tikzpicture}}%
\@endpart}
\def\@spart#1{%
\startcontents%
\phantomsection
{\thispagestyle{empty}%
\begin{tikzpicture}[remember picture,overlay]%
\node at (current page.north west){\begin{tikzpicture}[remember picture,overlay]%
\fill[ocre!20](0cm,0cm) rectangle (\paperwidth,-\paperheight);
\node[anchor=north east] at (\paperwidth-1.5cm,-3.25cm){\parbox[t][][t]{15cm}{\strut\raggedleft\color{white}\fontsize{30}{30}\sffamily\bfseries#1}};
\end{tikzpicture}};
\end{tikzpicture}}
\addcontentsline{toc}{part}{\texorpdfstring{%
\setlength\fboxsep{0pt}%
\noindent\protect\colorbox{ocre!40}{\strut\protect\parbox[c][.7cm]{\linewidth}{\Large\sffamily\protect\centering #1\quad\mbox{}}}}{#1}}%
\@endpart}
\def\@endpart{\vfil\newpage
\if@twoside
\if@openright
\null
\thispagestyle{empty}%
\newpage
\fi
\fi
\if@tempswa
\twocolumn
\fi}

\newcommand{\thechapterimage}{}%
\newcommand{\chapterimage}[1]{\renewcommand{\thechapterimage}{#1}}%
\def\@makechapterhead#1{%
{\parindent \z@ \raggedright \normalfont
\ifnum \c@secnumdepth >\m@ne
\if@mainmatter
\begin{tikzpicture}[remember picture,overlay]
\node at (current page.north west)
{\begin{tikzpicture}[remember picture,overlay]
\node[anchor=north west,inner sep=0pt] at (0,0) {\IfFileExists{fig/\thechapterimage}{\includegraphics[width=\paperwidth]{\thechapterimage}}{}};
\draw[anchor=west] (\Gm@lmargin,-9cm) node [line width=2pt,rounded corners=15pt,draw=ocre,fill=white,fill opacity=0.5,inner sep=15pt]{\strut\makebox[22cm]{}};
\draw[anchor=west] (\Gm@lmargin+.3cm,-9cm) node {\huge\sffamily\bfseries\color{black}\thechapter. #1\strut};
\end{tikzpicture}};
\end{tikzpicture}
\else
\begin{tikzpicture}[remember picture,overlay]
\node at (current page.north west)
{\begin{tikzpicture}[remember picture,overlay]
\node[anchor=north west,inner sep=0pt] at (0,0) {\IfFileExists{fig/\thechapterimage}{\includegraphics[width=\paperwidth]{\thechapterimage}}{}};
\draw[anchor=west] (\Gm@lmargin,-9cm) node [line width=2pt,rounded corners=15pt,draw=ocre,fill=white,fill opacity=0.5,inner sep=15pt]{\strut\makebox[22cm]{}};
\draw[anchor=west] (\Gm@lmargin+.3cm,-9cm) node {\huge\sffamily\bfseries\color{black}#1\strut};
\end{tikzpicture}};
\end{tikzpicture}
\fi\fi\par\vspace*{270\p@}}}

\def\@makeschapterhead#1{%
\begin{tikzpicture}[remember picture,overlay]
\node at (current page.north west)
{\begin{tikzpicture}[remember picture,overlay]
\node[anchor=north west,inner sep=0pt] at (0,0) {\IfFileExists{fig/\thechapterimage}{\includegraphics[width=\paperwidth]{\thechapterimage}}{}};
\draw[anchor=west] (\Gm@lmargin,-9cm) node [line width=2pt,rounded corners=15pt,draw=ocre,fill=white,fill opacity=0.5,inner sep=15pt]{\strut\makebox[22cm]{}};
\draw[anchor=west] (\Gm@lmargin+.3cm,-9cm) node {\huge\sffamily\bfseries\color{black}#1\strut};
\end{tikzpicture}};
\end{tikzpicture}
\par\vspace*{270\p@}}
\makeatother

\usepackage[colorlinks=true,
            linkcolor=ocre,
            urlcolor=blue,
            citecolor=gray]{hyperref}
\hypersetup{hidelinks,backref=true,pagebackref=true,hyperindex=true,colorlinks=true,breaklinks=true,urlcolor= ocre,bookmarks=true,bookmarksopen=false,pdftitle={Bandits on Graphs and Structures},pdfauthor={Michal Valko}}
\usepackage{bookmark}
\bookmarksetup{
open,
numbered,
addtohook={%
\ifnum\bookmarkget{level}=0 
\bookmarksetup{bold}%
\fi
\ifnum\bookmarkget{level}=-1 
\bookmarksetup{color=ocre,bold}%
\fi
}
} 
\usepackage{amsthm}
\usepackage{amsmath}
\usepackage{amssymb}
\usepackage{bm}
\usepackage{dsfont}

\DeclareMathOperator*{\argmax}{arg\,max}
\DeclareMathOperator*{\argmin}{arg\,min}
\DeclareMathOperator*{\arginf}{arg\,inf}

\DeclareMathOperator{\polylog}{polylog}

\DeclareMathOperator{\polyloglog}{polyloglog}

\newcommand{\set}[1]{\left\{#1\right\}}

\newcommand{\II}[1]{\mathds{1}_{\left\{#1\right\}}}
\newcommand{\I}{{\mathds{1}}}

\newcommand{\ra}{\rightarrow}

\newcommand{\Bernoulli}{\mathrm{Bernoulli}}

\newcommand{\specialcell}[2][c]{%
 \begin{tabular}[#1]{@{}c@{}}#2\end{tabular}}

\newtheorem{assumption}{Assumption}
\newcommand{\R}{\mathbb{R}}
\newcommand{\realset}{\mathbb{R}}

\newcommand{\NN}{{\mathbb N}}
\newcommand{\1}{\mathds{1}}

\newcommand{\E}{\mathbb{E}}
\newcommand{\EE}[1]{\mathbb{E}\left[#1\right]}
\newcommand{\EEt}[1]{\mathbb{E}_t\left[#1\right]}
\newcommand{\EEs}[2]{\mathbb{E}_{#1}\left[#2\right]}
\newcommand{\EEc}[2]{\mathbb{E}\left[#1\left|#2\right.\right]}
\newcommand{\EEcc}[2]{\mathbb{E}\left[\left.#1\right|#2\right]}

\newcommand{\PP}[1]{\mathbb{P}\left[#1\right]}

 \newcommand{\PPc}[2]{\mathbb{P}\left[#1\left|#2\right.\right]}

\newcommand{\pa}[1]{\left(#1\right)}
\newcommand{\sqpa}[1]{\left[#1\right]}
\newcommand{\ac}[1]{\left\{#1\right\}}
\newcommand{\ev}[1]{\left\{#1\right\}}
\newcommand{\card}[1]{\left|#1\right|}

\newcommand{\norm}[1]{\left\|#1\right\|}
\newcommand{\onenorm}[1]{\norm{#1}_1}

\newcommand{\abs}[1]{\left|#1\right|}

\newcommand*{\MyDef}{\mathrm{\tiny def}}
\newcommand*{\eqdefU}{\ensuremath{\mathop{\overset{\MyDef}{=}}}}
\newcommand*{\eqdef}{\mathop{\overset{\MyDef}{\resizebox{\widthof{\eqdefU}}{\heightof{=}}{=}}}}
\newcommand{\transpose}{^\mathsf{\scriptscriptstyle T}}

\newcommand{\cA}{\mathcal{A}}

\newcommand{\cE}{\mathcal{E}}
\newcommand{\F}{\mathcal{F}}

\newcommand{\cG}{\mathcal{G}}
\newcommand{\cH}{\mathcal{H}}
\newcommand{\cI}{\mathcal{I}}

\newcommand{\cL}{\mathcal{L}}
\newcommand{\calL}{\cL}

\newcommand{\cO}{\mathcal{O}}
\newcommand{\tcO}{\widetilde{\cO}}
\newcommand{\OO}{\mathcal{O}}
\newcommand{\tOO}{\wt{\OO}}
\newcommand{\cP}{\mathcal{P}}

\newcommand{\Sw}{\mathcal{S}}

\newcommand{\cT}{\mathcal{T}}
\newcommand{\T}{\cT}

\newcommand{\cV}{\mathcal{V}}

\newcommand{\cX}{\mathcal{X}}
\newcommand{\X}{\cX}

\newcommand{\bI}{{\bf I}}
\newcommand{\bM}{{\bf M}}
\newcommand{\bO}{\boldsymbol{O}}
\newcommand{\bp}{\boldsymbol{p}}

\newcommand{\br}{{\bf r}}

\newcommand{\bQ}{{\bf Q}}

\newcommand{\bff}{{\bf f}}

\newcommand{\bk}{{\bf k}}
\newcommand{\bK}{{\bf K}}
\newcommand{\bL}{{\bf L}}

\newcommand{\bu}{{\bf u}}

\newcommand{\bv}{{\bf v}}
\newcommand{\bV}{{\bf V}}
\newcommand{\bw}{{\bf w}}

\newcommand{\bx}{{\bf x}}

\newcommand{\bZ}{{\bf Z}}

\newcommand{\eps}{\varepsilon}
\renewcommand{\epsilon}{\varepsilon}
\renewcommand{\hat}{\widehat}
\renewcommand{\tilde}{\widetilde}
\renewcommand{\bar}{\overline}

\newcommand{\balpha}{{\boldsymbol \alpha}}
\newcommand{\talpha}{\widetilde{\indn}}

\newcommand{\tTheta}{{\widetilde\Theta}}

\newcommand{\bLambda}{{\boldsymbol \Lambda}}

\newcommand{\nothere}[1]{}

\newcommand{\hloss}{\hat\ell}
\newcommand{\bloss}{\boldsymbol  \ell}
\newcommand{\hbl}{\hat{\bloss}}
\newcommand{\hbL}{\wh{\bL}}
\newcommand{\wh}{\widehat}
\newcommand{\ti}{_{t,i}}
\newcommand{\wt}{\widetilde}

\usepackage{xspace}

\newcommand{\LP}{\texttt{LP}\xspace}
\newcommand{\CMG}{\texttt{CMG}\xspace}
\newcommand{\FPL}{\texttt{FPL}\xspace}

\newcommand{\UCB}{\texttt{UCB}\xspace}
\newcommand{\MOSS}{\texttt{MOSS}\xspace}
\newcommand{\UCBE}{\texttt{UCB-E}\xspace}
\newcommand{\ImprovedUCB}{\texttt{ImprovedUCB}\xspace}

\newcommand{\CUCB}{\texttt{CUCB}\xspace}
\newcommand{\EXP}{\texttt{Exp3}\xspace}
\newcommand{\exph}{\EXP}
\newcommand{\LinearTS}{\texttt{LinearTS}\xspace}
\newcommand{\ThompsonSampling}{\texttt{ThompsonSampling}\xspace}
\newcommand{\SpectralEliminator}{\texttt{\textcolor[rgb]{0.5,0.2,0}{SpectralEliminator}}\xspace}
\newcommand{\LinearEliminator}{\texttt{\textcolor[rgb]{0.5,0.2,0}{LinearEliminator}}\xspace}
\newcommand{\LinUCB}{\texttt{LinUCB}\xspace}
\newcommand{\LinRel}{\texttt{LinRel}\xspace}
\newcommand{\KernelUCB}{\texttt{\textcolor[rgb]{0.5,0.2,0}{KernelUCB}}\xspace}
\newcommand{\SupKernelUCB}{\texttt{\textcolor[rgb]{0.5,0.2,0}{SupKernelUCB}}\xspace}
\newcommand{\GPUCB}{\texttt{GP-UCB}\xspace}
\newcommand{\OFUL}{\texttt{OFUL}\xspace}
\newcommand{\OPM}{\texttt{\textcolor[rgb]{0.5,0.2,0}{OPM}}\xspace}
\newcommand{\CLUB}{\texttt{CLUB}\xspace}
\newcommand{\GOBLin}{\texttt{GOB.Lin}\xspace}
\newcommand{\UCBN}{\texttt{UCB-N}\xspace}
\newcommand{\UCBmaxN}{\texttt{UCB-MaxN}\xspace}

\newcommand{\SpectralUCB}{\texttt{\textcolor[rgb]{0.5,0.2,0}{SpectralUCB}}\xspace}
\newcommand{\CheapUCB}{\texttt{\textcolor[rgb]{0.5,0.2,0}{CheapUCB}}\xspace}
\newcommand{\SpectralTS}{\texttt{\textcolor[rgb]{0.5,0.2,0}{SpectralTS}}\xspace}
\newcommand{\SupLinRel}{\texttt{SupLinRel}\xspace}
\newcommand{\SupLinUCB}{\texttt{SupLinUCB}\xspace}
\newcommand{\imb}{\texttt{\textcolor[rgb]{0.5,0.2,0}{IMLinUCB}}\xspace}
\newcommand{\NetBandits}{\texttt{NetBandits}\xspace}
\newcommand{\BARE}{\texttt{\textcolor[rgb]{0.5,0.2,0}{BARE}}\xspace}
\newcommand{\ELP}{\texttt{ELP}\xspace}
\newcommand{\ELPP}{\texttt{ELP.P}\xspace}
\newcommand{\expix}{\texttt{\textcolor[rgb]{0.5,0.2,0}{Exp3-IX}}\xspace}
\newcommand{\expset}{\texttt{Exp3-SET}\xspace}
\newcommand{\expdom}{\texttt{Exp3-DOM}\xspace}
\newcommand{\expg}{\texttt{Exp3.G}\xspace}

\newcommand{\fplix}{\texttt{\textcolor[rgb]{0.5,0.2,0}{FPL-IX}}\xspace}

\newcommand{\expwix}{\texttt{\textcolor[rgb]{0.5,0.2,0}{Exp3-WIX}}\xspace}

\newcommand{\expixt}{\texttt{Exp3-IXt}\xspace}
\newcommand{\expcoop}{\texttt{Exp3-Coop}\xspace}
\newcommand{\expres}{\texttt{\textcolor[rgb]{0.5,0.2,0}{Exp3-Res}}\xspace}

\newcommand{\StoSOO}{\texttt{\textcolor[rgb]{0.5,0.2,0}{StoSOO}}\xspace}
\newcommand{\POO}{\texttt{\textcolor[rgb]{0.5,0.2,0}{POO}}\xspace}
\newcommand{\DOO}{\texttt{DOO}\xspace}
\newcommand{\SOO}{\texttt{SOO}\xspace}
\newcommand{\Zooming}{\texttt{Zooming}\xspace}
\newcommand{\UCT}{\texttt{UCT}\xspace}
\newcommand{\HCT}{\texttt{HCT}\xspace}
\newcommand{\SHOO}{\POO}
\newcommand{\HOO}{\texttt{HOO}\xspace}
\newcommand{\ATB}{\texttt{ATB}\xspace}
\newcommand{\TZ}{\texttt{TaxonomyZoom}\xspace}
\newcommand{\Direct}{\texttt{DiRect}\xspace}
\newcommand{\SiRI}{\texttt{\textcolor[rgb]{0.5,0.2,0}{SiRI}}\xspace}

\newcommand{\greedy}{\texttt{Greedy}\xspace}
\newcommand{\opm}{\texttt{\textcolor[rgb]{0.5,0.2,0}{OPM}}\xspace}

\newcommand{\reg}{\gamma}
\newcommand{\hmu}{\hat{\mu}}

\newcommand{\hs}{\hat{\sigma}}

\newcommand{\etat}{\eta_t}
\newcommand{\gammat}{\gamma_t}
\newcommand{\nodes}{{\textcolor[rgb]{0.3,0.8,0.0}{N}}}
\newcommand{\rounds}{{\textcolor[rgb]{0.3,0.0,0.8}{T}}}
\newcommand{\td}{{\textcolor[rgb]{0.6,0.0,0.6}{\tilde{d}}}}
\newcommand{\matL}{{\textcolor[rgb]{0.6,0.0,0.6}{L}}}
\newcommand{\matK}{{\textcolor[rgb]{0.6,0.0,0.6}{K}}}
\newcommand{\effd}{{\textcolor[rgb]{0.6,0.0,0.6}{d}}}

\newcommand{\indn}{{\textcolor[rgb]{0.6,0.0,0.6}{\alpha}}}
\newcommand{\indnstar}{{\textcolor[rgb]{0.6,0.0,0.6}{\alpha^\star}}}
\newcommand{\cliquen}{{\textcolor[rgb]{0.6,0.0,0.6}{\chi}}}
\newcommand{\erdosr}{{\textcolor[rgb]{0.6,0.0,0.6}{r}}}
\newcommand{\detD}{{\textcolor[rgb]{0.6,0.0,0.6}{D}}}
\newcommand{\detDstar}{{\textcolor[rgb]{0.6,0.0,0.6}{D_\star}}}
\newcommand{\infibeta}{{\textcolor[rgb]{0.6,0.0,0.6}{\beta}}}
\newcommand{\mas}{{\textcolor[rgb]{0.6,0.0,0.6}{\texttt{mas}}}}
\newcommand{\nodeset}{\cV}
\newcommand{\edgeset}{\cE}
\newcommand{\regret}{R_\rounds}

\newcommand{\sumT}{\sum_{t = 1}^\rounds}
\newcommand{\sumt}{\sum_{t=1}^\rounds}

\newcommand{\sumi}{\sum_{i=1}^{\nodes}}

\newcommand{\tj}{_{t,j}}

\newcommand{\pti}{p\ti}

\newcommand{\ptj}{p_{t,j}}

\newcommand{\oti}{o\ti}

\newcommand{\loss}{\ell}
\newcommand{\hLoss}{\hat{L}}
\newcommand{\hL}{\wh{L}}
\newcommand{\noise}{\xi}

\newcommand{\dnew}{\effd_{\scriptsize\mbox{new}}}
\newcommand{\gweight}{s}
\newcommand{\avgalpha}{\indnstar_{\text{avg}}}
\newcommand{\rkdual}{r_k^{\circ}}

\newcommand{\ridual}{r_i^{\circ}}
\newcommand{\rstardual}{r_\star^{\circ}}

\newcommand{\Ddualset}{\mathcal D^{\circ}}
\newcommand{\node}[2]{(#1,#2)}

\newcommand{\todo}[2][]{}
\newcommand{\todom}[1]{}
\newcommand{\todomi}[1]{}

\usepackage{algorithm}
\usepackage{algorithmic}
\usepackage{subcaption}
\usepackage{float}
\usepackage{wrapfig}

\newcommand{\citep}{\parencite}
\newcommand{\citet}{\textcite}

\begin{document}

\begingroup
\thispagestyle{empty}
\begin{tikzpicture}[remember picture,overlay]
\coordinate [below=12cm] (midpoint) at (current page.north);
\node at (current page.north west)
{\begin{tikzpicture}[remember picture,overlay]
\node[anchor=north west,inner sep=0pt] at (0,0) {\includegraphics[width=\paperwidth]{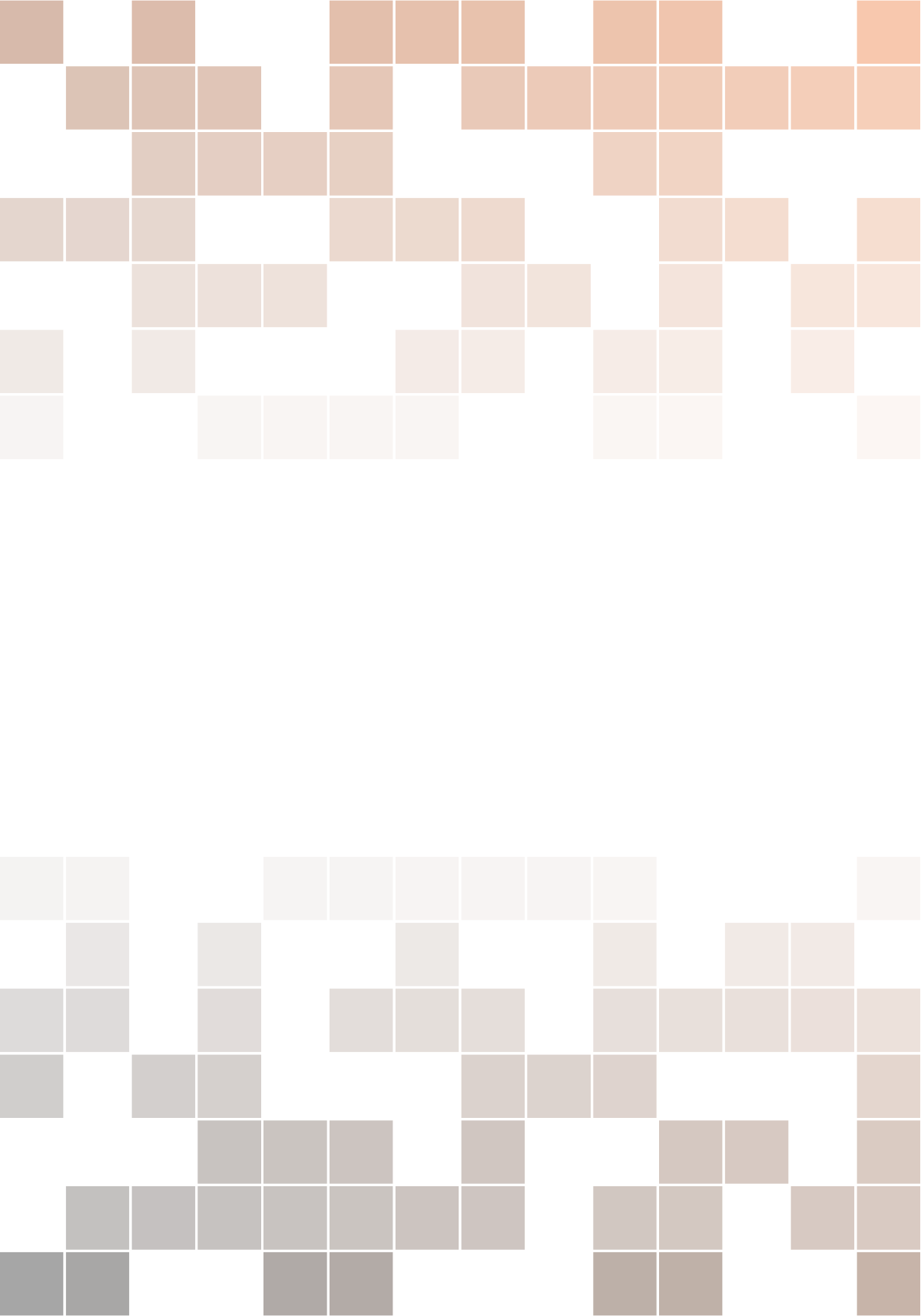}}; 
\draw[anchor=north] (midpoint) node [fill=ocre!30!white,fill opacity=0.6,text opacity=1,inner sep=1cm]{\Huge\centering\bfseries\sffamily\parbox[c][][t]{\paperwidth}{\centering Bandits on Graphs and Structures\\[15pt] 
{\Large habilitation \`a diriger des recherches}\\[20pt] 
{\huge Michal Valko}}}; 
\end{tikzpicture}};
\end{tikzpicture}
\vfill
\endgroup


\newpage
~\vfill
\thispagestyle{empty}
{\fontfamily{ptm}\selectfont 

 \vspace{-3cm} \hspace{-1cm}\fbox{
\begin{minipage}[c]{16cm}
\vspace{.2cm}
\begin{center}
{\large\sc\bf\sffamily HABILITATION \`A DIRIGER DES RECHERCHES DE
\\
\vspace{.1cm}
L'\'ECOLE NORMALE SUP\'ERIEURE DE CACHAN} 
\vspace{.2cm}
\end{center}
\end{minipage}
}

\vspace{1.0cm}
\begin{center}
  \includegraphics[width=0.2\columnwidth]{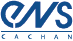}
\end{center}
\vspace{0.5cm}

\begin{center}
{\bf {\sffamily Habilitation \`a Diriger des Recherches}}

\vspace{0.8cm}

{\sffamily Sp{\'e}cialit{\'e} :} {\bf\sffamily Math\'ematiques}\\

{\sffamily \vspace{1cm} pr\'esent\'ee par \vspace{.5cm}}

{\large \bf\sffamily  Michal VALKO} \\
\vspace{0.3cm}
 \includegraphics[width=0.10\columnwidth]{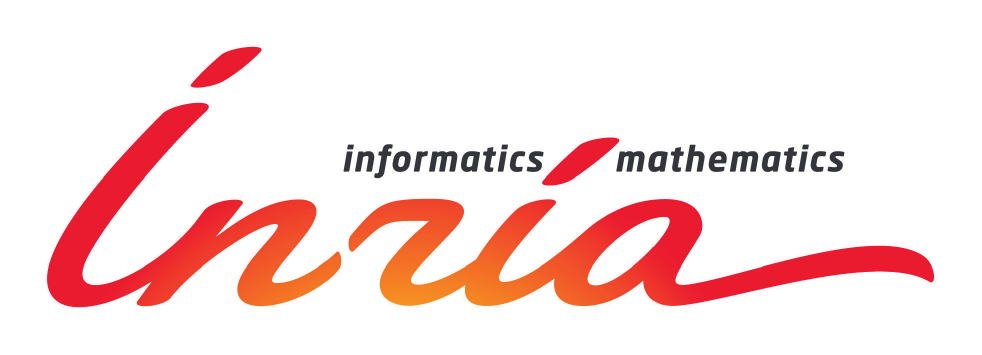}
\end{center}

\vspace{1cm}
\begin{center}
\hspace{-0.3cm}\hrulefill \hspace{.2cm}

\vspace{.2cm}

\hspace{-.5cm}{\bf \Large \sffamily{Bandits on Graphs and Structures}}

\vspace{-.1cm}

\hspace{-0.3cm}\hrulefill \hspace{.2cm}

\end{center}

\vspace{2cm}
{\sffamily

\begin{tabular}{lllll}
{\bf\sffamily Rapporteurs :} 
& {M.}   & Aur\' elien & {\bf\sffamily GARIVIER} & Universit\' e Toulouse\\
& {M.}   & G\'abor & {\bf\sffamily LUGOSI}     & Universitat Pompeu Fabra\\
& {M.}   & Vianney  & {\bf\sffamily PERCHET}      & ENSAE ParisTech \\
\end{tabular}

{\sffamily\vspace{1.5cm} Soutenue le {\bf\sffamily 15 Juin 2016} devant le jury
compos{\'e} de}
}
\bigskip

\bigskip

{\sffamily
\begin{tabular}{lllll}
{M.}   & Nicolas & {\bf\sffamily VAYATIS}  & ENS de Cachan  & Garant \& Examinateur  \\ 
{M.}   & Aur\' elien & {\bf\sffamily GARIVIER}  & Universit\' e Toulouse & Pr\' esident \& Rapporteur\\
{M.}   & G\'abor & {\bf\sffamily LUGOSI}      &  Universitat Pompeu Fabra & Rapporteur\\
{M.}   & Vianney & {\bf\sffamily PERCHET}    & ENSAE ParisTech & Rapporteur\\
{M.}   & Nicol\`o  & {\bf\sffamily CESA-BIANCHI}      &   Universit\`a  di Milano   & Examinateur\\
{M. } & Mark & {\bf\sffamily HERBSTER}      & University College London & Examinateur\\
{M. } & R\' emi & {\bf\sffamily MUNOS}      &  DeepMind \& Inria & Examinateur
\end{tabular}
}
%
%
%
%


\chapterimage{chapter_head_1.pdf} 
\pagestyle{empty} 
\tableofcontents 
\newpage


{\sffamily \noindent Many thanks to the committee \dots}
 \begin{center}
  \includegraphics[width=1\columnwidth]{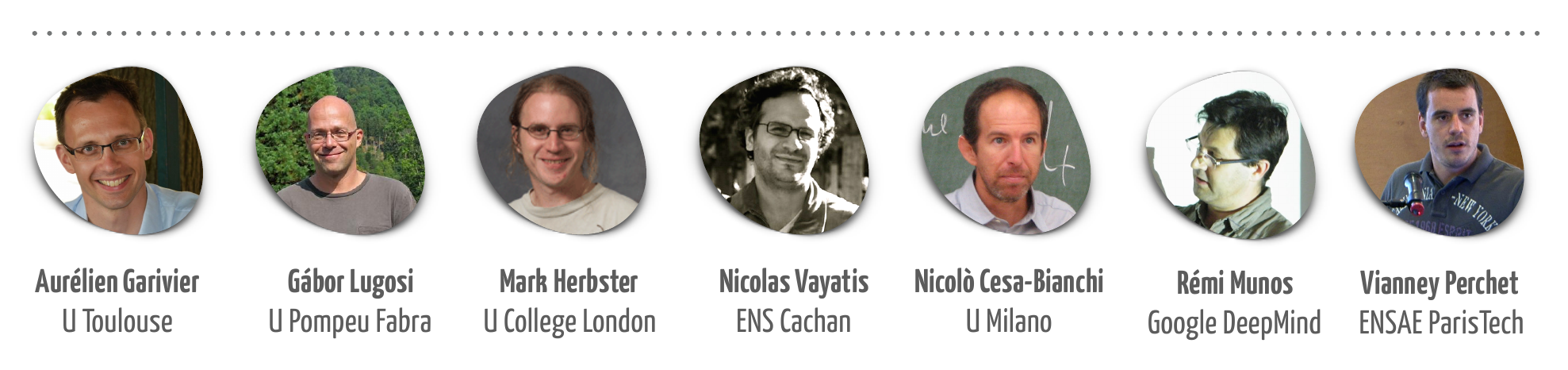}
 \end{center}

{\sffamily  \hspace{-0.25cm} \dots coauthors \dots}
 \begin{center}
  \includegraphics[width=1\columnwidth]{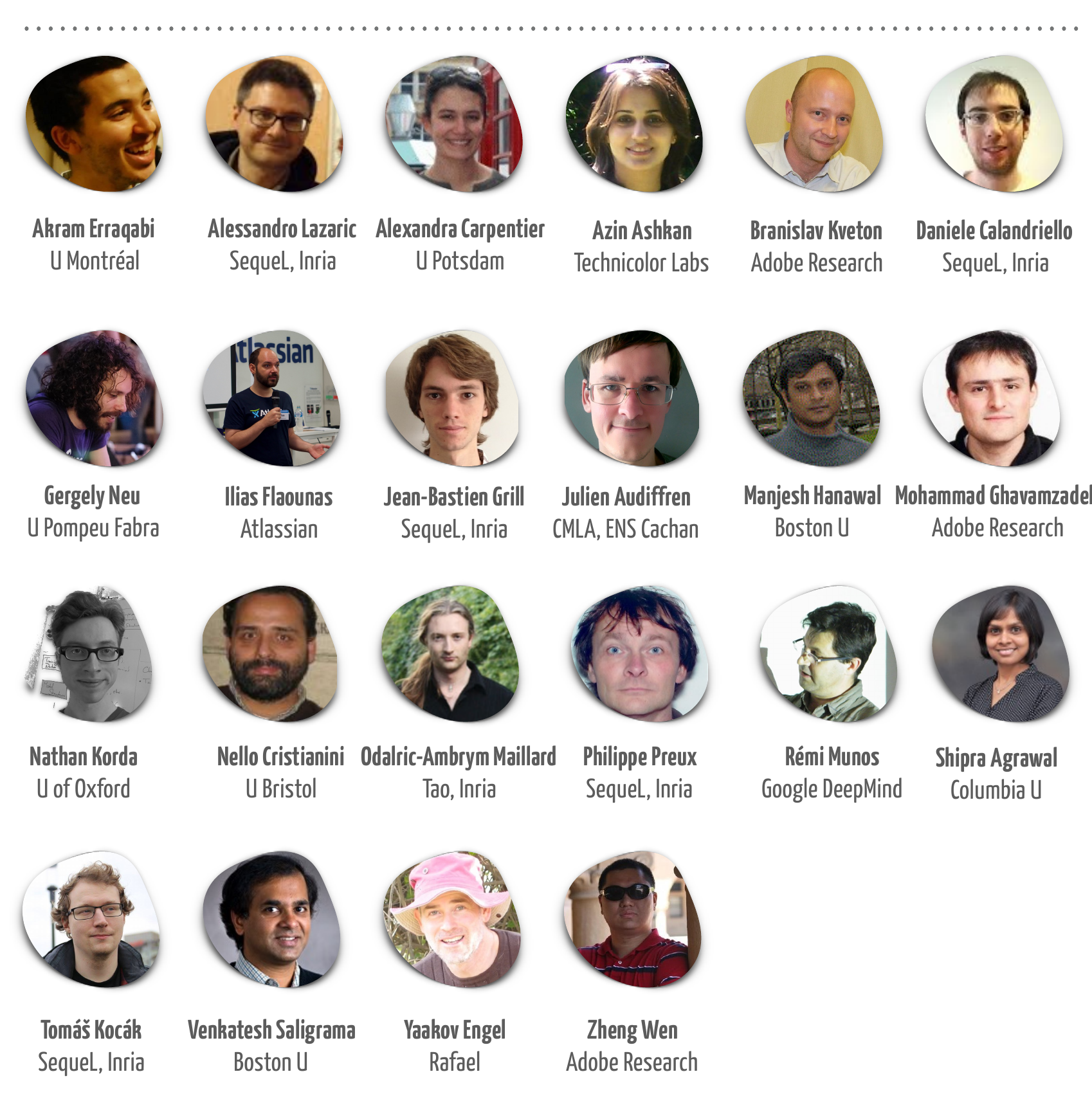}
 \end{center}

{\sffamily \noindent \dots friends, colleagues, supporters, reviewers, students, future readers, and family.}

\cleardoublepage 
\pagestyle{fancy} 
\chapter*{Summary}
\addcontentsline{toc}{chapter}{\textcolor{ocre}{Summary}}

The goal of this thesis is to investigate the \emph{structural} properties of certain \emph{sequential} problems in order to 
bring the solutions closer to a practical use. In the first part, we put a special emphasis on structures, that 
can be represented as \emph{graphs} on actions, in the second part we study the large action 
spaces that can be  of \emph{exponential} size in the number of base actions or even \emph{infinite}. 

\section*{Graph bandits}
\emph{Bandit problems} are online decision-making problems where the only feedback given to the learner is a (noisy) reward of the chosen decision. In early sequential decision-making research, we treated each of the decisions \emph{independently}. While this is enough when the number of actions is very small, it becomes difficult (both theoretically and in practice) when the set of potential actions comprises larger sets, such as a set of movies or products in a recommender system. The minimax regret guarantees scale as $\Theta(\sqrt{\nodes \rounds})$, where $\nodes$ is the number of actions and $\rounds$ is the time horizon. If $\nodes$ happens to be large (such as the number of movies that can be in millions), these guarantees are weak. Luckily, the problems become easier if there is an \emph{efficient information sharing} between the actions.
For instance, we will study the benefits of homophily (similar actions give similar rewards) and side information.
 Our goal is to take advantage of these similarities in order to (provably) learn \emph{faster}. With respect to the guarantees we aspire  to give, we aim to replace $\nodes$ (number of actions = number of nodes in a graph) with some \emph{graph-dependent quantity}, possibly smaller than $\nodes$ if the graph structure is helpful.
This part aspires to be a \emph{survey} of all work done in graph bandits.

\paragraph{Smoothness of rewards in graph bandits}
Consider an example when the actions are  nodes on a given graph, e.g.,  recommending a movie in a graph of movies with movie similarities on edges. Then in many realistic situations, the rewards (i.e., the ratings) are smooth on this graph. This smoothness in graph bandits is a structural property that we leveraged in \emph{spectral bandits} \parencite{valko2014spectral} to deliver a \UCB-style algorithm \parencite{burnetas1996optimal,auer2002finite} whose performance does not scale with the number of nodes but rather with the \emph{effective dimension}, a quantity related to the \emph{number of relevant eigenvectors} of the associated graph Laplacian, which is often much smaller in practice. Later, we gave a computationally faster algorithm based on \ThompsonSampling \parencite{kocak2014spectral}.  Furthermore, we extended the results to the cost-sensitive setting called \emph{cheap bandits} \parencite{hanawal2015cheap}, relevant for radar applications, where the reward is an average of several nodes. 

\paragraph{Side observations in graph bandits}
Another setting, where we can learn faster using a graph structure, is the case when we receive some \emph{side information}, as formalized by \textcite{mannor2011from}. This setting is a \emph{partial observability model}, capturing situations where the information conveyed to the learner is between the full information and the bandit feedback. In the simplest variant, we assume that in addition to its own loss, the learner also gets to observe losses of some other actions --- neighbor nodes on a graph. For instance, if we optimize click-through rate in the newsfeed recommendation, the additional information  comes from the fact that several news feeds can refer to the same content, giving us the  opportunity to infer clickthroughs for a number of assignments. This additional feedback allows for new algorithms where the regret guarantees improve from $\cO(\sqrt{\nodes\rounds})$ to $\cO(\sqrt{\indn \rounds})$ where~$\indn$ is the \emph{independence number} of the graph and can range from $1$ (complete graphs) to $\nodes$ (empty graphs). With the new \emph{implicit exploration} technique employed in the \expix algorithm  \parencite{kocak2014efficient}, we were able to relax the assumption needed by \textcite{alon2013from} which required the knowledge of the graph before the action was chosen. This relaxation was also important, because it removed the costly exploration phase previously needed, including a computation of the dominating set in each time step.  It also removed the need for the doubling trick and the need to aggregate several algorithms to appropriately tune the learning rate. We further extended the setting to the \emph{combinatorial} side observation case, when the learner can select a subset of the nodes (potentially constrained by  combinatorial constraints). This can be relevant for example in learning of bipartite matching between the users and the recommendations. We provided the \FPL version of implicit exploration for this setting, achieving the regret of $\cO(m^{3/2}\sqrt{\indn \rounds})$, where $m$ is the number of nonzero components. Finally, in some graphs, the side observations are perturbed by \emph{noise}. This is the case in \emph{sensor networks}, where the communication reliability is a decreasing function of the distance. For this case, we extended this setting to the noisy information case \parencite{kocak2016online}, delivering an algorithm with $\cO(\sqrt{\indnstar \rounds})$, where $\indnstar$ is the \emph{effective independence number}.

\paragraph{Influence maximization in graph bandits}

For most of the graph bandits approaches, the algorithms need to have access to at least some portions of the graph to properly update their loss estimates. This is however not always possible in practice. Therefore, we formalized revealing bandits \parencite{carpentier2016revealing}, for the \emph{influence maximization} in the stochastic setting, where the only information that the learner receives is the set of the influenced nodes.  The goal is to find the most influential node without knowing the probabilities of influence between the node pairs. For this setting, we delivered a minimax optimal algorithm, \emph{bandit revelator} (\BARE) and showed that it attains $\cO(\sqrt{\detDstar \rounds})$ cumulative regret, where$\detDstar$ is the \emph{detectable dimension}, small when the unknown underlying graph has a small number of very influential nodes.
Our setting considers only a simple model of \emph{local} influence, but opens a way to study influence 
maximization in a sequential setting for more global influence models known 
in \emph{computational social sciences} \parencite{kempe2015maximizing}.
\section*{Stochastic bandits in large structured domains}
Not all structured action spaces are graphs and in the second part we address bandit setting 
with \emph{large} (exponential or infinite) action spaces with \emph{stochastic} rewards.
While in the case of graph bandits we could opt to ignore the graph and use known multi-arm bandit
strategies whose regret is worse by scaling with number of nodes $\nodes$; when the number 
of actions is exponential or infinite, this is no longer even an option. Using the \emph{structure} 
in the large-action setting \emph{is essential}.

\paragraph{Kernel bandits}
Linear (contextual) bandits allow for information sharing between the rewards through the unknown, but \emph{fixed} 
vector of weights via linear combination of features. This allows the regret bounds of  linear bandit algorithms (\LinRel, 
\cite{auer2002using}; \LinUCB, \cite{li2010contextual}; \OFUL, \cite{abbasi2011improved}; \LinearTS, \cite{agrawal2013thomson}; \LinearEliminator, \cite{valko2014spectral})
to scale not with the number of actions, but with the size of the vector of weights $D$, i.e., the dimension of the context.
In \emph{kernel bandits}, the rewards may be a  smooth function of contexts as given by some similarity function, while the set of the decision can still be discrete (\GPUCB, \cite{srinivas2009gaussian}; \KernelUCB, \cite{valko2013finite}). However, when the reward is an arbitrary linear function of context in the related RKHS, the dimension of the fixed vector of weights expressing the linear
combination can be infinite, and, therefore, the straightforward use of linear bandit analysis
does not apply. We show, however, that the regret of \KernelUCB can be bounded in terms of the
\emph{effective dimension}, measuring the decay of eigenvalues of the covariance matrix in kernel regression
and show its link to the \emph{maximal information gain},  appearing in the analysis of \GPUCB.

\paragraph{Polymatroid bandits}
When the potential actions in sequential learning form a combinatorial set, a very useful and specific structure is a \emph{polymatroid}, where the associated offline learning problem with a known model can be solved optimally by a simple \greedy algorithm. Since some interesting problems are instances of sequential optimization on a polymatroid (e.g., the recommendation of \emph{diverse items}) we propose and analyze \OPM \parencite{kveton2016learning}, optimistic polymatroid optimization algorithm  suited for this setting with a regret bound scaling with the \emph{rank of the polymatroid}.


\paragraph{Bandits for function optimization}
A very general structure is an arbitrary function. If we are to optimize such function sequentially, we may do it by extending the ideas from the bandit theory. In this case, the decision set is continuous, and in the more challenging setting, the (reward) function may be only locally smooth around one of its optima, while the \emph{smoothness is unknown}. We first provided an algorithm for an easy class of functions (\StoSOO, \cite{valko2013stochastic}, \cite{preux2014bandits}) and later for a much wider class of difficult-to-optimize functions (\POO, \cite{grill2015black-box}).

\paragraph{Infinitely many-arms bandits}
Even in the \emph{continuous} case when \emph{no topology} on the decision set is given to the learner, the learner still can take advantage of the structure if there is a certain quantity of near-optimal decisions. This setting was formally defined by \textcite{berry1997bandit} as \emph{infinitely many-arms bandits}, when the learner can sample from an infinite pool of decisions. Our contribution is the \SiRI algorithm \parencite{carpentier2015simple}, that is nearly minimax optimal in the \emph{simple regret} setting and potentially useful in the best feature selection when we face a large number of candidates. The regret guarantees  depend on a $\infibeta$-parameter  characterizing the distribution of
near-optimal arms.

%
\vskip 1cm
\paragraph{Notation }
By default, we use $\rounds$ for the number of rounds and $\nodes$ for the number of actions, e.g., the number of nodes 
in a graph. For any $a \in \NN^+,$
$[a]$ stands for the set of first $a$ positive natural numbers, 
$[a] \eqdef \{1, 2, 3, \dots, a \}$.
The rest of the notation is introduced when it is used. We also highlight the graph- or problem-dependent 
quantities as $\effd, \indn, \indnstar, \cliquen, \mas, \erdosr, \detDstar, \matL, \matK,\td,$ and $\infibeta$.


\part{Graph bandits}


\chapterimage{chapter_head_2.pdf} 
\noindent
In this section, we consider the variations of the following setting. 
There is a (known or unknown) graph $\cG$, with the node set 
$\nodeset = \left\{v_1, \dots, v_\nodes \right\}$ of $\nodes$ nodes
and the edge set $\edgeset$. 
Every round $t = 1, \dots, \rounds,$ the learner picks $I_t$-th node (decision, arm, or action) $v_{I_t}$. 
At the same time, the environment (possibly adversarial) independently picks 
a vector of losses $\bloss_t \in [0,1]^\nodes$ and the learner
suffers the loss  $\loss_{t,I_t}$. As usual in online settings \citep{cesa-bianchi2006prediction}, the performance 
is measured in terms of (total expected) regret, which is the difference between
a total loss received and the total loss of the best single action chosen in
hindsight,
 \begin{align}
 \label{def:regretgeneral}
 R_\rounds = \max_{i\in [\nodes]} \EE{\sum_{t=1}^\rounds \left(\loss_{t,I_t} - \loss_{t,i}\right)} 
 \end{align}
 where the expectation integrates over the random choices made by the learning
algorithm. Equivalently, instead of losses $\bloss_t,$ we can consider rewards $\br_t$.
If the nodes have associated context vectors, we will denote them by $\bx_1, \dots, \bx_\nodes,$
with $\bx_i \in \R^D$.  The goal of the learner is to minimize  $R_\rounds$.

\paragraph{General lower bound} Every setting in this part can be potentially
treated with known algorithms and analyses for multi-arm bandits by \emph{ignoring 
the presence of the graph} and considering the nodes as independent decisions.
Ignoring the graph structure and using the known algorithms 
for either the stochastic or adversarial bandits is limited by the following lower bound.
\begin{theorem}[Multi-arm bandit lower bound by \textcite{auer2002nonstochastic}]
\label{thm:lowerboundauer}
For any number of actions $\nodes \geq 2$, there exists a distribution of losses (rewards) such that
\[
\EE{R_\rounds} \geq \frac{1}{20} \min \left \{ \sqrt{\nodes\rounds}, \rounds \right \}.  
\]
\end{theorem}
This part  addresses the situations a with large number of nodes (actions, decisions, arms) $\nodes$, which
makes the regret scale unfavorably with $\nodes$. In the rest of this chapter 
we describe the approaches that leverage the graph structure for specific 
settings with the regret bounds that essentially replace~$\nodes$ 
in the lower bound of Theorem~\ref{thm:lowerboundauer}
 with a possibly much smaller, \emph{graph-dependent} quantity.

In the following three chapters, we consider three groups of graph bandit setups. For
each of them, we aim to find efficient regret-minimization algorithms and find out an appropriate
problem dependent quantity:

\begin{itemize}
\item  \textbf{Smoothness of rewards} on a given graph is a setup where we exploit situations where neighboring nodes give similar rewards. In a specific, \emph{spectral bandit} setting, we show that the relevant quantity is the \emph{number of relevant 
eigenvectors} of the Laplacian of $\cG$.
\item  \textbf{Side observations} extend the bandit feedback to case where we receive \emph{additional} reward information from the nodes adjacent to $v_{I_t}$. We show that several variations of this setup are linked to the 
\emph{independence number} of $\cG$.
\item  \textbf{Influence maximization} is a problem with a specific reward structure that is a function of the whole graph. In this setting, we aim at finding the nodes with the maximum influence on the rest of the nodes. For a simple
model of \emph{local influence}, we define a new \emph{detectable dimension} of the problem.
\end{itemize}

\chapter{Smoothness of rewards}\label{chap:smoothness}
\begin{wrapfigure}[16]{R}{0.40\textwidth}
\vspace{-65pt}
\hspace{5pt}
\includegraphics[width=2.2in]{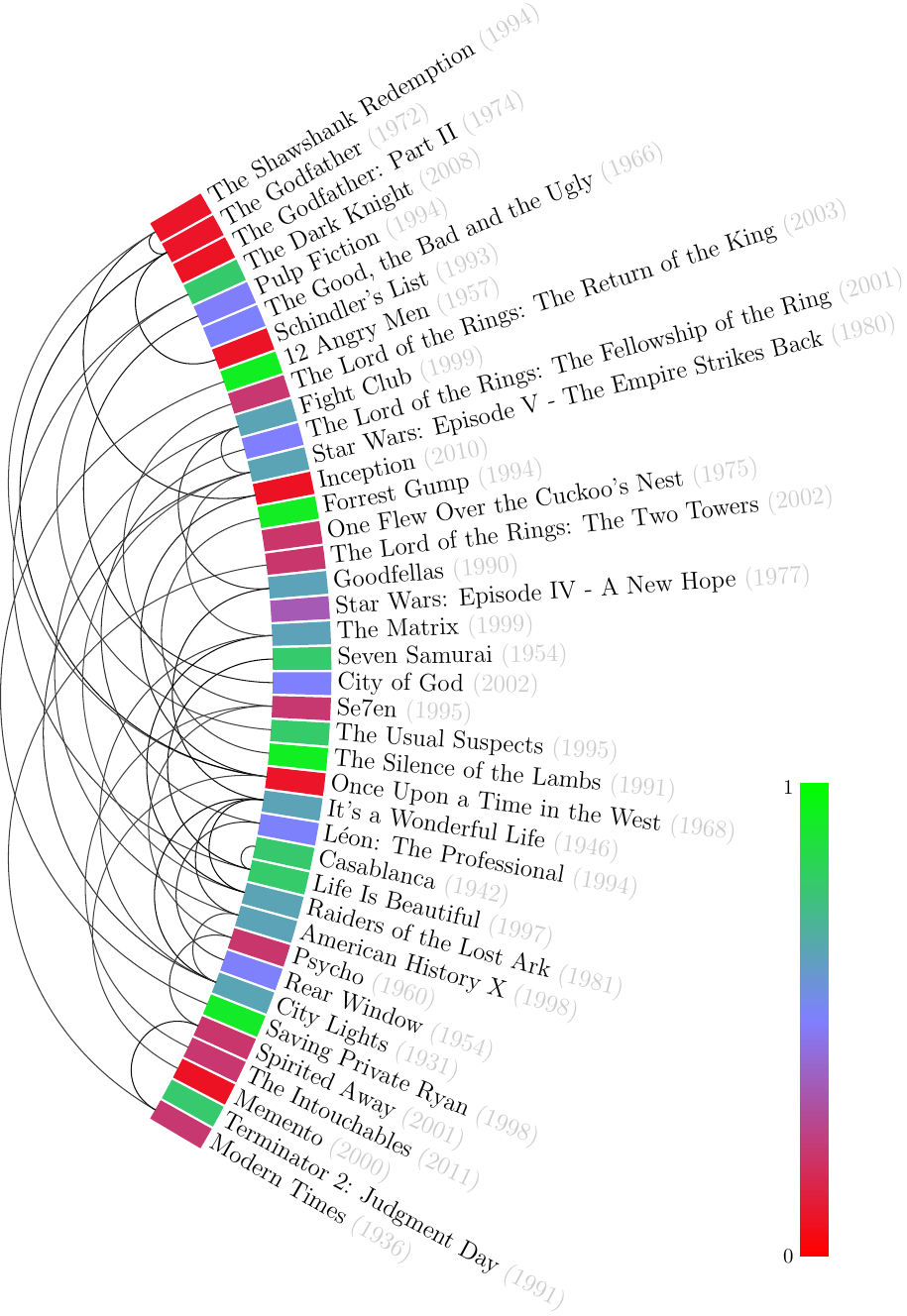}
\caption{User preference for movies.}
\label{fig:userpreference}
\vspace{-3cm}
\end{wrapfigure}

\emph{A smooth graph function} is a function on a graph that returns similar
values on neighboring nodes. This concept arises frequently in manifold and
semi-supervised learning \parencite{zhu2008semi-supervised}, and reflects the fact
that the outcomes on the neighboring nodes tend to be similar. It is well-known
\parencite{belkin2006manifold,belkin2004regularization} that a smooth graph function
can be expressed as a linear combination of the eigenvectors of the graph
Laplacian with smallest eigenvalues. Therefore, the problem of learning such a
function can be cast as a regression problem on these eigenvectors. We will  
bring this concept to bandits as a special case of the (stochastic) setting defined in the beginning of
this part. 
In particular, we study a
bandit problem where the arms are the nodes of a graph and the expected payoff
of pulling an arm is a smooth function on this graph.

 One
application is \textit{targeted advertisement} in social networks. Here, the
graph is a social network and our goal is to discover a part
of the network that is interested in a given product. Interests of people in a
social network tend to change smoothly \parencite{mcpherson2001birds}, because
friends tend to have similar preferences. Therefore, we take advantage of
this structure and formulate this problem as learning a smooth preference
function on a graph.

Another application of our work is \textit{recommender systems}
\parencite{jannach2010recommender}. In content-based recommendation
\parencite{chau2011apolo},
the user is recommended items that are
similar to the
items that the user rated highly in the past. The assumption is that users
prefer similar items similarly. The similarity of the items can be measured for
instance by a nearest neighbor graph \parencite{billsus2000learning}, where
each item is a node and its neighbors are the most similar items (Figure~\ref{fig:userpreference}).

In both  applications described above, the learner (advertiser)
has rarely the budget (time $\rounds$) to try all the options even once. 
Furthermore, imagine that the learner is a movie recommender system
and would ask the user to rate all the movies before it starts producing relevant 
recommendations. Such a recommender system would be of little value. 
Yet, many bandit algorithms start with pulling each arm once. 
This is something that we cannot afford here and therefore, contrary to standard  
 bandits, we consider the case $\rounds \ll \nodes$.

\begin{figure}
  \begin{center}
\includegraphics[viewport = 112 348 497 558,clip,width=0.45\columnwidth]{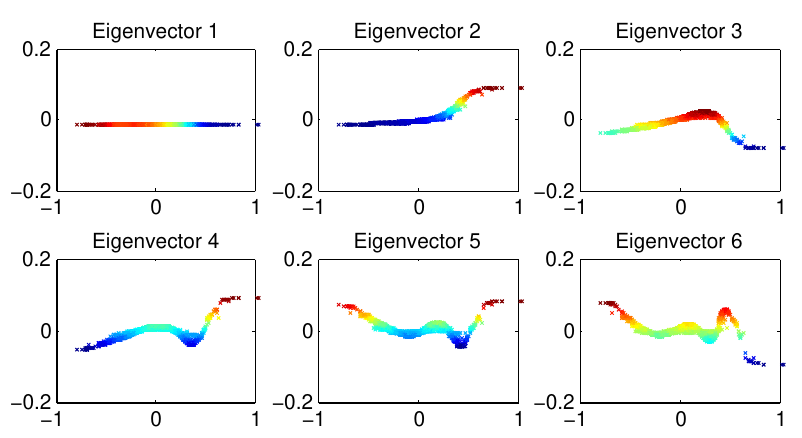}
\includegraphics[width=0.32\columnwidth]{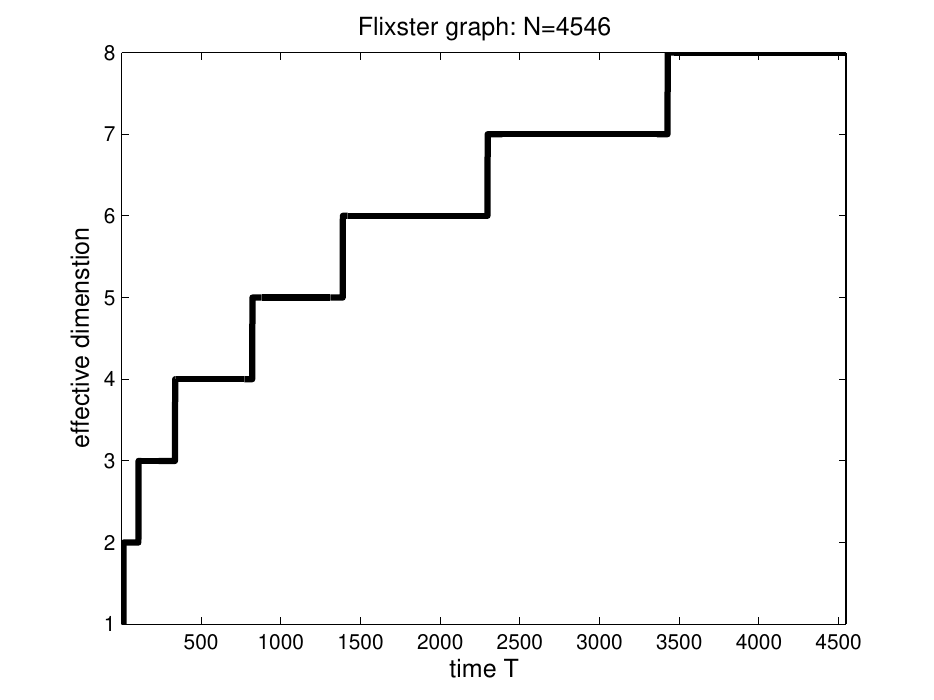}
\caption{\textbf{Left:} Eigenvectors from the Flixster data corresponding to the smallest
few eigenvalues projected onto the first principal component. Colors
 indicate the values. \textbf{Right}: Effective dimension as a function of time.}
  \label{fig:flixster_eigenvectors}
 \end{center}
\end{figure}

\section{Spectral bandits}\label{sec:spectral}

If the smooth graph function can be expressed as a linear combination of $k$
eigenvectors of the graph Laplacian, and $k$ is small and known, our learning
problem can be solved using ordinary linear\footnote{spectral bandits are therefore a special subcase  of kernel bandits~\parencite{srinivas2009gaussian,valko2013finite}} bandits
\parencite{auer2002using,li2010contextual,agrawal2013thomson}. In practice, $k$ is problem specific and
unknown. Moreover, the number of features $k$ may approach the number of nodes
$\nodes$. Therefore, proper regularization is necessary, so that the regret of the
learning algorithm does not scale with $\nodes$. We are interested in the setting
where the regret is independent of $\nodes$ and, therefore, this problem is
nontrivial.

There are several  ways to define the \textit{smoothness} of the function $f$ 
with respect to the graph $G$. We are using the one that is standard in spectral 
clustering and semi-supervised learning, defined as
\begin{align}
\label{eq:smoothness}
S_G(f) = \frac{1}{2}\sum_{i,\,j \in [\nodes]}^\nodes w_{i,j}(f(i)-f(j))^2 = \bff\transpose\cL\bff = \bff\transpose\bQ\bLambda\bQ\transpose\bff = \|\balpha\|^2_{\bLambda} = \sum_{i = 1}^\nodes \lambda_i \alpha_i^2,
\end{align}
where $\bff = (f(1),\,\dots,\,f(\nodes))\transpose$ is the vector of the function values, 
$\bQ\bLambda\bQ\transpose$ is the eigendecomposition of graph laplacian $\cL$
where $\bLambda$ is the diagonal matrix with entries $0 = \lambda_1\leq
\lambda_2\leq \dots\leq \lambda_\nodes$, and 
$\balpha = \bQ\transpose\bff$ is the representation of vector $\bff$ in the eigenbasis. 
The assumption on  smoothness of the reward function with respect to the 
underlying graph is reflected in our belief that the value of the $S_G(f)$ is small 
and therefore, components of $\balpha$ corresponding to the large eigenvalues 
should be small as well. If $\balpha^\star$ it the true (unknown) parameter vector, 
then in the stochastic setting, the reward of the chosen node $v_{I_t}$ is assumed to be
\[
  r_t =\langle \bx_{I_t} , \balpha^\star \rangle + \varepsilon_t,
\]
where the noise $\varepsilon_t$ is be $R$-sub-Gaussian for any $t$.
In our setting, we have $\bx_v \in \R^D$ and $\|\bx_v\|_2\leq 1$
for all $\bx_v$.  The goal of the recommender is to minimize the cumulative
regret with respect to the strategy that always picks the best nodes
w.r.t.~$\balpha^\star$ and the definition of cumulative (pseudo) regret~\eqref{def:regretgeneral} simplifies to
\[
R_\rounds  = \rounds \max_{v \in \nodeset}  f_{\balpha^\star}(v) -  \sum_{t=1}^\rounds f_{\balpha^\star}\left(v_{I_t}\right), 
\text{ where } f_{\balpha^\star}(v) = \langle \bx_{v} , \balpha^\star \rangle.
\]

\subsection{Effective dimension for spectral bandits}
The main benefit of expressing the reward vector in 
the spectral basis of the graph Laplacian
 \eqref{eq:smoothness} is that when only $\effd \ll \nodes$ eigenvectors
 are enough to express the reward function well, then 
 we can learn faster than with linear bandits, where 
 the regret scales with $D$ that denotes the ambient dimension, 
 which is equal to $\nodes$ in the spectral setting.

In general, we assume 
a set of $K$ vectors $\bx_1,\dots, \bx_K\in\R^\nodes$ such that $\|\bx_i\|_2\leq 1$
for all $i$. For the spectral bandits, we have $K = \nodes$.
Moreover, since $\bQ$ is an orthonormal matrix, $\|\bx_i\|_2 = 1$.
Finally, since the first eigenvalue of a graph Laplacian is always zero,
 $\lambda^\cL_1 = 0$, we use $\bLambda = \bLambda_{\cL} + \lambda \bI$,
 with some positive regularizer $\lambda$,
in order to have $\lambda_1 = \lambda > 0$. 
In order to present our algorithms and analyses,
we introduce a notion of \textit{effective dimension} $\effd$.
\begin{definition}\label{def:effectived}
 Let the \textbf{effective dimension} be the largest $\effd$ such
that:
\[(\effd-1) \lambda_\effd  \leq \frac{\rounds}{\log(1 + \rounds/\lambda)}\]
\end{definition}
The effective dimension $\effd$ (whose precise definition comes from our analysis) is small when the coefficients $\lambda_i$ grow
rapidly above $\rounds$. This is the case when the dimension of the space $D$ (and
$K$) is much larger than $\rounds$, such as in graphs from social networks with very
large number of nodes $\nodes$. In contrast, when the coefficients are all small,
then $\effd$ may be of the order of $\rounds$, which would make the regret bounds useless.
Figure~\ref{fig:flixster_eigenvectors}  (right) shows how $\effd$ behaves compared to $D$ on 
generated graphs and graphs built from real-world data.\footnote{We set $\bLambda$ to
$\bLambda_{\cL} + \lambda \bI$ with $\lambda = 0.01$,
where $\bLambda_{\cL}$ is the graph Laplacian of the respective graph.}

The dependence of the effective dimension on $\rounds$ comes from the fact,
that $\effd$ is related to the number of \emph{nonnegligible} dimensions
characterizing the space where the solution to the penalized least-squares lies, since this solution is  constrained to an ellipsoid defined
by the inverse of the eigenvalues.
 In fact, for a small $\rounds$, the axes of the ellipsoid corresponding to the large eigenvalues of $\cL$ are negligible with respect to the overall regret.
Therefore, when $\rounds$ tends to infinity, all directions matter, thus the solution can be
anywhere in a (bounded) space of dimension~$D = \nodes$. On the contrary, for a smaller
$\rounds$, the
ellipsoid possesses a smaller number of nonnegligible dimensions.
Notice that it is natural that this effective dimension depends on $\rounds$ as
we consider the setting $\rounds<\nodes$.
If we
wanted to avoid $\rounds$ in the definition of $\effd$, we could define it as well in
terms
of~$\nodes$ by replacing$\rounds$ by $\nodes$ in Definition~\ref{def:effectived}, but this
would
only loosen its value.

\paragraph{Lower bound}
While the known lower bound for linear bandits is $\Omega(\sqrt{D\rounds})$, 
we can show a similar lower bound for spectral bandits $\Omega(\sqrt{\effd\rounds})$, featuring 
the effective dimension.  The main idea is to construct a graph 
composed of $\effd$ almost disconnected components (Figure~\ref{fig:lowerbound_graph})
and then reduce the setting to
 $\effd$-arm bandits with $\Omega(\sqrt{\effd\rounds})$ lower bound (Theorem~\ref{thm:lowerboundauer}).

\begin{figure}
 \begin{center}
  \includegraphics[width=0.5\columnwidth]{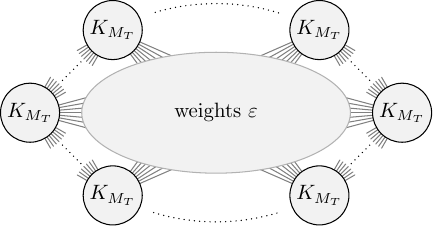}
  \caption{Weights within blocks $K_{M_\rounds}$ have value 1, otherwise $\varepsilon$.
  $K_{M_\rounds}$ is a complete graph on $M_\rounds$ nodes, with $M_\rounds$ being a function of $\rounds.$ 
  }
 \label{fig:lowerbound_graph}
 \end{center}
\end{figure}

\subsection{Algorithms for spectral bandits}

Having expressed the rewards as a linear combination of eigenvectors, 
we can directly modify \LinUCB \parencite{li2010contextual} to use \textit{spectral
penalty}~\eqref{eq:smoothness} for the regularized
least-squares estimate $\widehat\balpha_t$
\[
\widehat\balpha_t = \argmin_{\bw\in\R^\nodes} \left( \sum_{s=1}^{t}\left[\bx_{I_s}\transpose \bw  - r_{I_s}\right]^2 + \|\bw\|_{\bLambda}^2  \right).
\]
This gives us \SpectralUCB with the regret scaling as $\tilde\cO(\effd\sqrt{\rounds})$.
\begin{theorem}[Regret of \SpectralUCB by \textcite{valko2014spectral}]
\label{thm:tucb}
Let $\effd$ be the effective dimension and $\lambda$ be the minimum
eigenvalue of $\bLambda$. If $\| \balpha\|_{\bLambda} \leq C$
and for all $\bx_a$, $ \bx_a\transpose\balpha \in [-1, 1]$,  then the
cumulative regret of \SpectralUCB  is with probability at
least $1-\delta$ bounded as
\begin{align*}
R_\rounds \leq  &\left(4 R\sqrt{ \effd \log(1 + \rounds/\lambda) + 2\log(1/\delta)} + 2C +2\right)  \sqrt{4\effd \rounds \log (1 + \rounds/\lambda)}.
\end{align*}

\end{theorem}
\begin{remark}
The constant $C$ needs to be such that $\| \balpha \|_{\bLambda} \leq C$.
If we set $C$ too small, the true $\balpha$ will lie outside of the region and far from $\widehat\balpha_t$, causing the algorithm to underperform.
Alternatively, $C$ can be time dependent, e.g.,~$C_t = \log \rounds$. In such case, we
do not need to know an upper bound on $\| \balpha \|_{\bLambda}$
in advance, but our regret bound would only hold after some
$t$,  when $C_t \geq \| \balpha \|_{\bLambda}$.
\end{remark}

\noindent
It is known that the available upper bound for \LinUCB \parencite{li2010contextual}, \LinearTS \parencite{agrawal2013thomson} or \OFUL \parencite{abbasi2011improved} is not optimal for
the linear bandit
setting with finite number of arms in terms of dimension $D$.  On the other
hand, the algorithms \SupLinRel or \SupLinUCB
achieve the optimal $\sqrt{D\rounds}$ regret.
In the following, we likewise provide an
algorithm that also scales better with $\effd$
and achieves $\sqrt{\effd\rounds}$ regret.
The algorithm is called \SpectralEliminator  \parencite{valko2014spectral}
and works in phases, eliminating the arms
that are not promising. 
The  phases are defined by the time indexes $t_1=1\leq t_2\leq \dots$
and depend on some parameter $\beta$.
The algorithm is in spirit similar to the \ImprovedUCB by~\textcite{auer2010ucb}.
The main idea
of \SpectralEliminator is to divide the time steps into sets in order
to introduce independence and allow
the Azuma-Hoeffding inequality \parencite{azuma1967weighted} to be applied.
In the following theorem, we characterize the performance
of \SpectralEliminator and show
that the upper bound on regret has $\sqrt{\effd}$ improvement 
over \SpectralUCB.

\begin{theorem}[Regret of \SpectralEliminator by \textcite{valko2014spectral}]
\label{thm:eliminator}
Choose the phase starts as  $t_j=2^{j-1}$. Assume all rewards are in $[0,1]$
and $\|\balpha\|_\bLambda\leq C$. For any $\delta>0$, with probability at least
$1-\delta$, the cumulative regret of \SpectralEliminator algorithm run with
parameter $\beta=2R \sqrt{\!14\log(2K\!(1+\log_2 \rounds)/\delta)} + C$ is bounded as:
\begin{align*}
R_\rounds \leq 2 &+ 16\left(2R \sqrt{14 \log\frac{2K(1+\log_2
\rounds)}{\delta}}+C + \frac{1}{2}\right) 
\times\sqrt{\effd\rounds\log_2(\rounds) \log\left(1 + \rounds/\lambda\right)}
\end{align*}
\end{theorem}

\newpage
\begin{remark}
If we use $\bLambda = \bI$ in \SpectralEliminator, we get a new algorithm,
\LinearEliminator,
which is a competitor to \SupLinRel~\parencite{auer2002using}
or \SupLinUCB~\parencite{chu2011contextual}
and as a corollary to Theorem~\ref{thm:eliminator}
also enjoys $\tilde\cO(\sqrt{D\rounds})$ upper bound on the cumulative regret.
Compared to \SupLinRel or \SupLinUCB, \LinearEliminator and its analysis
are much simpler.
\end{remark}

\subsection{Scalability and computational complexity}

There are three main computational issues to address in
order to make the proposed algorithms scalable:
the computation of $\nodes$ UCBs, matrix inversion, and obtaining the eigenbasis
which serves as an input to the algorithm.
First, to speed up the computation of $\nodes$ UCBs in each time step, we use the lazy
updates technique~\citep{desautels12parallelizing} which maintains a sorted
queue of UCBs and in practice leads to substantial
speed gains. Second, to speed up matrix inversion we do
iterative matrix inversion \citep{zhang2005schur}.

Finally, while the eigendecomposition of a general matrix is 
computationally difficult, Laplacians are symmetric diagonally
dominant (SDD). This enables us to use fast SDD solvers such as \CMG by
\parencite{koutis2011combinatorial}. Furthermore, using \CMG we can find good
approximations to the first
$L$ eigenvectors in $\cO(L m \log m)$ time, where $m$ is the number of edges in
the graph (e.g., $m=10\nodes$ in the Flixster experiment).
\CMG can easily work with $\nodes$ in millions.
In general, we have $L = \nodes$ but from our experience, a smooth reward function
can
be often approximated by dozens of eigenvectors. In fact, $L$ can be considered
as an upper bound on the number of eigenvectors we actually need.
Furthermore, by choosing small $L$ we not only reduce the complexity of
eigendecomposition but also the complexity of the least-square problem being
solved in each iteration.

Choosing a small $L$ can significantly reduce the computation
but it is important to choose $L$ large enough so that still less than
$L$ eigenvectors are enough. This way, the
problem that we solve is still relevant and our analysis applies.
In short, the problem cannot be solved trivially by choosing first $k$
relevant eigenvectors because $k$ is unknown. Therefore, in practice, we choose
the largest $L$ such that our method is able to
run. 

Even with all those improvements, we may have to recompute 
the UCBs for many arms.  As in linear bandits, 
\ThompsonSampling \parencite{thompson1933likelihood} provides 
more computationally efficient alternative 
and we can easily derive a \ThompsonSampling equivalent of \SpectralUCB. 
This variant is called \SpectralTS \parencite{kocak2014spectral} 
and its upperbound also scales as $\tilde\cO(\effd\sqrt{\rounds})$.

\begin{theorem}[Regret of \SpectralTS by \textcite{kocak2014spectral}]
\label{mainTheorem}
Let $\effd$ be the effective dimension and $\lambda$ be the minimum eigenvalue
of $\bLambda$. If $\|\balpha\|_\bLambda\leq C$ and for all $\bx_a$,
$\bx_a\transpose \balpha \in[-1,1]$, then the cumulative regret of \SpectralTS
 is with probability at least $1-\delta$ bounded as
\begin{align*}
R_\rounds\leq\,&\frac{11g}{p}\sqrt{\frac{4+4\lambda}{\lambda}\effd\rounds\log\frac{\lambda+\rounds}{\lambda}}+\frac{1}{\rounds}+ \frac{g}{p}\left(\frac{11}{\sqrt{\lambda}}+2\right)\sqrt{2\rounds\log\frac{2}{\delta}},
\end{align*}
where $p = 1/(4e\sqrt{\pi})$ and
\begin{align*}
g =\, &\sqrt{4\log (\rounds\nodes)}\left(R\sqrt{6\effd\log\left(\frac{\lambda+\rounds}{\delta\lambda}\right)}+C\right)	+R\sqrt{2\effd\log\left(\frac{(\lambda+\rounds)\rounds^2}{\delta\lambda}\right)}+C.
\end{align*}
\end{theorem}
\begin{remark}
Substituting $g$ and $p$, we see that the regret bound scales as
$\effd\sqrt{\rounds\log{\nodes}}$. Note that $\nodes=D$ could be exponential in $\effd$ and we need to
consider factor $\sqrt{\log{\nodes}}$ in our bound. On the other hand, if~$\nodes$ is
indeed exponential in $\effd$, then our algorithm scales with
$\log{D}\sqrt{\rounds\log{D}}=\log(D)^{3/2}\sqrt{\rounds}$ which is even better.
\end{remark}
\begin{remark}
Since \ThompsonSampling is a Bayesian approach, it requires a prior to run
and we choose it here to be a Gaussian. However, this does not pose any
assumption whatsoever about the actual data
both for the algorithm and the analysis.
The only assumptions we make about the data are: (a) that the mean payoff is
linear in the features, (b) that the noise is sub-Gaussian, and (c) that we know
a bound on the Laplacian norm of the mean reward function.
We provide a frequentist bound on the regret (and not an
average over the prior) which is a much stronger worst case result.
\end{remark}
\section{Related approaches to smoothness on graphs}
In this section, we review other graph bandit approaches that assume smoothness but 
either have a different objective or they assume smoothness in some other form. 

\subsection{Spectral bandits with different objectives}
In the follow-up work on spectral bandits, there have been algorithms
optimizing other objective function than the cumulative regret.
First, in some sensor networks, sensing a node (pulling and arm)
has an associated cost \parencite{narang2013signal}. In a 
particular, \emph{cheap bandit} setting \parencite{hanawal2015cheap}, 
it is cheaper to  get an average of rewards of a set of nodes
than a specific reward of a single one. For this setting, we proposed 
\CheapUCB  \parencite{hanawal2015cheap} that reduces the cost of sampling by 1/4
as compared to \SpectralUCB, while maintaining 
$\tilde\cO(\effd\sqrt{\rounds})$ cumulative regret.
Next, \citet{gu2014online} study the online classification setting 
on graphs with bandit feedback, very similar to spectral bandits.
The analysis of their algorithm delivers essentially the same bound
on the regret, however, they need to know the number of relevant eigenvectors~$\effd$.
Moreover, \textcite{ma2015active} consider several variants of \emph{$\Sigma$-optimality}
that favors specific exploration when selecting the nodes.

\subsection{Smoothness of linear parameter vectors}
Spectral bandit strategies are relevant to recommender systems but only consider a single user. 
However, the information sharing is possible and desirable also between the users. This is considered 
in \emph{gang of bandits} \parencite{cesa-bianchi2013gang}, a graph bandit setting 
where each node represents user $i$, and is a linear bandit itself with parameter $\bw_i$, 
unknown to the learner.
Each round, the learner gets a user index $i_t$ (node) with a set of contexts 
$C_t$ and has to chose a $\bar\bx_t \in C_t$. The graph, in this case, 
represents a network of users and it is the parameters $\{\bw_i\}_i$ that
are assumed to be smooth on the given graph in a Laplacian way.
The \GOBLin algorithm \parencite{cesa-bianchi2013gang} exploits 
this smoothness and after each feedback and the local update of $\widehat\bw_i$, 
also brings $\{\widehat \bw_i\}_i$ closer together.

\begin{theorem} [Regret of \GOBLin by \textcite{cesa-bianchi2013gang}]
With probability $1-\delta$, the cumulative regret of \GOBLin is upperbounded as 
\[ 
R_\rounds \leq 4 \sqrt{\rounds \left( R \ln \frac{m_\rounds}{\delta} + L\left(\bu_1, \dots, \bu_\nodes\right) \right)  \ln \abs{m_\rounds} }. 
\]
where $R$ is the sub-Gaussianity of the noise, $m_\rounds$ is a $(D\nodes) \times (D\nodes)$  block-diagonal matrix with block being the covariance matrices of each node (linear bandit) and
\[
L\left(\bu_1, \dots, \bu_\nodes\right) = \sum_{i \in \nodeset} \norm{\bu_i}^2 + \sum_{(i,j) \in \edgeset} \norm{\bu_i - \bu_j}^2,
\]
\end{theorem}
The term $\sum_{(i,j) \in \edgeset} \norm{\bu_i - \bu_j}^2$ in the bound above reflects the smoothness of the reward
vectors among the nodes (user) and can be thought of as the vector version of the smoothness constant 
$C$ in spectral bandit bound (Theorem~\ref{thm:tucb}). The value 
of $\ln \abs{m_\rounds}$  can be of order $\tilde\cO\left(D\nodes\right)$.

\subsection{Clusters of linear bandits, unimodal bandits, and reward from multiple nodes}
A slightly stronger assumption  is considered by
\textcite{gentile2014online}, where
the nodes of the graph (users) can be clustered with respect to some unknown underlying clustering
and the nodes within a cluster exhibit similar behavior. 
The regret bound of their \CLUB algorithm scales roughly with the 
number of clusters instead of the number of nodes, but can be even better 
if there are big clusters with identical arms.
\textcite{li2015online} later extended the approach to \emph{double clustering} where both the users and the items
are assumed to appear in clusters (with the underlying clustering unknown to the learner)
and \textcite{korda2016distributed} consider a distributed extension. 

Yet another assumption of a special graph reward structure is exploited by unimodal bandits~\citep{yu2011unimodal,combes2014unimodal}.
One of the settings considered by \citet{yu2011unimodal} is a graph bandit setting where every path in the graph has unimodal rewards and
therefore also imposes a specific  kind of smoothness with respect to the graph topology.

In networked bandits~\citep{fang2014networked}, the learner picks a node, but besides receiving the reward from that node, its reward is the sum of the rewards
of the picked node and its neighborhood. The algorithm of \textcite{fang2014networked}, \NetBandits, can also deal with changing topology, however, this has to be always 
revealed to the learner before it makes its decision.

\section{Perspectives for graph-smooth rewards}
We outline some future extensions of bandit learning on graphs with smooth rewards.

\subsection{Improvements for the effective dimension for spectral bandits}
While the effective dimension is related to the number of relevant eigenvectors,
its precise definition (Definition~\ref{def:effectived}) comes from the 
analysis of the regularized covariance matrix used in  least-squares regression \parencite[][Lemma~6]{valko2014spectral}. 
One possible improvement is to define the effective dimension as an earlier upper bound in the analysis, in particular, define it as
\[
\dnew = \frac{\max\log\prod_{i=1}^\nodes\left(1+\frac{t_i}{\lambda_i}\right)}{\log\left(1+\frac{\rounds}{\lambda}\right)}
\]
where the $\max$ is taken over all possible non-negative integers $\{t_1,\,\dots,\,t_\nodes\}$, such that $\sum_{i = 1}^\nodes t_i=\rounds.$
This improves the scaling of the regret bound with respect to $\effd$ (Figure~\ref{fig:effd_real}) and since we use $\effd$
in the algorithm to avoid computation of the determinants, this new definition has also a practical impact.
\begin{figure}[ht]
 \begin{center}
   \vspace{1em}
  \includegraphics[width=0.32\columnwidth]{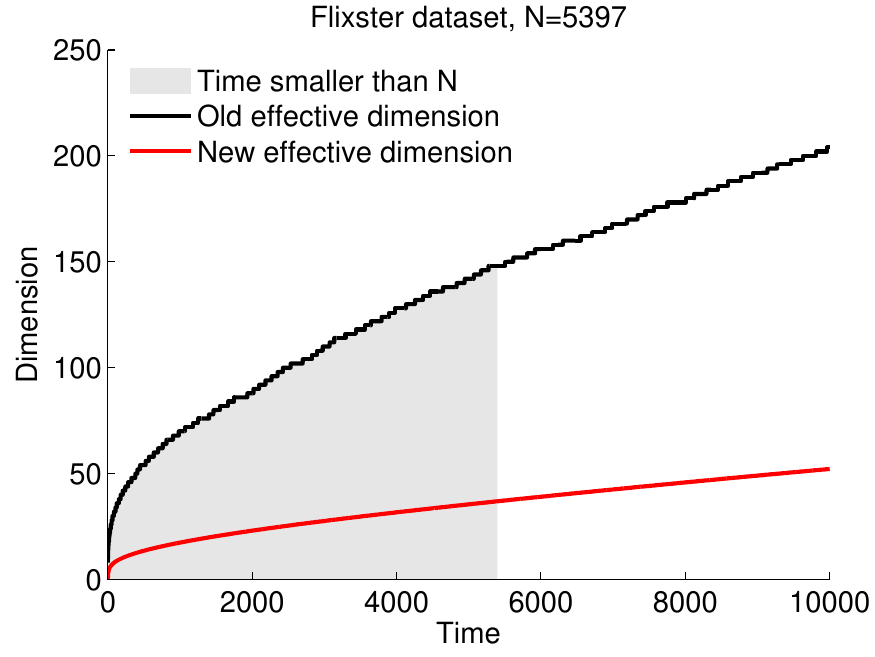}
  \includegraphics[width=0.32\columnwidth]{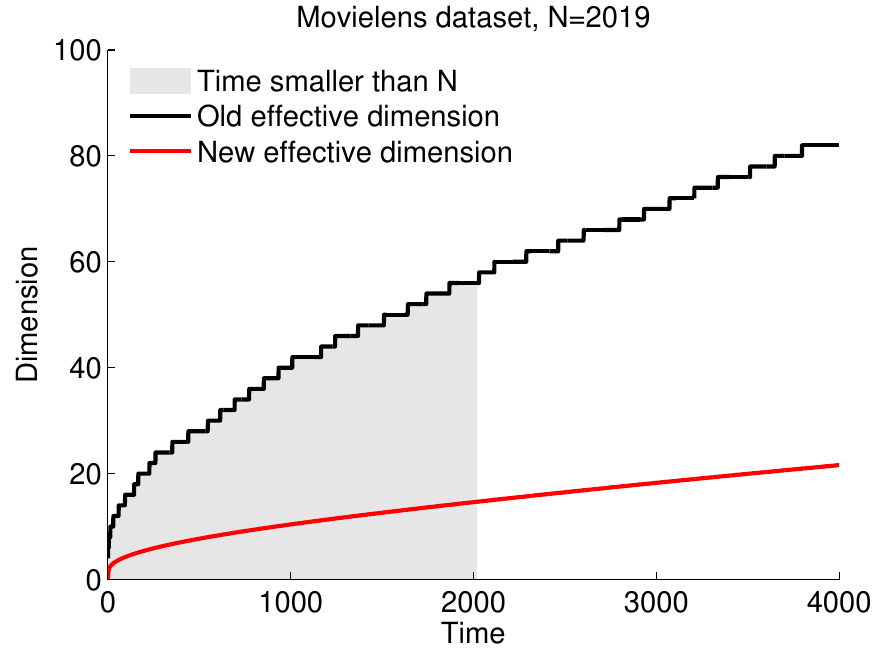}
  \includegraphics[width=0.32\columnwidth]{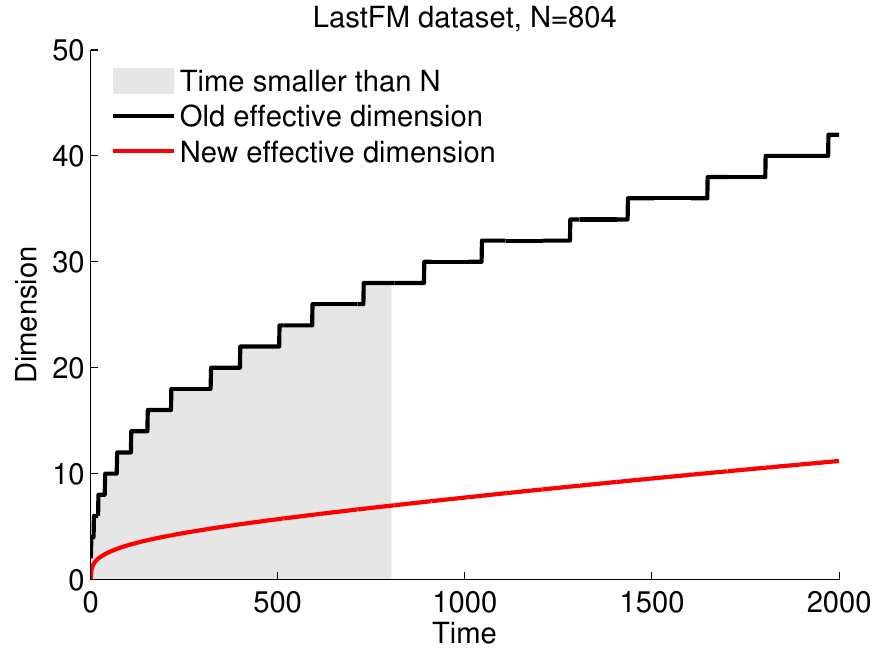}
  \vspace{1em}
  \caption{Difference between $\dnew$ and $2\effd$ for real world datasets. From left to right: Flixster dataset with $\nodes=5397$, MovieLens dataset with $\nodes=2019$, and LastFM dataset with $\nodes=804$.}
 \label{fig:effd_real}
 \end{center}
\end{figure}

\noindent Notice that $\effd$ does not depend on the reward and while we have $\Omega(\sqrt{\effd\rounds})$ lower bound in $\effd$, it is rather both effective dimension $\effd$ and the upper bound 
on the reward smoothness $C$ that together reflect the difficulty of spectral bandits. 
If the rewards are not smooth ($C$ is large), then the problem is as difficult as learning $\nodes$-arm bandits, no matter how small $\effd$ 
is.\footnote{This does not violate our upper bounds that obviously depend on $C$.}
Therefore, an interesting and more fundamental open question is the better understanding of the problem difficulty of spectral bandits with perhaps a \emph{single} 
measure of difficulty.

\subsection{Applicability to recommender systems}
The oracle strategy for spectral bandits would always pick the most rewarding node. This is not desirable in many online 
recommender systems as this would mean watching the same movie or listening to the same song all the time. 
This is however not limiting, because, similar to linear bandits, the weight vector is the only thing that we are learning. 
This means that we can  restrict the set of available arms to the ones that were not pulled yet and 
\SpectralUCB or \SpectralTS and their analyses extend to that situation.

On the other hand, unlike in linear bandits, changing the arm set in general (e.g., adding new arms) would require changes in the approach. 
The reason is that the spectral basis is assumed to be fixed (as is the standard basis for linear bandits) and is $\nodes$-dimensional (there are $\nodes$
eigenvectors for~$\nodes$ nodes). Adding a new arm (node) would often require the update of the basis and projecting the current estimate of the
weight vector to this new basis.\footnote{Note that elimination algorithms, such as \SpectralEliminator achieving $\tilde\cO(\sqrt{\effd\rounds})$ regret, 
do not extend easily to the changing sets of arms.}

One aspect of spectral bandits is that it replaces the costly \emph{feature-engineering} step needed for 
linear bandits with a pairwise similarity and this way circumvents feature selection. 
On the other hand, this approach is limited to a single user recommendation, 
unlike gang of bandits~\parencite{cesa-bianchi2013gang}, which however uses a linear bandit
for each user needing feature construction.  A useful future work would be the exploitation 
of smoothness in \emph{both} the item and the user space.

A standard approach to smoothness or similarity of the rewards in the recommender systems
is based on \emph{low-rank matrix factorization} of the user-item matrix.
Although there are already first results, studying this approach in bandit setting
both for matrix factorization \parencite{mary2015bandits,guillou2015collaborative,guillou2016scalable} and
probabilistic matrix factorization  \parencite{prisadnikov2014exploration,tu2015bandit,kawale2015efficient}
based either on \UCB or \ThompsonSampling, they are mostly empirical.
One of the difficulties is the nonconvexity of the non-negative matrix factorization
and another one is the interplay between the rows and columns in the user-item matrix. 
Besides a more theoretical understanding of this approach, it would be interesting to relate it 
to spectral bandits and understand the tradeoffs coming from spectral smoothness vs.\,low-rank assumptions.

\chapter{Side observations}\label{chap:side}

\begin{figure}[t]
\begin{minipage}[b]{0.45\linewidth}
\begin{center}
\includegraphics[width = 2.2em]{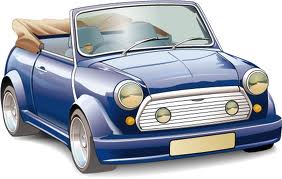}
\ 
\includegraphics[width = 2.2em]{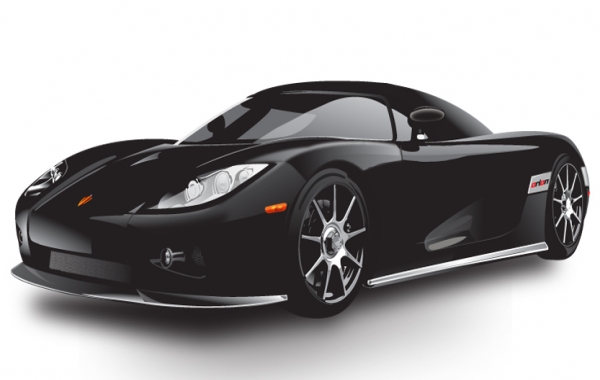}
\ 
\includegraphics[width = 2.2em]{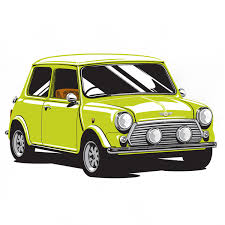}
\ 
\includegraphics[width = 2.2em]{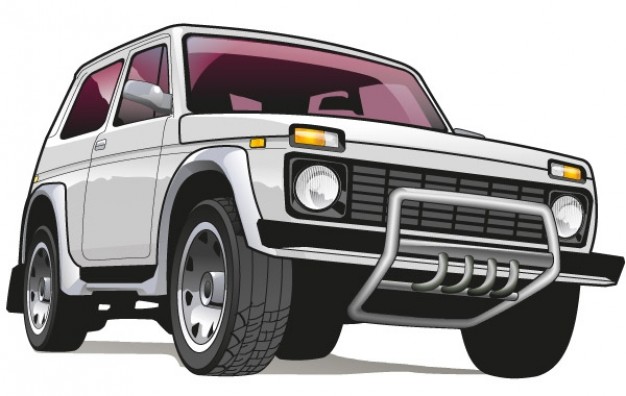}
\ 
\includegraphics[width = 2.2em]{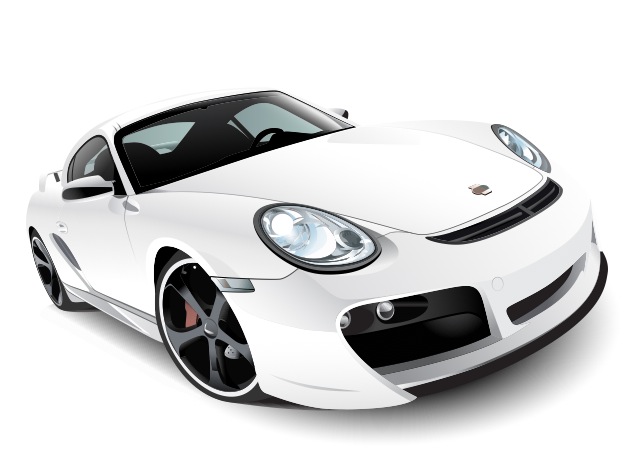}
\includegraphics[width = 2.2em]{normal_car1.jpg}
\begin{center}
\begin{tikzpicture}
\vspace{-1em}
    \draw [->,line width =	 2.2pt] (0.2,0) -- (1.2,1.5);
    \draw [->,line width =	 2.2pt] (-0.2,0) -- (-1.2,1.5);
\end{tikzpicture}
\end{center}
\begin{center}
\vspace{-1em}
\includegraphics[width = 4em]{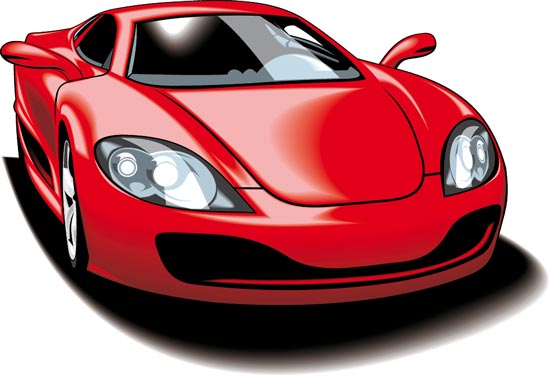}
\end{center}
\end{center}
\end{minipage}
\qquad
\begin{minipage}[b]{0.45\linewidth}
\includegraphics[width=13em]{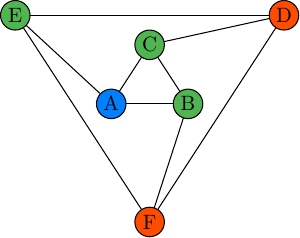}
\end{minipage}
\caption{Side observations on \textbf{undirected} graphs. \textbf{Left:} \emph{Recommendation example}: When a provider
sees the interest in a particular sport car, they can assume the interest in other sport cars.
 \textbf{Right:} \emph{Illustration of the feedback:}
Whenever the learner asks for node $A$, it receives the feedback also for the nodes $B$, $C$, and $E$, but not for the rest.}
\label{fig:graphsideundirected}
\end{figure}

\begin{figure}[t]
\begin{minipage}[b]{0.45\linewidth}
\begin{center}
\includegraphics[width = 2.2em]{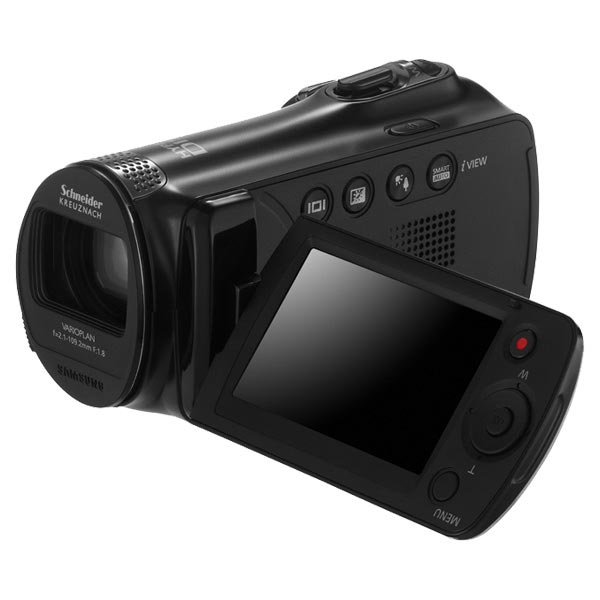}
\ \ \ 
\includegraphics[width = 2.2em]{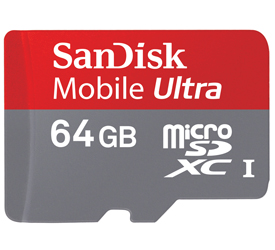}
\ \ \ 
\includegraphics[width = 2.45em]{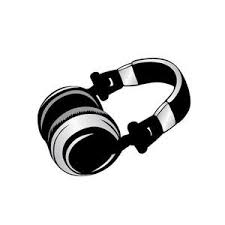}
\ \ \ 
\includegraphics[width = 1.2em]{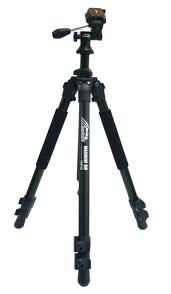}
\ \ \ 
\includegraphics[width = 2.2em]{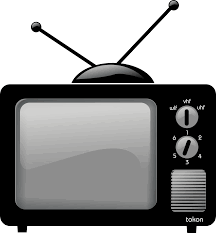}
\begin{center}
\begin{tikzpicture}
\vspace{-1em}
    \draw [->,line width =	 2.2pt] (0.2,0) -- (1.2,1.5);
    \draw [->,line width =	 2.2pt] (-0.2,0) -- (-1.2,1.5);
\end{tikzpicture}
\end{center}
\begin{center}
\vspace{-1em}
\includegraphics[width = 4em]{camcorder.jpg}
\end{center}
\end{center}
\end{minipage}
\qquad
\begin{minipage}[b]{0.45\linewidth}
\includegraphics[width=14em]{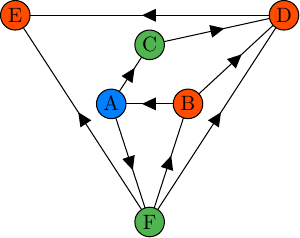}
\end{minipage}
\caption{Side observations on \textbf{directed} graphs. \textbf{Left:}\@ \emph{Recommendation example}: When a provider
sees the interest in video cameras, they can assume the interest in SD cards and tripods, but not necessarily vice versa.
 \textbf{Right:}\@ \emph{Illustration of the feedback:}
Whenever the learner asks for node $A$, it receives the feedback also for the nodes $C$ and $F$, but not for the rest.}
\label{fig:graphsidedirected}
\end{figure}
How to take advantage of richer feedback  than a bandit one? In some situations, we can freely access 
or infer feedback for the actions that the online learner did not take. In \emph{recommender systems}, this can 
be inferring interest about similar items (Figure~\ref{fig:graphsideundirected}, left) or  
accessories (Figure~\ref{fig:graphsidedirected}, left).  

Another motivation is an online interaction in \emph{sensor networks}, with sensors distributed in the area and
where each sensor collects some information about the environment and can communicate with nearby sensors
to share this information. Therefore, when the learner asks for data from a particular sensor, it can access the information from other, geographically close sensors.

If we equate the actions (sensor, choices, arms) with the nodes of a graph and the side information with the
edges (Figure~\ref{fig:graphsideundirected}, right; and Figure~\ref{fig:graphsidedirected}, right)
then we can see this setting as graph bandits with side observations.
This setting was formally defined by \textcite{mannor2011from} 
as an intermediate feedback protocol between bandit feedback and full information (learning with experts).  
The graph in this setting represents the \emph{observation system} of side observations.

\begin{figure}[H]
\centering
\fbox{
\begin{minipage}{.95\textwidth}
{\bfseries Parameters}: \\
\phantom{eye}set of arms $[\nodes]$, number of rounds $\rounds$.\\
{\bfseries For all $t=1,2,\dots,\rounds$ repeat}
\begin{enumerate}
 \item The environment picks a loss function $\loss_t:[\nodes]\ra [0,1]$ and a directed graph $\cG_t$ with 
edge weights in $[0,1]$.
 \item Based on its previous observations (and possibly some source of randomness), the learner picks an action
$I_t\in[\nodes]$.
 \item The learner suffers loss $\loss_{t,I_t}$.
 \item The learner observes  $\cG_t$ and the feedback 
 \[
\loss_{t,j} \text{ for all } j \text{ for all } (I_t \ra j)\in \cG_t
 \] 
\end{enumerate}
\end{minipage}
}
\caption{The protocol of bandit learning with side observations.}
\label{fig:protocolside}
\end{figure}
\noindent Figure~\ref{fig:protocolside} shows the learning protocol that we consider in this chapter. The different 
approaches vary depending whether the losses are stochastic or adversarial, whether the graphs are fixed 
or can change, whether they need to be revealed to the learner before it chooses the action or only after,
and whether the graphs are directed or undirected.
Table~\ref{tab:sidealgos} lists the algorithms for the adversarial case with some of their
properties that we later discuss in detail.
\begin{table}[htp]
\begin{center}
\begin{tabular}{|c|c|c|c|}
\hline
Algorithm & Reference & orientation  & graph $\cG_t$ \\
\hline
\ELP & \cite{mannor2011from} & (un)directed  & known before \\
\expset & \cite{alon2013from} & undirected  & only after  \\
\expdom & \cite{alon2013from} & (un)directed  & known before \\
\expix, \fplix & \cite{kocak2014efficient} & (un)directed  & only after \\ 
\expg & \cite{alon2015online} & (un)directed & only after \\
\hline
\end{tabular}
\end{center}
\caption{Graph bandit algorithms for learning with side observations with nonstochastic losses}
\label{tab:sidealgos}
\end{table}%
Most of the algorithms for the setting are  graph variants of \EXP (template shown in Algorithm~\ref{alg:expgraph})
and vary by how they define their node sampling distribution (Line~\ref{expgrap:line:prob}, Algorithm~\ref{alg:expgraph})
and how they construct their loss estimates (Line~\ref{expgrap:line:lossestimaes}, Algorithm~\ref{alg:expgraph}).
We will discuss these choices in the rest of the chapter. Before that, we
define two graph-dependent quantities that will be used to state the regret bounds.

 \begin{algorithm}[H]
\caption{\EXP template for graph bandits with side observations}
\label{alg:expgraph}
\begin{algorithmic}[1]
 \STATE \textbf{Input:} Set of actions $\Sw = [\nodes]$,
 parameters
$\gammat\in(0,1)$, $\eta_t>0$ for $t\in[\rounds]$.
\FOR{$t = 1$ {\bfseries to} $\rounds$}
\STATE  $w\ti \gets (1/\nodes)\exp{(-\etat\hL_{t-1,i})}$ for $i\in
[\nodes]$
 \STATE An adversary privately chooses losses $\loss\ti$ for $i\in[\nodes]$
and generates a graph $\cG_t$
\STATE \textcolor{cyan}{Necessary for some algorithms:}  Observe graph $\cG_t$  \label{expgrap:line:observe}
\STATE $W_t \gets \sum_{i=1}^\nodes w_{t,i}$
\STATE \textcolor{ocre}{Define probabilities:}  $p_{t,i}$, default:  $p_{t,i} =  \frac{w_{t,i}}{W_t}$ \label{expgrap:line:prob} 
\STATE Choose $I_t  \sim \bp_t=
(p_{t,1},\dots,p_{t,\nodes})$
\STATE Observe graph $\cG_t$
\STATE Observe pairs $\{i,\loss_{t,i}\}$ for $(I_t \ra
i)\in \cG_t$
\STATE  $\oti \gets \sum_{(j \ra i)\in
\cG_t}\ptj$ for  $i\in [\nodes]$
\STATE \textcolor{ocre}{ Define loss estimates:} $\hloss\ti$, default: $\hloss\ti \gets
\frac{\loss_{t,i}}{\oti}\I_{\{(I_t \ra i)\in \cG_t\}}$ for $i\in  [\nodes]$  \label{expgrap:line:lossestimaes}
 \STATE  $\hLoss\ti \gets \hLoss_{t-1,i}+ \hloss\ti$ for  $i\in [\nodes]$
\ENDFOR
\end{algorithmic}
\end{algorithm}

\begin{definition}
The \textbf{independence set} of graph  $\cG_t$ is a set of nodes, for which no pair is adjacent. The maximum possible size of 
such set is called \textbf{independence number} $\indn_t$.
\end{definition}

\begin{figure}[H]
 \begin{center}
  \includegraphics[width = 0.5\textwidth]{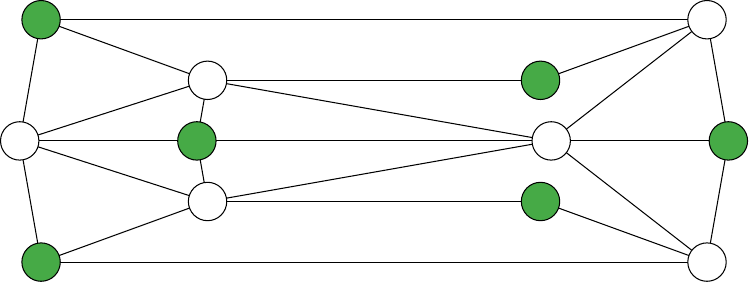}
  \caption{\emph{Example:} Independence set of size 6.}
 \label{fig:independenceset}
 \end{center}
\end{figure}

\begin{definition}
The \textbf{clique-partition number} $\cliquen_t$ of  graph $\cG_t$ is the smallest number of cliques that partition all the nodes.
\end{definition}

\begin{remark}
If $\indn_t$ and $\cliquen_t$ are the independence and the clique-partition  numbers of the same graph, then 
 $\indn_t \leq \cliquen_t$, since any clique can have at most one node from the independence set.

\end{remark}



\section{Undirected side observations}\label{sec:undirside} 
In this section, we consider undirected observations from Figure~\ref{fig:graphsideundirected}, 
which means that in every round,  graph $\cG_t$ is undirected (symmetric).
The first algorithm, \ELP \parencite{mannor2011from}, 
uses unbiased loss estimates (default setting in Line~\ref{expgrap:line:lossestimaes}, Algorithm~\ref{alg:expgraph}). Moreover, \ELP's sampling probability distribution over the nodes  (Line~\ref{expgrap:line:prob}, Algorithm~\ref{alg:expgraph}) is 
\[ 
 p_{t,i} =  \pa{1-\gamma_t}  \frac{w_{t,i}}{\sum_{j=1}^\nodes w_{t,j}} + \gamma_ts_{t,i},  \text{ where }
\ac{s_{t,i}}_{i \in \ac{\nodes}} = \argmax_{\forall i \ s_{t,i}  \ge 0, \sum_i s_{t,i} = 1 } \min_{j \in \sqpa{\nodes}} \sum_{(j \ra l)\in \cG_t} s_{t,l}, 
\]
where $\ac{s_{t,i}}_{i \in \ac{\nodes}}$ can be found using linear programming (\LP), given the graph is revealed by the environment 
in Line~\ref{expgrap:line:observe} (Algorithm~\ref{alg:expgraph}).  Intuitively, \ELP \emph{mixes} in a distribution that is not uniform
but is aware of the observation system (graph) and thus aims at distributing the exploration equally.
Furthermore, \ELP uses gains instead of losses. 

\begin{theorem}[Regret of \ELP by \textcite{mannor2011from}]\label{thm:elpregret}
Setting the learning rate $\eta_t = \sqrt{\pa{\log \nodes}/(3\sumt{\indn_t})}$
and mixing rate $\gamma_t = \eta_t/\pa{\min_{j \in \sqpa{\nodes}} \sum_{(j \ra l)\in \cG_t} s_{t,l}}$, the expected 
regret of \ELP is upper bounded as
\[R_\rounds \leq \sqrt{3\pa{\textstyle\sumt{\indn_t}}\log \nodes}. \]
\end{theorem}
\ELP needs to see $\cG_t$ revealed before the action is taken, to run the \LP to tune its learning rate.
This was fixed later by \expset \citep{alon2013from}, which does not need either. \expset uses losses instead of rewards, 
and differs from \EXP only by the loss estimates (same as for \ELP). For the sampling distribution, it uses simple \EXP weighting without mixing, 
\[ 
 p_{t,i} =  \frac{w_{t,i}}{\sum_{j=1}^\nodes w_{t,j}}, 
\]
and thus does not need to know the graph in Line~\ref{expgrap:line:observe} (Algorithm~\ref{alg:expgraph}) to provide essentially the same guarantees on the regret.
\begin{theorem}[Regret of \expset by \textcite{alon2013from}]
Setting the learning rate $\eta_t = \sqrt{\pa{2 \log \nodes}/(\sumt{\indn_t}})$, the expected 
regret of \expset is upper bounded as
\[R_t \leq \sqrt{2\pa{\textstyle\sumt{\indn_t}}\log \nodes}. \]
\end{theorem}

\paragraph{Knowledge of $\cG_t$}
Note that both \ELP and \expset, set their learning parameter $\eta_t$ as a function of  $\sumt{\indn_t}$. This can be 
however avoided for both, by running an additional \EXP algorithm on top it, at the price of an additional $\log$ factor \parencite{mannor2011from}.
Therefore, \expset can avoid any knowledge of~$\cG_t$ before it picks a node. 
However, note that both of them need some knowledge of the graph \emph{after} the node $I_t$ is picked.
In particular, they need it to construct the loss estimates in Line~\ref{expgrap:line:lossestimaes} of Algorithm~\ref{alg:expgraph}.
Since the algorithms update not only the loss estimate of $I_t$-th node but also of its neighbors in $\cG_t$, the algorithms
require also the knowledge of the neighbors of neighbors of~$I_t$, the \emph{second neighborhood} of $I_t$.
This is shared by many algorithms in this chapter. While the knowledge of the first neighborhood is a very reasonable 
assumption (we know from which nodes the observations came from), the knowledge of the second neighborhood
may not be always available in practical deployments.

\section{Directed side observations} 
We now turn our attention to directed graphs from Figure~\ref{fig:graphsidedirected}.
\ELP from Section~\ref{sec:undirside} can be used without modification, but the upper bound given by \textcite{mannor2011from}
only gives the version of Theorem~\ref{thm:elpregret}, with clique-partition number  $\cliquen_t$ instead of independence number $\indn_t$.
\expset can be also used, however \textcite{alon2013from}, show the graph and the distribution of the sampling probabilities
for which the key quantity (coming from Line~\ref{expgrap:line:lossestimaes}, Algorithm~\ref{alg:expgraph}) cannot be upper bounded
by the independence number $\indn$.
Therefore \textcite{alon2013from} designed  \expdom whose guarantees were proved to be functions of the \emph{independence number}, 
which gives either equal or a better guarantee on the regret.
To control the problematic quantity (discussed later), \expdom controls the loss estimates $\hloss_{t,i}$  by mixing in a uniform distribution in Line~\ref{expgrap:line:prob} of Algorithm~\ref{alg:expgraph},
supported on the dominating set of the directed graph $\cG_t$ (set of the nodes that have the directed edges to the rest of the graph). 
This achieves the desired bound  but comes with a few disadvantages. First, $\cG_t$ has to be revealed to the learner
at the beginning of each round and so we get the same limitation as for \ELP. Second, 
depending on the size of the dominating set, \expdom needs to run $\log \nodes$ instances to properly set the node sampling distribution.
Finally, since the rounds where to use each instance are random, \expdom needs to use the doubling trick to optimally set~$\gamma_t$ and $\eta_t$.

\subsection{Implicit exploration and \expix} 
\expdom of  \textcite{alon2013from} needed to know $\cG_t$ before choosing the action, to control the loss estimates.
In this section we show how to achieve a similar behavior \emph{without} the knowledge of~$\cG_t$ \parencite{kocak2014efficient}.
In particular, we propose the simplest exploration scheme imaginable, which consists of \emph{merely pretending to
explore}. Precisely, we simply sample our action~$I_t$ from the distribution defined as the default setting
without explicitly mixing with any exploration distribution.
Let $\F_{t-1} = \sigma(I_{t-1},\dots,I_1)$ capture the interaction 
history up to time $t$.
 Our key trick is to define the loss estimates for
all arms $i$ as
\[
 \hloss_{t,i} = \frac{\loss_{t,i}}{o_{t,i} + \gamma_t} \II{(I_t \ra i)\in \cG_t},
 \quad \mbox{where} \quad o_{t,i} = \EEc{O_{t,i}}{\F_{t-1}} \eqdef \PPc{(I_t \ra 
i)\in \cG_t}{\F_{t-1}}
\]
and $\gamma_t>0$ is a parameter of our algorithm.
It is easy to check that $\hloss_{t,i}$ is a \emph{biased} estimate of $\loss_{t,i}$. The nature of this bias, however,
is very special.
 First, observe that $\hloss_{t,i}$ is an \emph{optimistic} estimate of~$\loss_{t,i}$ in the sense that
$
\EEc{\hloss_{t,i}}{\F_{t-1}}\le \loss_{t,i}$. That is, our bias always ensures 
that,
 on expectation,
we underestimate the loss of any fixed arm
$i$. Even more importantly, our loss estimates also satisfy
\begin{equation}
\begin{split}\label{eq:lbias}
\EEcc{\sum_{i=1}^\nodes p\ti\hloss\ti}{\F_{t-1}} & = \sum_{i=1}^\nodes \pti\loss\ti +
\sum_{i=1}^\nodes \pti\loss\ti\left(\frac{\oti}{\oti+\gammat}-1\right)
\\
&= \sum_{i=1}^\nodes \pti\loss\ti - \gamma_t \sum_{i=1}^\nodes \frac{\pti\loss\ti}{\oti+\gammat},
\end{split}
\end{equation}
that is, the bias of the estimated losses \emph{suffered by our algorithm} is directly controlled by $\gamma_t$. As we
will see in the analysis, it is sufficient to control the bias of our own estimated performance as long as we can
guarantee that the loss estimates associated with any fixed arm are optimistic---which is precisely what we have.
Note that this slight modification ensures that the denominator of $\hloss_{t,i}$ is lower bounded by $p_{t,i} +
\gamma_t$, which is a very similar property as the one achieved by the exploration scheme used by \expdom.
We call the above loss estimation method \emph{implicit exploration} or IX, as it gives rise to the same effect as
explicit exploration without actually having to implement any exploration policy. In fact, explicit and implicit
explorations can both be regarded as two different approaches for 
bias-variance tradeoff: while explicit
exploration biases the \emph{sampling distribution} of $I_t$ to reduce the variance of the loss estimates, implicit
exploration achieves the same result by biasing \emph{the loss estimates themselves}.

From this point on, we take a somewhat more predictable course and define our algorithm \expix as a variant of \EXP
using the IX loss estimates. One of the twists is that \expix is actually
based on the adaptive-learning-rate variant of \EXP 
by \citet{auer2002nonstochastic}, which avoids the necessity of prior knowledge of the 
observability graphs in order to set a proper learning rate.
This algorithm is 
defined by setting $\hL_{t-1,i} =
\sum_{s=1}^{t-1} \hloss_{s,i}$ and for all $i\in[\nodes]$ computing the weights as
 \[
 w_{t,i} = (1/\nodes)e^{-\eta_t \hL_{t-1,i}}.
\]
 These weights are then used to construct the sampling distribution of $I_t$ as
defined in
 Line~\ref{expgrap:line:prob} of Algorithm~\ref{alg:expgraph}. 
As a result \expix does not even need to know the number of rounds $\rounds$ and our regret bound
scales with the \emph{average} independence number $\bar{\indn}$ of the graphs played by the adversary rather than the
largest of these numbers. \expix employs adaptive learning rate
and unlike \expdom, it does not need to use a doubling trick to be anytime or
to aggregate outputs of multiple algorithms to optimally set their learning
rates. The upper bound on the regret is stated below.

\begin{theorem}[Regret  of \expix by \textcite{kocak2014efficient}]\label{cor:mainExp}
The regret  of \expix satisfies
\[\regret \leq 4\sqrt{\pa{\nodes + 2\textstyle\sumt\left(H_t \indn_t+1\right)}\log \nodes},\]
where
\[
 H_t = \log\left(1+\frac{\lceil \nodes^2\sqrt{t\nodes/\log \nodes}\rceil+\nodes}{\indn_t}\right) =\OO(\log (\nodes \rounds)).
\]
\end{theorem}

\subsection{\expg}\label{ss:expg}
Learning on graphs with directed side observations is the special setting of a more general graph 
feedback related to partial monitoring, for which \textcite{alon2015online} proposed the \expg algorithm.
\expg also follows the template of Algorithm~\ref{alg:expgraph} and mixes in a uniform distribution 
over the nodes for the sampling distribution,
\[ 
 p_{t,i} =  \pa{1-\gamma_t}  \frac{w_{t,i}}{\sum_{j=1}^\nodes w_{t,j}} + \frac{\gamma_t}{\nodes},
\]
which means that it does not need to know the graph $\cG_t$ for this step. The analysis of 
\expg differs from the typical analysis of \EXP-style algorithms by using an improved second-order regret
bound that considers separately small and large losses for a better control of variance.
\expg with proper tuning also achieves $\OO\left(\sqrt{\indn \rounds \log\left(\nodes \rounds\right)}\right)$ regret bound 
and can be generalized to the case when the graph is changing and when $\indn$ (used for parameter tuning)
is unknown using either the doubling trick or an adaptive learning rate employed by \expix. 
\expg and its analysis can be however used in the more general feedback settings discussed in Section~\ref{ss:beyondbandits}.

\subsection{Combinatorial semi-bandit problems with side observations}

We now turn our attention to the setting of online combinatorial optimization (see
\cite{koolen10comphedge,cesa-bianchi2012combinatorial,audibert2014regret}). In this variant of the 
online learning problem, the learner has access
to a possibly huge action set $\Sw\subseteq\ev{0,1}^\nodes$ where each
action is represented by a binary vector $\bv$ of dimensionality $\nodes$. In 
what follows, we assume that $\onenorm{\bv}\le m$
holds for all $\bv\in\Sw$ and some $1\le m\ll \nodes$, with the case $m=1$ 
corresponding to the multi-armed bandit setting considered in the
previous section. 
In each round $t=1,2,\dots,\rounds$ of the decision process, the learner picks an action $\bV_t\in\Sw$ and incurs a loss of
$\bV_t\transpose\bloss_t$. 
At the end of the
round,
the learner receives some feedback based on its decision $\bV_t$ and the loss vector $\bloss_t$. 
The regret of the learner
is defined as
\[
 R_\rounds = \max_{\bv\in\Sw} \EE{\sum_{t=1}^\rounds \pa{\bV_t - \bv}\transpose\bloss_t}.
\]
In this section, we
define a new
feedback scheme
situated between the semi-bandit and the full-information schemes. In particular, we assume that the learner gets
to observe the losses of some other components not included in its own decision
vector $\bV_t$. Similarly to the model of~\citet{alon2013from}, the relation
between the chosen action and the side observations are given by a directed 
observability  $\cG_t$. We refer to this feedback scheme as \emph{semi-bandit
with side observations}.
As an example, consider the situation shown on
Figure~\ref{fig:newsFeedsA}. 
In this simple example, we want to suggest one out of three news feeds to each user, that is, we want to choose a
matching on the graph shown on Figure~1a which covers the users. Assume that
news feeds~2 and~3 refer
to the same content, so \emph{whenever we assign news feed~2 or~3 to any of 
the users, we learn the value of both of
these assignments}. The relations between these assignments can be described by a graph structure
(shown on Figure~\ref{fig:newsFeedsB}), where nodes represent user-news feed assignments, and
edges mean that the
corresponding assignments reveal the clickthroughs of each other. For a more compact representation, we can group the
nodes by the users, and rephrase our task as having to choose one node from each group. Besides its own
reward, each selected node reveals the rewards assigned to all their neighbors.

\begin{figure}[H]
\centering
\begin{subfigure}[t]{0.60\textwidth}
\centering
\includegraphics[height = 3cm]{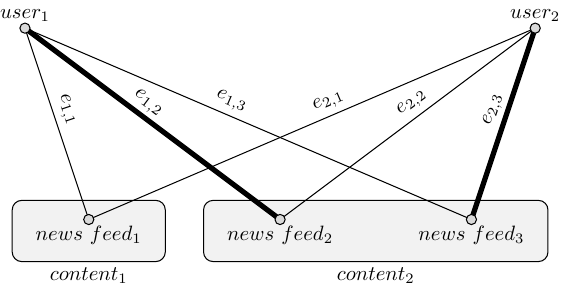}
\caption{The thick edges represent one
potential matching of users to feeds, grouped news feeds show the same content.}
\label{fig:newsFeedsA}
\end{subfigure}
\quad
\begin{subfigure}[t]{0.295\textwidth}
\centering
\includegraphics[height = 3cm]{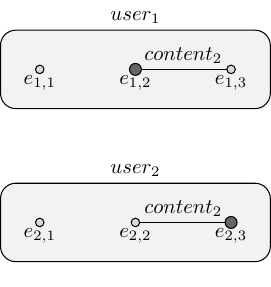}
\caption{Connected feeds mutually reveal each
others clickthroughs.}
\label{fig:newsFeedsB}
\end{subfigure}
\caption{Users and news-feeds example of complex actions with side observations.}
\label{fig:newsFeeds}
\end{figure}
\noindent
Figure~\ref{fig:complex actions} shows what happens in general, as 
Figure~\ref{fig:graphsidedirected} (right) does for simple actions.
\begin{figure}[H]
 \begin{center}
  \includegraphics[width = 0.7\textwidth]{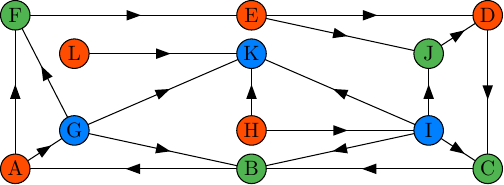}
  \caption{\emph{Illustration of the feedback for complex actions:} Whenever the learner asks for nodes $G$, $I$, and $K$, it receives the feedback also for the nodes $B$, $C$, $J$, and $F$, but not for the rest.}
 \label{fig:complex actions}
 \end{center}
\end{figure}

\noindent While we could extend \expix to this setting, combinatorial \expix could rarely be
implemented efficiently---we refer to \textcite{cesa-bianchi2012combinatorial,koolen10comphedge} for some 
positive examples.
As one of the main concerns in this chapter is computational efficiency, we take a different approach:
we propose a variant of \FPL \parencite{kalai2003efficient,hannan1957approximation} that efficiently implements the idea of implicit exploration in
combinatorial semi-bandit problems with side observations.
In each round~$t$, \FPL~bases its decision on some estimate $\hbL_{t-1}=\sum_{s=1}^{t-1}\hbl_s$ of the total losses
$\bL_{t-1} = \sum_{s=1}^{t-1} \bloss_s$ as follows:
\begin{equation}\label{eq:fpl}
 \bV_t = \argmin_{\bv\in\Sw} \bv\transpose\left(\eta_t \hbL_{t-1} - \bZ_t\right).
\end{equation}
Here, $\eta_t>0$ is a parameter of the algorithm and $\bZ_t$ is a perturbation vector with components drawn
independently from an exponential distribution with unit expectation. The power
of \FPL~lies in that it only requires an oracle that solves the (offline)
optimization problem $\min_{\bv\in\Sw} \bv\transpose \bloss$ and thus can
be
used to turn any efficient offline solver into an online optimization algorithm with strong guarantees.
To define our algorithm precisely, we need some further notation.
  We redefine $\F_{t-1}$ to be $\sigma(\bV_{t-1},\dots,\bV_1)$, 
 $O_{t,i}$ to be the indicator of the observed \textit{component} and let
\[
 q_{t,i} = \EEc{V_{t,i}}{\F_{t-1}} \qquad\mbox{and}\qquad o_{t,i} = \EEc{O_{t,i}}{\F_{t-1}}.
\]
The most crucial point of our algorithm is the construction of our loss estimates. To implement the idea of implicit
exploration by optimistic biasing,  we apply a modified version of the geometric resampling method of
\citet{neu2013efficient} constructed as follows: Let $\bO'_t(1),\bO'_t(2),\dots$ 
be independent copies\footnote{Such
independent copies can be simply generated by sampling independent copies of $\bV_t$ using the \FPL rule \eqref{eq:fpl}
and then computing $\bO_t'(k)$ using the observability 
$\cG_t$. Notice that this procedure requires no interaction between the learner
and the environment, although each sample requires an oracle access.} of $\bO_t$
and let
$U_{t,i}$ be geometrically distributed random variables for all $i \in [\nodes]$ with
parameter $\gamma_t$. We let
\begin{equation}\label{eq:resamp}
 K_{t,i} = \min\left(\ev{k: O_{t,i}'(k) = 1}\cup\ev{U_{t,i}}\right)
\end{equation}
and define our loss-estimate vector $\hbl_t\in\realset^\nodes$ with its $i$-th element 
as
\begin{equation}\label{eq:grix}
 \hloss_{t,i} = K_{t,i} O_{t,i} \loss_{t,i}.
\end{equation}
By definition, we have $\EEc{K_{t,i}}{\F_{t-1}} = 1/(o_{t,i} +
(1-o_{t,i})\gamma_t)$, implying that our loss
estimates are \emph{optimistic} in the sense that they lower bound the losses in expectation:
\[
 \EEcc{\hloss_{t,i}}{\F_{t-1}} = \frac{o_{t,i}}{o_{t,i} + (1-o_{t,i})\gamma_t} \loss_{t,i} \le \loss_{t,i}.
\]
Here we used the fact that $O_{t,i}$ is independent of $K_{t,i}$ and has expectation $o_{t,i}$ given $\F_{t-1}$.
We call this algorithm Follow-the-Perturbed-Leader with Implicit
eXploration (\fplix, \cite{kocak2014efficient}).
Note that the geometric resampling procedure can be terminated as soon as $K_{t,i}$ becomes well-defined for all $i$
with $O_{t,i}=1$. As noted by
\citet{neu2013efficient}, this requires generating at most $\nodes$ copies of $\bO_t$ 
on expectation. As each of these copies
requires one access to the linear optimization oracle over $\Sw$, we conclude
that the expected running time of \fplix
is at most $\nodes$ times that of the expected running time of the oracle. A
high-probability guarantee of the
running time can be obtained by observing that
$U_{t,i}\le \log\pa{\frac1\delta}/\gamma_t$ holds with
probability at least $1-\delta$ and thus we can stop sampling after at most
$\nodes\log\pa{\frac \nodes\delta}/\gamma_t$
steps with probability at least $1-\delta$.
The regret guarantee for \fplix using the approximation $\talpha_t$ of $\indn_t$ is stated below.
\begin{theorem}[Regret of \fplix by \textcite{kocak2014efficient}]
Assume that for all $t\in[\rounds]$, $\indn_t/C \le \talpha_t \le \indn_t \le \nodes$ for 
some $C>1$. Setting $\eta_ t =
\gamma_t = \sqrt{\pa{\log{\nodes} + 1}/\pa{m\pa{\nodes+\sum_{s 
=1}^{t-1}\talpha_s}}}$ and 
assuming $m\nodes>4$,
the regret of \fplix satisfies
 \[
  R_\rounds \le H m^{3/2} \sqrt{\pa{\nodes + C\textstyle\sum_{t = 1}^{\rounds} \indn_t}(\log \nodes + 
1)}, \quad  \mbox{where $H=\OO(\log (m\nodes \rounds))$}.
 \]
\end{theorem}
\section{Noisy side observations}\label{sec:noisyside}  
Until now in this chapter, we studied situations when the learner observes losses associated with some additional actions besides its own loss. 
This setting fails to address one important practical concern: 
in reality, one can rarely expect \emph{perfect} side-observations to be available. In the current section, we propose a 
similar model that can incorporate \emph{imperfect} side-observations corrupted by various levels of noise, depending on 
the problem structure.

\noindent
As an illustration of noisy setting, consider the problem of controlling solar panels so as to maximize their power
production. In this problem, the learner has to repeatedly decide about the orientation of the
panels so as to find alignments with strong sunshine. Besides the amount of the energy
being actually produced in the current alignment, the learner can also possibly base its decisions on measurements of
sensors installed on the solar panel. However, the observations
generated by these sensors can be of variable quality depending on visibility conditions, the quality of the sensors and
the alignment of the panels. Overall, this problem can be seen as a bandit problem with noisy side-observations fitting
into our framework, where actions correspond to alignments and the noisy side observations give 
information about similar alignments.

Formally, the learning protocol (Figure~\ref{fig:protocolnoisyside}) 
additionally assumes the knowledge of the  weight of 
each arc $i\ra j$ in $\cG_t$, which is denoted as $s_{t,(i,j)}$ and  assumed to lie in $[0,1]$. 
The feedback that the learner in the noisy setting is  
\[
c_{t,i} = \gweight_{t,(I_t,i)}\cdot \loss_{t,i} + \left(1-\gweight_{t,(I_t,i)}\right)\cdot \noise_{t,i}
\] 
for every arm $i$, where $\noise_{t,i}$ 
is the 
\emph{observation noise} (c.f.\@ another illustration on Figure~\ref{fig:fishingnoisy}).
We assume that each $\noise_{t,i}$ is zero-mean, satisfies $|\noise_{t,i}|\le R$ for some 
known constant $R\ge 0$, and is generated independently of all other noise terms and the history of the 
process.\footnote{We are mainly interested in the setting where $R = \Theta(1)$, that is, we are neither in the easy 
case where $R$ is close to zero or the hard one where it may be as large as $\Omega(\sqrt{\rounds})$.}

\begin{figure}[H]
\begin{center}
\includegraphics[width=0.45\textwidth]{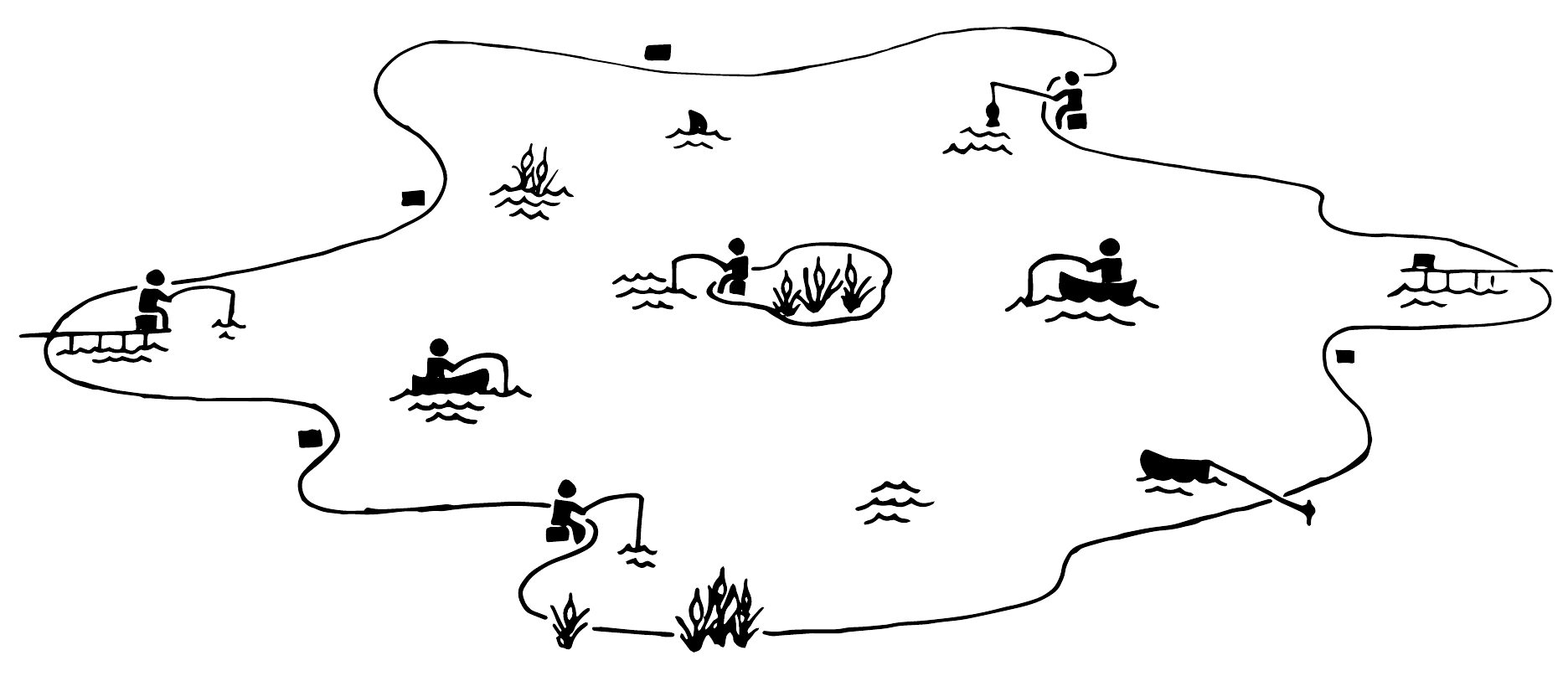}
\includegraphics[width=0.45\textwidth]{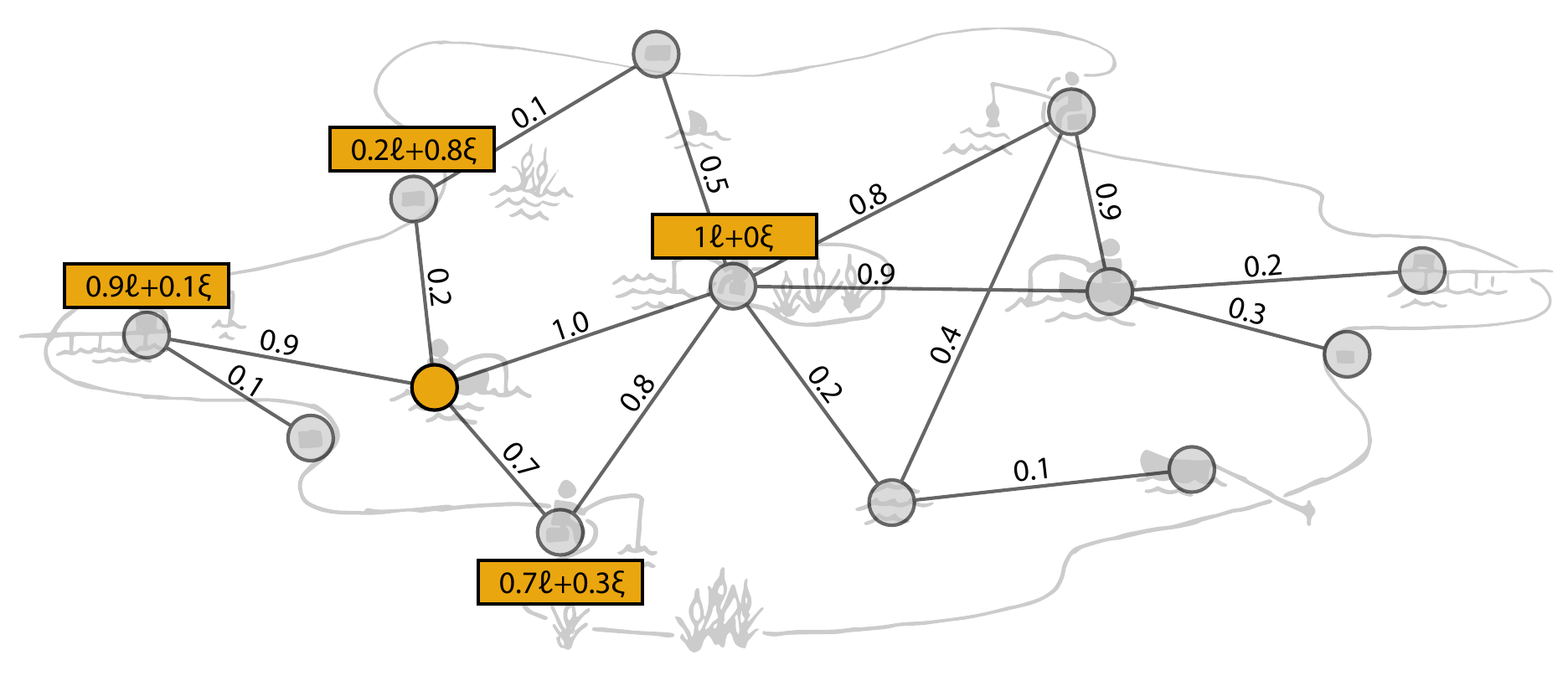}
\end{center}
\caption{\emph{Noisy feedback on a fishing example} \parencite{wu2015online,kocak2016online}: A~fisherman picks a fishing spot daily and gets the yield while imperfectly observing the yields of neighbors.}
\label{fig:fishingnoisy}
\end{figure}

\begin{figure}[t]
\centering
\fbox{
\begin{minipage}{.95\textwidth}
{\bfseries Parameters}: \\
\phantom{eye}set of arms $[\nodes]$, number of rounds $\rounds$.\\
{\bfseries For all $t=1,2,\dots,\rounds$ repeat}
\begin{enumerate}
 \item The environment picks a loss function $\loss_t:[\nodes]\ra [0,1]$ and a directed \textbf{weighted} graph $\cG_t$ with 
edge weights in $[0,1]$.
 \item Based on its previous observations (and possibly some source of randomness), the learner picks an action
$I_t\in[\nodes]$.
 \item The learner suffers loss $\loss_{t,I_t}$.
 \item The learner observes  $\cG_t$ and the feedback 
 \[
c_{t,i} = s_{t,(I_t,i)}\cdot \loss_{t,i} + \left(1-s_{t,(I_t,i)}\right)\cdot \noise_{t,i}
 \] 
 for every arm $i\in[\nodes]$.
\end{enumerate}
\end{minipage}
}
\caption{The protocol of online learning with \textbf{noisy} observations.}
\label{fig:protocolnoisyside}
\end{figure}
\noindent
Intuitively, in the case when the noise level of side observations does not change with time, 
a possible strategy one can think of is to use only the observations from the \emph{most reliable} sources and ignore
the rest. Having made the distinction between \emph{reliable} and \emph{unreliable}, the learner could model the observation
structure in the framework of \textcite{mannor2011from}, by treating every reliable observation
as \emph{perfect}. This approach raises two concerns. First, determining the cutoff for unreliable observations that
allows the \emph{most efficient} use of information is a highly nontrivial design choice. 
As we show later, knowing the \emph{perfect cutoff} would help us to improve performance over the
pure bandit setting without side observations. Second, one has to address the \emph{bias} arising from handling every
reliable observation as perfect. While one can think of many obvious ways to handle this bias by appropriate weighting
observations, none of these solutions are directly compatible with the model of \textcite{mannor2011from}.
A central concept in our performance guarantees is a new graph property that we call \emph{effective independence  
number}, defined as follows:
\begin{definition}
Let $\cG$ be a weighted directed graph with $\nodes$ nodes and edge weights  $s_{i, j}$ bounded in $[0,1]$. For all
$\varepsilon\in[0,1]$, let $\cG(\varepsilon)$ be the (unweighted) directed graph where arc $i\ra j$ is present if and only
if $\gweight_{i,j} \ge \varepsilon$ in $\cG$. Letting $\indn(\varepsilon)$ be the independence number of
$\cG(\varepsilon)$, the \textbf{effective independence number} of $\cG$ is defined as
\[
 \indnstar = \min_{\varepsilon\in[0,1]}\frac{\indn(\varepsilon)}{\varepsilon^2}\cdot
\]
\end{definition}
We first consider an algorithm that bases its decisions on the following estimates of each $\loss_{t,i}$:
\begin{equation}\label{eq:cand1}
\hloss\ti^{\ (\mbox{\textsc{b}})} = \frac{c_{t,i}}{\sum_{j=1}^\nodes p\tj s_{t,(j,i)} + \gamma_t }.
\end{equation}
where \textsc{b} stands for \emph{\textbf{b}asic}.
Here, $\gamma_t\ge 0$ is a so-called \emph{implicit exploration} (or, in short, IX) parameter first used by 
\citet{kocak2014efficient} for decreasing the variance of importance-weighted estimates. Notice that setting 
$\gamma_t = 0$, makes the estimates above unbiased since 
\[
 \EEcc{c_{t,i}}{\F_{t-1}} = \pa{\sum_{j=1}^\nodes p\tj s_{t,(j,i)}}\cdot \loss_{t,i},
\]
where we used our assumption that $\EE{\noise_{t,i}}=0$. Using these estimates in our algorithmic template \exph (see 
Algorithm~\ref{alg:expgraph}), one would expect to get reasonable performance guarantees. Unfortunately 
however, we were not able to prove a performance guarantee for the resulting algorithm. 

A close examination reveals that the reason for the poor performance of the above algorithm is the large variance of 
the estimates~\eqref{eq:cand1} which is caused by including observations from unreliable sources with small 
weights. One intuitive idea is to explicitly draw the line between reliable and unreliable sources by cutting 
connections with weights under a certain threshold. This effect is realized by the estimates
\begin{equation}\label{eq:cand2}
\hloss\ti^{\ (\mbox{\textsc{t}})} = \frac{c_{t,i}\II{s_{t,(I_t,i)}\ge \varepsilon_t}}{\sum_{j=1}^\nodes p\tj s_{t,(j,i)} 
\II{s_{t,(j,i)}\ge \varepsilon_t} + \gamma_t},
\end{equation}
where $\varepsilon_t\in[0,1]$ is a threshold value and \textsc{t} stands for \emph{\textbf{t}hresholded}. We call the algorithm 
resulting from using the above estimates in Algorithm~\ref{alg:expgraph} \expixt, standing for ``\exph with 
Implicit eXploration and Truncated side-observation weights''. Thanks to the thresholding operation, the variance of 
the loss estimates can be nicely controlled and it becomes possible to prove a strong performance guarantee for \expixt.
Note that if we choose $\varepsilon_t = \arg\min_{\varepsilon\in[0,1]}\frac{\indn_t(\varepsilon)}{\varepsilon^2}$ 
for 
all $t$, this essentially becomes $\tOO(\sqrt{\avgalpha \rounds})$ where $\avgalpha = \frac 1\rounds \sum_{t=1}^\rounds 
\indnstar_t$ is the average effective independence number of the sequence of graphs played by the environment. Note 
however that tuning $\varepsilon_t$ can be a very challenging task in practice, since computing independence 
numbers in general is known to be NP-hard. Even worse, computing the \emph{effective} independence number 
of a weighted graph can require computing up to $\nodes^2$ independence numbers. We propose an adaptive 
algorithm (\expwix) that does not need to tune this parameter and still manages to guarantee the same regret bound
without having to estimate any effective 
independence numbers. The key element of this algorithm is using loss estimates of the form
\begin{equation}\label{eq:est}
\hloss\ti = \frac{\gweight_{t,(I_t,i)} \cdot c\ti}{\sum_{j=1}^\nodes p\tj \gweight_{t,(j,i)}^2 + \gamma_t},
\end{equation} 
for which we prove the following guarantee.
\begin{theorem}[Regret of \expwix by \textcite{kocak2016online}]
\label{thm:mainTheoremWIX}
For all $t$, let $\indnstar_t$ be the effective independence number of $\cG_t$. Then, setting 
$\eta_t = \sqrt{\frac{{\log \nodes}}{{2(1+R+R^2)(\nodes + \sum_{s=1}^{t-1}Q_s)}}}$
and $\gamma_t =R\eta_t$, the regret of \expwix is bounded as
\[
R_\rounds = \tOO\pa{(1+R)\sqrt{\nodes+\sum_{t=1}^\rounds \indnstar_t}}.
\]
\end{theorem}

\section{Stochastic losses} 
In this section, we discuss few results for a simpler setting, when the node losses are coming from some fixed distribution. 
\textcite{caron2012leveraging}  proposed \UCBN and \UCBmaxN that closely follow \UCB, but in addition, they use side observations for better reward estimates (\UCBN) or choose one of the neighboring nodes with a better empirical estimate (\UCBmaxN).
These modifications enable to improve the guarantees of \UCB, i.e.,~the regret does not scale with the  number of nodes but  with the \emph{clique partition number}. 
Later, \textcite{buccapatnam2014stochastic} improved the results of~\textcite{caron2012leveraging} with \LP-based solutions and  
guarantees scaling  with the \emph{minimum dominating set}
and \textcite{kolla2016collaborative} considered a collaborative setting.

\subsection{Gaussian losses and side observations}
\textcite{wu2015online} considered an essentially identical model from Section~\ref{sec:noisyside} in the stochastic case.
In particular, they study partial-observability model for online learning: there, side observations are modeled as zero-mean Gaussian random 
variables with \emph{variance} depending on the chosen action. It is easy to see that their model and ours can capture 
exactly the same type of problems as in the adversarial setting: a side observation with zero variance in their model corresponds to a perfect 
observation with weight 1 while useless noise is equivalently represented by infinite-variance or 
zero-weight observations.  \textcite{wu2015online} assume that the losses are 
i.i.d.~Gaussian random variables while the results of  Section~\ref{sec:noisyside} hold without any assumptions made on the sequence of losses. The 
main contributions of \textcite{wu2015online} are (i) a general problem-dependent lower bound on the regret and (ii) 
algorithms that work under the assumption that all the useful (i.e., finite-variance) side-observations have the same 
variance. This latter assumption does not use the full strength of the framework  where the variance of side 
observations can vary for different actions.

\section{Lower bounds and high-probability bounds}
While the independence number $\indn$ can be much lower that the number of nodes $\nodes$, we may wonder 
whether it is the  \emph{right quantity} describing the \emph{difficulty} of the setting.
To support this, \textcite{mannor2011from} gave an $\Omega(\sqrt{\indn \nodes})$ bound in the undirected setting 
for an \emph{unchanging sequence} of graphs, $\cG_t = \cG, \indn_t = \indn$. 
Later, \cite{alon2013from} extended this lower bound to the directed case, still for unchanging sequence of graphs.
 
While the upper bounds in this chapter were given on the \emph{expected} regret, some algorithms also 
come with regret guarantees in \emph{high probability}.
\textcite{alon2014nonstochastic} gave a high-probability bound for \ELPP, a modified version of \ELP that 
with probability $1-\delta$ achieves the regret of $\OO\left(\sqrt{\log(\nodes/\delta)\sumt\mas(\cG_t)}\right)$,
where $\mas(\cG_t)$ is the size of the \emph{maximal acyclic subgraph}. While for undirected\footnote{where we consider two edges between the same nodes going the opposite direction}  graphs $\mas(\cG_t) =  \indn(\cG_t)$,  for directed graphs $ \indn(\cG_t) \leq \mas(\cG_t)$ in general and therefore
the bound is not as tight.
For \expix, \citet{neu2015explore} proved that with probability $1-\delta$, the cumulative regret of \expix is bounded by $\tcO(\sqrt{\indn \nodes})$,
which matches the lower bound of \textcite{mannor2011from} up to logarithmic factors.

Concerning the \emph{noisy} side observations, \textcite{wu2015online} showed an $\Omega(\sqrt{\indn \nodes}/\eps)$ lower bound 
on the regret for the special case of graphs, with all weights $\gweight_{ij}$ equal to either $0$ or $\epsilon$.
Note that this lower bound matches the upper bound of \expwix  (Theorem~\ref{thm:mainTheoremWIX}), 
since in that case $\indnstar = \indn / \eps^2$.

\section{Perspectives for side observations} 
In this section, we describe new challenges, related settings, and open problems for graph bandit learning with side observations.

\subsection{Beyond bandits}\label{ss:beyondbandits}
Besides the side observation models mentioned above, several other partial-observability models have been considered in the literature.
The most general of these settings is the \emph{partial-monitoring} framework considered by \citet{bartok2011minimax,bartok2014partial}. 
Unlike the side observation model, this framework is most useful for identifying and handling feedback structures that are \emph{more restrictive}
than bandit feedback. In contrast, learning with side observations deals with feedback structures that are strictly more expressive than
plain bandit feedback. Similarly to \textcite{bartok2011minimax}, the recent work of \citet{alon2015online} also considers a generalization of
the partial-observability models of \citet{mannor2011from} and \citet{alon2013from} that may be more restrictive than 
bandit feedback.

Specifically,  \citet{alon2015online} consider directed graphs with possible \emph{self-loops}. For a particular node, a self-loop means that 
whenever this node is selected, its loss is observed. Therefore, online learning on graphs with side observations, as defined
by \citet{mannor2011from} and considered above is a special case when all the self-loops are always present. More restrictive feedback
schemes emerge when some of the self-loops are not available, which means that the learner does not observe the loss of the chosen action, 
but still occurs this loss. Similarly to \textcite{bartok2011minimax}, they found that there are 3 classes of problems with 
$\tTheta(\sqrt{\rounds})$, $\tTheta(\rounds^{2/3})$, and $\tTheta(\rounds)$ regret and interestingly provide a \emph{complete characterization} of the settings
classifying all possible graphs in these three categories. Furthermore, a generalized version of \expg (Section~\ref{ss:expg}) can be 
used to attain these rates.

\textcite{cesa-bianchi2016delay} study yet another learning setting when the nodes 
cooperate to solve a nonstochastic bandit problem by communicating up to 
$\effd$ hops on the graph. 
Their \expcoop algorithm is shown to scale with $\indn_{\leq \effd}$, which 
is the independence number of the $\effd$-th power of the connected communication graph $\cG$.
Furthermore, \citet{ghosh2015ising} study a quite non-standard setting with Ising graph model.

Before to research in graph bandits and the quest for tight finite-time regret bounds, there 
was a prior work in economics and social sciences that studied the asymptotic convergence of learning for 
specific social models \parencite{bala1998learning,bala2001conformism,ellison1993rules,gale2003bayesian}.

\subsection{Graph generators}
One of the main practical drawback on the settings and algorithms presented in this chapter is the need
to see some parts of the graph, at least after the action was chosen.
Indeed, all previous algorithms for the studied setting
\citep{mannor2011from,alon2013from,kocak2014efficient}
require the environment to reveal a substantial part of a graph, at least 
after the side observations have been revealed. Specifically, these algorithms
require the knowledge of the \textit{second neighborhood} (the set of neighbors of the 
neighbors) of the chosen action in order to update their internal loss estimates.
On the other hand, they are able to handle arbitrary graph structures, 
potentially chosen by an adversary and prove performance guarantees 
expressed using graph properties based on cliques or independence sets.
In fact, it is difficult to get rid of this constraint, since
\textcite{cohen2016online} show that achieving nontrivial advantages from side observations may be impossible without perfectly known side-observation 
graphs when an adversary is allowed to pick \emph{both} the losses and the side-observation graphs. 
However, the situation is easier if we know something more about how $\cG_t$ is generated.

\subsubsection{Erd\H os--R\'enyi side-observation graphs} 
\begin{figure}
\begin{center}
\includegraphics[width = .45\textwidth]{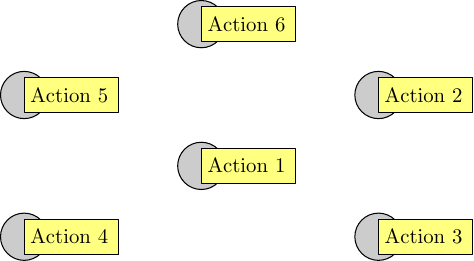}
\hspace{1cm}
\includegraphics[width = .45\textwidth]{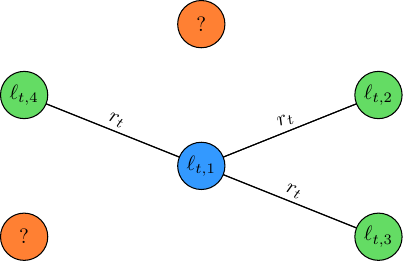}
\end{center}
\caption{\textbf{Left:}\@ The learner selects one of the actions (e.g., Action 1).  \textbf{Right:}\@ The nature generates an Erd\H os-R\'enyi graph with parameter $\erdosr_t$, where $\erdosr_t$ can be chosen by an adversary.}
\label{fig:sideobsER}
\end{figure}

Erd\H os--R\'enyi (ER) graphs \parencite{erdos1959on} are well studied random graphs where each edge is
generated uniformly at random with probability $\erdosr$ (Figure~\ref{fig:sideobsER}). If this probability is
fixed, but a new graph $\cG_t$ can be generated every round, then the regret of \expset
is of $\OO\left(\frac{2\left(\log \nodes \right)\rounds\left(1-\left(1-\erdosr\right)\right)^\nodes}{\erdosr} \right)$
\parencite{alon2013from}. Furthermore, generalizing the lower bound of \textcite{mannor2011from},
\textcite{alon2013from} also proved a $\Omega\Big(\sqrt{\rounds/\erdosr}\Big)$ lower bound for this setting 
in the case of a fixed graph. However, \expset still needs to have the knowledge of $\erdosr$
and to have the parts of the graph revealed after the actions. An interesting direction 
would be an algorithm that would not require this knowledge, since the probability 
of the side observation is $\erdosr$. Therefore, we can strive for an algorithm with 
$\tcO\left(\sqrt{\rounds/\erdosr}\right)$ regret in the fixed $\erdosr$ case and $\tcO\Big(\sqrt{\sumT (1/\erdosr_t) }\Big)$
in case of changing $\erdosr$. Note that when $\erdosr < 1/\nodes$, these bound are worse
than ignoring all side observations (the case of \EXP) and therefore the most interesting 
would be a procedure that does not do worse than \EXP.

In the case if $\erdosr_t$ is not to small, we provided \expres \parencite{kocak2016onlinea}, 
an algorithm that can efficiently estimate the losses without explicitly estimating $\erdosr_t$.
The main challenge in our setting is leveraging side observations \emph{without 
knowing $\erdosr_t$}. Had we had access to the exact value of $\erdosr_t$, we would be able to define the 
following estimate of $\loss_{t,i}$:
\begin{equation*}\label{eq:optest}
 \hloss_{t,i}^\star = \frac{O\ti\loss_{t,i}}{p_{t,i} + (1-p_{t,i})\erdosr_t} 
\end{equation*}
It is easy to see that the loss estimates defined this way are unbiased in the 
sense that
$\EEcc{\hloss_{t,i}}{\F_{t-1}} = \loss_{t,i}$ for all $t$ and~$i$.
It is also straightforward to show that an appropriately tuned instance of the \exph algorithm of 
\textcite{auer2002nonstochastic} fed with these loss
estimates is guaranteed to achieve a regret of $\cO(\sqrt{\sum_t(1/\erdosr_t)\log \nodes})$ (see also 
\cite{seldin2014prediction}). 
or any fixed $t,i$, we now describe an efficiently computable surrogate $G\ti$ for the geometrically distributed random variable $G\ti^\star$ 
with parameter $o\ti$ that will be used for constructing our loss estimates. In particular, our strategy will be to construct several 
independent copies $\ev{O'_{t,i}(k)}$ of $O_{t,i}$ and choosing $G_{t,i}$ as the index $k$ of the first copy with $O'_{t,i}(k)=1$. It is 
easy to see that with infinitely many copies, we could exactly recover $G_{t,i}^\star$; our actual surrogate is going to be weaker thanks to 
the smaller sample size. For clarity of notation, we will omit most  explicit references to $t$ and $i$, with the understanding that all 
calculations need to be independently executed for all pairs $t,i$. 

Let us now describe our mechanism for constructing the copies $\ev{O'(k)}$.
Since we need 
independence of $G\ti$ and $O\ti$ for our estimates, we use only side observations from actions $[\nodes]\setminus\ev{I_t,i}$. First, let's 
define $\sigma$ as a uniform random permutation of $[\nodes]\setminus\ev{I_t,i}$.
For all $k\in[\nodes-2]$, we define $R(k) = O_{t,\sigma(k)}$. Note that due to the construction, $\{R(k)\}_{k = 1}^{\nodes-2}$ are pairwise 
independent Bernoulli random variables with parameter $\erdosr_t$, independent of $O\ti$. Furthermore, knowing $p\ti$ we can define 
$P(1),\,\ldots,\,P(\nodes-2)$ as pairwise 
independent Bernoulli random variables with parameter~$p\ti$. Using $P(k)$ and $R(k)$ we define the random variable $O'(k)$ as 
\[
O'(k) = P\left(k\right) + \left(1-P\left(k\right)\right)R\left(k\right)
\]
for all $k\in[\nodes-2]$. Using independence of all previously defined random variables, it is easy to check that the 
variables $\{O'(k)\}_{k = 1}^{\nodes-2}$ are pairwise independent Bernoulli random 
variables with expectation $o\ti = p\ti + (1-p\ti)\erdosr_t$. Now we are ready to define $G\ti$ as
\begin{equation*}\label{eq:est_G}
G\ti = \min\ev{k\in[\nodes-2]:O(k)'=1}\cup\ev{\nodes-1}.
\end{equation*}
We can show that $G\ti$ follows a truncated geometric law in the sense that
\[
 \PP{G\ti = m} = \PP{\min\ev{G\ti^\star,\nodes-1} = m}
\]
holds for all $m\in[\nodes-1]$. Using all this notation, we construct an estimate of $\loss_{t,i}$ as
\begin{equation}\label{eq:est_R}
 \hloss_{t,i} = G_{t,i} O_{t,i} \loss_{t,i}.
\end{equation}
The rationale underlying this definition of $G_{t,i}$ is rather delicate. First, note that $p\ti$ is deterministic  given the history 
$\F_{t-1}$ and therefore, does not depend on
$O\ti$. Second, $O_{t,i}$ is also independent of$O_{t,j}$ for $j\not\in\{i,I_t\}$. As a result, $G\ti$ is independent of $O\ti$, and we 
can use the identity $\EEt{G_{t,i}O_{t,i}} = \EEt{G_{t,i}}\EEt{O_{t,i}}$. 
Using the estimates from Equation~\ref{eq:est_R} in Line~\ref{expgrap:line:lossestimaes} of Algorithm~\ref{alg:expgraph}, 
we get the \expres algorithm. The next theorem states our main result concerning \expres with an adaptive learning rate.
\begin{theorem}[Regret of \expres by \textcite{kocak2016onlinea}]
Assume that $\erdosr_t\ge \frac{\log \rounds}{2\nodes-2}$ holds for all $t$ and set
$\eta_{t} = \sqrt{\frac{\log \nodes }{\nodes^2+\sum_{s = 1}^{t-1} \sumi p_{s,i}(\hloss_{s,i})^2}}.$
Then, the expected regret of \expres satisfies
\[
R_\rounds\leq 2\sqrt{\pa{\nodes^2 +  \sumT\frac{1}{\erdosr_t}}\log \nodes} + \sqrt{\rounds}.
\]
\end{theorem}
The most obvious question and currently an open problem is whether it is possible to remove our assumptions on the values of $\erdosr_t$. We can only give a definite answer in the simple case when all~$\erdosr_t$-s are identical: In this case, one can think of simply computing the empirical frequency $\widehat{\erdosr}_t$ of all 
previous side observations in round $t$ to estimate the constant $\erdosr$. 

Besides Erd\H os--R\'enyi graphs, another direction would be the extension of the known results 
to side information in Barab\'asi-Albert (\cite*{barabasi1999emergence})
or Watts-Strogatz (\cite*{watts1998collective}) model, 
or other models better suited for some real-world graphs (e.g., social networks).

\subsubsection{Side observations in the communities}
\begin{wrapfigure}[16]{R}{0.40\textwidth}
\vspace{-15pt}
\includegraphics[width=2.2in]{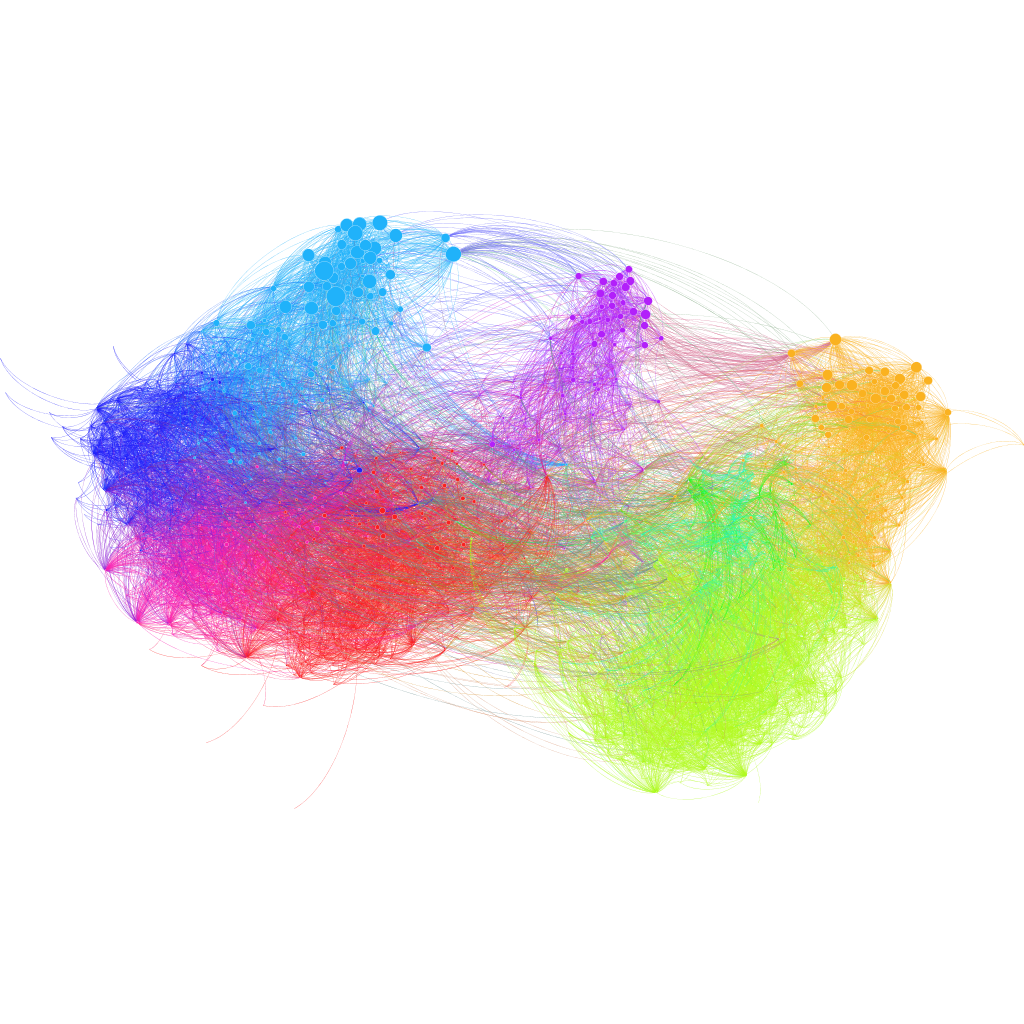}
\caption{Communities.}
\label{fig:communities}
\end{wrapfigure}
One typical target scenario for the setting 
in this chapter is advertising on social networks, where the advertiser
chooses a target user and besides their feedback receives (as side observations) 
also the feedback of their contacts. 
Social networks are often modeled as a set of (overlapping) communities (Figure~\ref{fig:communities})
and therefore an  extension is to consider an assumption 
that graphs we deal with have a community structure.
First, we can consider the communities as yet another model for random graphs. 
The most studied model is the \emph{stochastic block model} and its variants \citep{girvan2002community}. 
Second, we may have access to the community model and consider the case 
where each community gives side observations with their own probabilities, which are unknown to the learner.

\chapter{Influence maximization}

\emph{Product placement} is another marketing application that we target with graph bandits. An advertiser can offer a product to some  users in a hope that they will recommend the product to their contacts, i.e.,~to the neighboring nodes in a social network. The advertiser then observes the set of contacts that these users have influenced and that have  bought the product. The objective of the advertiser is to target \textit{influential users}, the nodes of the graph whose influence is the most important. Ideally, the advertiser would only offer products to the users with maximum influence.

Furthermore, there are many models of influence and some of the known ones were introduced in the seminal work on spreading 
the influence through a social network~\citep{kempe2003maximizing,kempe2015maximizing}.
In this chapter, we focus on \emph{local influence}, where a node on the graph influences
only its \emph{immediate neighborhood} and outline the road for more \emph{global models}.

We finished the previous chapter by stating that 
most of the existing approaches for active learning on graphs assume that either the \textit{entire graph}  is known in advance, or at least that a substantial \emph{part of the graph} is revealed to the learner after it selected the node. Typically, the algorithms require at least the knowledge of the set of neighbors of the neighbors of the nodes (\emph{second neighborhood}). 
This knowledge of the graph is crucial for existing learning algorithms \citep{mannor2011from,yu2011unimodal,caron2012leveraging,cesa-bianchi2013gang, alon2013from,gentile2014online,kocak2014efficient,gu2014online,valko2014spectral,
buccapatnam2014stochastic,alon2015online} to help them learn faster than in the case if no structure existed.
However, in some realistic scenarios, the graph information is \emph{not available}
to the learner beforehand. Typically, the operator of the social network would not freely reveal the social links and therefore the graph is not known to the advertiser. On the other hand, for instance, in the advertising example presented above, the advertiser has some local access to the social network in the sense that they can get information of the set of users that were influenced to purchase products through the other targeted customers. This information can be gathered through \emph{promotional codes} when the goal is \emph{product purchase}  or through \emph{likes}  in an \emph{information campaign}~\citep{caron2012leveraging}.

However, the existing graph bandit approaches do not allow to treat this scarce side information setting. 
Therefore, with the known tools, one can either (i) first thoroughly explore the graph and then apply existing graph bandit strategies, or (ii)  forget about the underlying graph structure and apply existing multi-arm bandit algorithms to the nodes of the graph. In both cases, it is necessary that the learner substantially explores the graph and therefore \emph{samples many nodes}, if not all of them. This is not very reasonable, for instance, in our marketing example, since graphs corresponding to social networks are usually  large. Moreover, the advertiser is unlikely to have a large enough budget to target all the nodes of the graph in order to learn which ones are the most  influential.

\section{Local influence and revelation bandits}\label{sec:localinfluence}

Let $\cG$ be a graph with $\nodes$ nodes. When a node $i$ is selected, it can influence the nodes of $\cG$, including itself.  Node $i$ influences each node $j$ with \emph{fixed} but \emph{unknown} probability $p_{i,j}$ (Figure~\ref{fig:revealing}). Let $\bM = (p_{i,j})_{i,j}$ be the $\nodes\times \nodes$ matrix that represents~$\cG$. 
We consider the following online, active setting. At each round (time) $t$, the learner chooses a node $k_t$ and observes which nodes are influenced by~$k_t$, i.e., the set $S_{k_t,t}$ of influenced nodes is \emph{revealed}. 
Given a budget of  $\rounds$ rounds, the objective is to maximize the number of \emph{influences} that the selected node exerts. Formally, our goal is to find the strategy maximizing the performance
\[\text{Reward}_\rounds = \sum_{t = 1}^\rounds \left|S_{k_t,t}\right|.\]

\begin{wrapfigure}[9]{R}{0.40\textwidth}
\vspace{-2.4cm}
\hspace{37pt}
\includegraphics[width=2.3in]{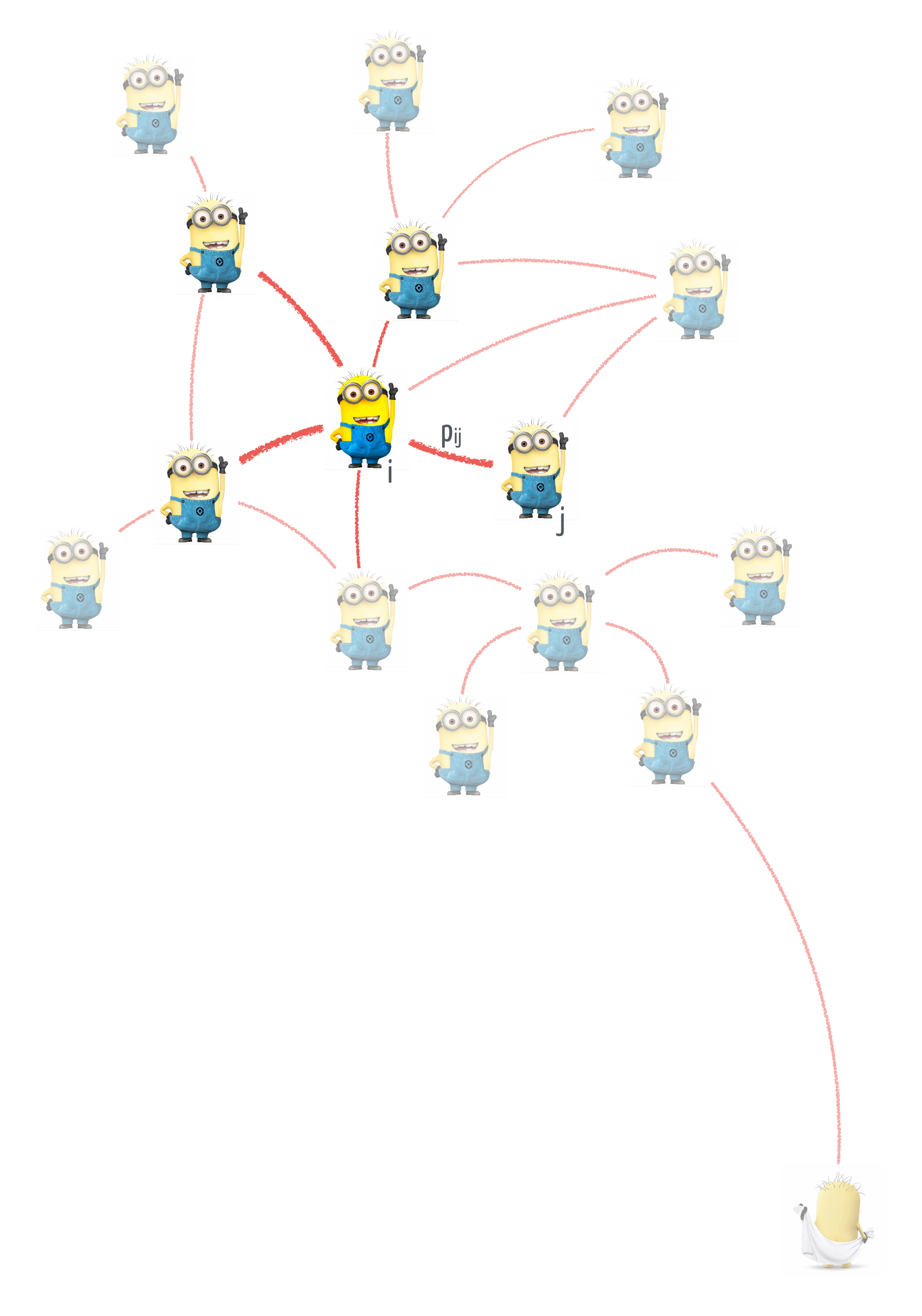}
\vspace{-3.8cm}
\caption{Influence probability $p_{i,j}$.}
\label{fig:revealing}
\vspace{-5cm}
\end{wrapfigure}
\noindent 
The \textit{influence} of node $k$, i.e., the expected number of nodes that node $k$ exerts influence on, is by definition
\[r_k = \EE{\left|S_{k,t}\right|} = \sum_{j \leq \nodes} p_{k,j}.\]
We also define the \textit{dual influence} of node $k$ as
\[\rkdual = \sum_{j \leq \nodes} p_{j,k}.\]
This quantity is the expected number of nodes that exert influence on node $k$.  For an undirected graph $\cG$,  $\bM$ is symmetric and $\rkdual = r_k$. However, in general, this is not the case, but we assume that the influence is up to a certain degree mutual.
 In other words, we assume that if a node is very influential, it also is subject to the influence of many other nodes. 

As the performance measure, we compare any \emph{adaptive strategy} for this setting with the optimal oracle that knows $\bM$. The oracle strategy always chooses one of the most influential nodes, which are the nodes whose expected number of influences $r_k$ is the largest. We call one of these node $k^\star$, such that
\[k^\star = \argmax_k \EE{\sum_{t = 1}^\rounds \left|S_{k,t}\right|} =  \argmax_k \rounds r_k.\]
Let the reward of this node be 
\[r_\star = r_{k^\star}.\]
Then, its expected performance, if it consistently sampled $k^\star$ over $\rounds$ rounds, is equal to
\[\EE{\text{Reward}^\star_\rounds} = \rounds r_\star.\]
The expected \emph{regret} of any adaptive strategy that is unaware of $\bM$, with respect to the  oracle strategy, is defined as the expected difference of the two, 
\[\EE{R_\rounds} = \EE{\text{Reward}^\star_\rounds} - \EE{\text{Reward}_\rounds}.\]
Dually, we define $\rstardual$ as the average number of influences received by the most influenced node,
\[\rstardual = \max_{k}\rkdual.\]

\noindent
First, note that the minimax-optimal rate in this setting is the same as in the restricted information case, 
when we ignore the identity of the influenced nodes and only use the number of them as a reward. To see that, one can, for instance, consider a network composed of isolated nodes with only a very small clique of most influential nodes, connected only to each other.  Another example is a graph where the fact of being influential is uncorrelated with the fact of being influenced and where, for instance, the most influential node is not influenced by any node. Therefore, when $\rounds \leq \nodes$,  there is no adaptive strategy in a minimax sense, also in this unrestricted setting we just defined.

However, the cases where the identity of the influenced nodes does not help, are somewhat pathological. Intuitively, they correspond to cases where the graph structure is not very informative for finding the most influential node. This is the case when there are many isolated nodes, and also in the case where observing nodes that are very influenced does not provide information on these nodes' influence. In many typical and more interesting situations, this is not the case. First, in these problems, the nodes that have high influence are also very likely to be subject being influenced, for instance, many interesting networks are symmetric and then it is immediately the case. Second, in realistic graphs, there is typically a small portion of the nodes that are noticeably more connected than the others~\citep{barabasi1999emergence}.

In order to rigorously define these nondegenerate cases, let us first define the function $\detD$ that controls the number of nodes with a given \emph{dual gap}, i.e.,~a given suboptimality with respect to the most influenced node.
\[ \detD(\Delta) = \left|\left\{i \leq \nodes : \rstardual - \ridual \leq \Delta\right\}\right|. \]
The function $\detD(\Delta)$  is a nondecreasing quantity dual to the arm gaps. Note that $\detD(r) = \nodes$ for any $r \geq \rstardual$ and that $\detD(0)$ is the number of most influenced nodes.
We now define the \emph{problem dependent} quantities that express the  difficulty of the problem and allow us to state our results.   
\begin{definition} 
 We define  the \textbf{detectable horizon} as the smallest integer $\rounds_\star>0$ such that
\[\rounds_\star\rstardual \geq \sqrt{\detDstar \rounds \rstardual},\]
when such $\rounds_\star$ exists and  $\rounds_\star = \rounds$ otherwise. Here, $\detDstar$ is  the \textbf{detectable dimension} defined as
\[\detDstar = \detD(\Delta_\star),\]
where the \textbf{detectable gap} $\Delta_\star$ is defined as 
\[\detDstar \eqdef 16\sqrt{\frac{\rstardual \nodes \log\left(\rounds \nodes\right)}{\rounds_\star}}+ \frac{144\nodes\log\left(\rounds \nodes\right)}{\rounds_\star}.\]
\end{definition}
\begin{remark} 
From the definitions above, the detectable dimension is the $\detDstar$ that corresponds to the smallest integer $\rounds_\star>0$ such that
\[\rounds_\star\rstardual \geq \sqrt{\detD\left(16\sqrt{\frac{\rstardual \nodes \log\left(\rounds \nodes\right)}{\rounds_\star}}+ \frac{144\nodes\log\left(\rounds \nodes\right)}{\rounds_\star}\right) \rounds \rstardual},\]
or $\detDstar=\nodes$ if such $\rounds_\star$ does not exist. It is therefore a well defined quantity. Moreover, since $\detD$ is nondecreasing and $\detD(0)$ is the number of most influenced nodes, then $\detDstar$ converges to the number of most influenced nodes as $\rounds$ tends to infinity. 
\end{remark}
\noindent
Finally let us write the influential-influenced gap as
\[\varepsilon_\star \eqdef r_\star - \max_{k\in \Ddualset} r_k,\]
where $\Ddualset \eqdef \{ i : \ridual  = \max_k \rkdual\}$. 
The quantity $\varepsilon_\star$ quantifies the gap between the most influential node overall vs.\@ the most influential node in the set of most influenced nodes.
\begin{remark} 
The quantity $\varepsilon_\star$ is small when one of the most influenced node is also very influential.
It is exactly zero when one of the most influential nodes happens to also be one of the most influenced nodes. For instance, the case   $\varepsilon_\star = 0$  appears in undirected social network models with mutual influence. 
\end{remark} 
\noindent
The graph structure is helpful when the $\detD$ function decreases quickly. 
To get an intuition, consider  a star-shaped graph which is  the most helpful and can have $\detDstar = 1$ even for a small $\rounds$. On the other hand, a bad case is a graph with many small cliques. The worst case is where all nodes are disconnected except $2$, where $\detDstar$ will be of order $\nodes$ even for a large $\rounds$. 

The detectable dimension $\detDstar$ is a problem dependent quantity  that represents the \emph{complexity of the problem} instead of $\nodes$.  In real networks, $\detDstar$ is typically smaller than the number of nodes $\nodes$. 
 As our analysis will show, $\detDstar$ represents the number of nodes that we can \emph{efficiently extract} from the mass of the $\nodes$ nodes in less than $\rounds$ rounds of the time budget. Our \emph{bandit revelator} algorithm, \BARE \parencite{carpentier2016revealing}, starts by the \emph{global exploration} phase and extracts a subset of cardinality less than or equal to $\detDstar$, that contains a very influential node, that is at most $\varepsilon_\star$ away from the most influential node. \BARE does this extraction  \textit{without scanning all the $\nodes$ nodes}, which could be impossible, anyway since we do not restrict to $\nodes \leq \rounds$.
In the subsequent \emph{bandit} phase, \BARE proceeds with scanning this smaller set of selected nodes to find the most influential one.

We now state our main theoretical result that proves a bound on the regret of \BARE.

\begin{theorem}[Regret of \BARE by \textcite{carpentier2016revealing}]\label{bare}
In the unrestricted local influence setting with information on the neighbors, \BARE satisfies, for a constant $C>0$,
\[\EE{R_\rounds} \leq C\min\left(r_\star \rounds, \detDstar r_{\star} + \sqrt{r_{\star}\rounds \detDstar} + \rounds \varepsilon_\star\right).\]
\end{theorem}
While detectable dimension $\detDstar$ behaves as we expect, it does not seem to be directly linked with 
some previously known graph concept (as it was the case for the side observations and independence number).
In fact, the graph-dependent only quantity is the function $\detD(\Delta)$, that quantifies the amount of $\Delta$-suboptimal most influenced nodes. However, the detectable dimension itself is tied to the bandit problem by essence (and the constants are due to the Bernstein bound) --- it is the quantity that realizes the optimal tradeoff between the regret suffered during the global exploration phase, and the regret suffered during the bandit phase. To support this claim, we give a lower bound that features this quantity. Notice that the influential-influence gap also appears in it.

\begin{theorem}[Lower bound for local influence setting by \textcite{carpentier2016revealing}]\label{bare2}
Let $\nodes \geq C\rounds>0$ where $C>0$ is a universal constant. Consider the set of  local influence setting and the set of all problems that have maximal influence bounded by $r$, detectable dimension smaller than $\detD \leq \nodes/2$ and influential-influence gap smaller than $\varepsilon$. Then the expected regret of the best possible algorithm in the worst case of these problems is lower bounded as
\[C''\min\left(r\rounds,  \detD r_\star + \sqrt{r\rounds \detD}  + \rounds \varepsilon\right),\]
where $C''$ is a universal constant.
\end{theorem}

\paragraph{Large scale setting} The quantity $\detDstar$ and \BARE become particularly appealing when we consider an interesting practical situation with a large number of graph nodes. For instance, even in a medium-sized social network, the advertiser 
would not have enough budget to target all the users and discover the most influential one, i.e.,\@ $\rounds \leq \nodes$. 
Notice again, that in the restricted setting, the regret of bandit strategies in this problem for $\rounds \ll \nodes$ is of order
$\rounds r_\star,$
which is larger than the regret of \BARE.

\section{Perspectives of bandit influence maximization}
In this section, we outline some extensions of the simple model of influence described above.

\subsection{Global models of influence}
In Section~\ref{sec:localinfluence}, we discussed the local influence model. 
In computational social sciences, we usually consider more involved, global models
of the influence spread over a social graph. The most known 
and studied are the models described in the seminal paper
of \textcite{kempe2003maximizing}, in particular, the \emph{independent cascade}
model. In this model, we consider a set of seed nodes $A_0 \subseteq \nodeset$
and a probability $p_{ij}$ associated with each edge. Independent cascade model 
defines an activation process of nodes, where at the beginning, all nodes in $A_0$
are active and subsequently every node~$i$ can activate its neighbor $j$ with 
probability $p_{ij}$ once, independent of the history of the process. This process
runs until no more activations are possible. Given the set of probabilities, 
$\ac{p_{ij}}_{ij}$, the goal in the (offline) influence maximization problem 
is to find such $A_0$ that maximizes the expected number of influenced nodes. 
Obviously, this property is trivially maximized for the whole node set $A_0 = \nodeset$, but 
we are typically interested in the  $\card{A_0}\leq k$, where $k$ is coming from the budget constraint
of how many people we can afford to reach. 
This \emph{offline} problem is NP-hard, but as the (expected) number of influence nodes
is a submodular set function, it can be approximated within the factor of $1-1/e$ \parencite{kempe2003maximizing}.

Similarly to the local influence (Section~\ref{sec:localinfluence}), in the bandit setting, 
the set of activation probabilities $\ac{p_{ij}}_{ij}$ are \emph{unknown} to the learner. 
In this simplest case, $k=1$ and we are interested in selecting a single, most influential node.
In general, $k\in\sqpa{\nodes-1}$ and this is an instance of combinatorial bandits.
We can consider several feedback settings:
\begin{enumerate}
\item \emph{full bandit: }the learner only observes the \emph{number} of influenced \emph{nodes}
\item \emph{node semi-bandit:} learner observes the \emph{identity} of the influenced \emph{nodes}
\item \emph{edge semi-bandit:} learner observes the \emph{identity} of the activated \emph{edges}
\end{enumerate}
Notice that in the edge semi-bandit setting, we observe a sample (one)
from $\Bernoulli(p_{ij})$ for each activated edge. Moreover, we also 
receive a sample (zero) from $\Bernoulli(p_{ij})$ for each nonactivated edge, 
when at least one of $i$ or $j$ nodes were activated. The node semi-bandit feedback is more challenging, since 
we do not observe the activation edges, and therefore we do not know what 
was the activation path for that node, which makes the estimation of $p_{ij}$ nontrivial.

For the full bandit feedback 
and $k=1$, we can obtain results similar to those as in the restricted setting 
considered in the local influence model. 
On the other hand, the understanding of the semi-bandit feedback for the influence maximization
problem is still an open problem. Nonetheless, we comment on some recent attempts and results. 
Recently, \textcite{lei2015online} investigated the combinations of offline influence maximization 
approaches with multi-arm bandit strategies for the online influence maximization 
in the edge semi-bandit case. \textcite{lei2015online} tried several combinations of bandit techniques ($\epsilon$-greedy, confidence-based methods)
and empirically showed that their methods perform well, however, they did not provide
any guarantees or analysis. \textcite{chen2015combinatorial} also considered 
combinatorial edge semi-bandit case and showed that the reward function 
of this problem is a special case of their general combinatorial semi-bandit 
case satisfying \emph{monotonicity} and \emph{bounded-smoothness} conditions.
Therefore, their algorithm (\CUCB) and analysis apply. However, their
analysis is general and distribution-dependent only and it is not clear 
how it relates to the structure of the graph. Furthermore,  both of their gap-dependent and gap-free bounds are problematic because they depend on the reciprocal of the minimum observation probability$p^\star$ of an edge. Consider a line graph with $L$ edges where all edge weights are $0.5$. Then $1/p^\star$ is~$2^{L - 1}$.
To avoid this problem, we proposed \parencite{wen2016influence}  \imb, a linear \UCB-like algorithm for edge semi-bandits that permits linear generalization and is suitable for large-scale problems. We bounded the regret of \imb when the structure of the network is a \emph{forest} \parencite{wen2016influence}. Our regret bounds are polynomial in all quantities of interest; reflect the structure and activation probabilities of the network; and do not depend on inherently large quantities, such as the reciprocal of the minimum probability of being influenced and the cardinality of the action set. The forest is important in practice because influence maximization in general graphs is computationally expensive, and known scalable approximations use forests to evaluate only most influential paths, such as in the maximum influence arborescence (MIA) model \parencite{chen2010scalable}. Furthermore, \textcite{vaswani2015influence} consider the more difficult, 
\emph{node} semi-bandit setting. In this setting, however, it is unknown 
which edge was alive and should have its estimate updated. 
\textcite{vaswani2015influence} decide to update 
one of the edges that could have been alive uniformly at random.
It is not clear whether it is possible to do better
and also, what is the equivalent of \emph{detectable 
dimension} for this model. 
Another direction is  to estimate the influence function
using the recent results studying 
learnability of influence in networks \parencite{narasimhan2015learnability}.
Finally, the problem gets even more challenging when we allow the influence probabilities to change \parencite{bao2016online},
when we allow the seed set to be chosen adaptively \parencite{vaswani2016adaptive},
or when we consider a continuous model  \parencite{farajtabar2016multistage}.
To sum up, bandit 
influence maximization under global models remains a very interesting open problem.

\subsection{Crawling bandits}
In Chapters~\ref{chap:smoothness} and~\ref{chap:side}, all methods needed to have access
to parts of the graph for various learning reasons. In the present chapter, we lifted the assumption 
on the knowledge of the edge set and the learner had to also \emph{estimate the graph structure} in order
to act on it.  Yet the learner was allowed to choose \emph{any node at any round}. In a more challenging case, even this possibility can be \emph{restricted}.

As we mentioned before, the inability of the learner to access the full graph 
as desired can come from some external factors. In the context of advertising
in social networks, the social network provider can have reasons 
to conceal the social graph: privacy, business advantage, or intention of charging for this information.
This poses an additional challenge for the learner who can only see (some) neighbors of the previously chosen nodes.
Such process resembles \emph{crawling} the websites through the links to collect some information
or discover interesting new sites. \textcite{singla2015information} formalizes a specific 
set of these constraints for general \emph{utility functions} using the parameters $l_{\rm deg}$ and $l_{\rm val}$, 
where $l_{\rm deg}$ quantifies \emph{selectability} of the new nodes and $l_{\rm val}$
the \emph{observation possibility} of the new nodes and 
notes that  in real-world social networks (such as Facebook or LinkedIn),
the visibility is usually restricted to $l_{\rm deg} = 1$  and  $l_{\rm val} = 1$ due
to privacy settings. This means that the learner can typically only see and access (select in the next step)
the local ($1$-hop) neighborhood of the nodes already selected.
\textcite{singla2015information} uses this restriction parametrization 
for specific set discovery problems.  It is an open 
problem what algorithm would be optimal for (cumulative) regret minimization.
This setting is also related to the \emph{volatile multi-armed bandits} where the set of possible arms changes \parencite{bnaya2013social}.

Note that the constraints on the visibility of the graphs are not only applicable in the influence maximization 
setting, they are relevant in other graph bandits, for instance in learning with side observations (Chapter~\ref{chap:side}).

\part{Stochastic bandits in large structured domains}
The whole previous part was dedicated to settings where the actions (arms) are the graph nodes.
Not all action spaces naturally form a graph and in this part we focus
on other \emph{structured spaces}.
In Chapter~\ref{chap:kernel}, we describe the frequentist analysis 
of kernelized bandits \parencite{valko2013finite}, closely related to Gaussian process bandits \parencite{srinivas2009gaussian}.
Kernelized bandits are a simple extension of linear bandits to reproducing kernel Hilbert spaces (RKHS).
In Chapter~\ref{chap:polymatroid}, we consider \emph{polymatroid bandits} \parencite{kveton2016learning},
that generalize the notion of linear independence to other structures,
where the optimization over combinatorial action spaces can be done
efficiently (in the offline case) using the simple \greedy algorithm.

While kernelized and polymatroids bandits are instances of discrete action 
spaces, in the rest of this part we give examples of a structure
in \emph{continuous} ones. 
First, in Chapter~\ref{chap:function}, 
we apply bandit strategies to \emph{black-box function optimization} with noisy evaluations, 
where the action space is a (bounded) continuous
domain of some unknown function $f$. The structure of rewards 
in this setting is the \emph{smoothness} around one of the optima of $f$.
However, in the most general setting, we treat the case 
when this smoothness is \emph{unknown} to the learner
and we  show that we are able to provide almost the same 
guarantees on the error (\emph{simple regret)} as if this smoothness was available.
Second, in Chapter~\ref{chap:infi} we look into another bandit 
setting with continuous arm set, but this time with \emph{no topological 
or metric assumptions} between the arms. In other words, no arm
can give any information about any other arm. 
This setting was formalized by \textcite{berry1997bandit}
as \emph{infinitely many arms} bandits and we 
focus on the \emph{simple regret} in this setting, same 
as in Chapter~\ref{chap:function}.

In the previous graph bandit part, the common thread was the study 
of graph-dependent quantities (independence number, detectable dimension, 
number of relevant eigenvectors, \dots)   for different settings that embodied 
different difficulties  of the problems. We studied algorithms that took advantage 
of the graph and were able to get faster rates as
functions of these graph-dependent quantities instead of  
the number of nodes  
$\nodes$. Our intention in this part is very similar.
What are the \emph{sizes of action sets} considered here?
First, the kernelization of linear bandits in kernel bandits takes the dependence
on the dimension $D$ of the context to the dimension of RKHS, that is possibly infinite. 
Second, the space of actions in polymatroid bandits is combinatorial 
(possibly exponential) in the number of items. Finally, in 
both bandits for function optimization and infinitely many arm bandits, 
the arms form a continuous set.
Henceforth, while in graph bandit part, we had a choice 
of ignoring the graph structure, treat the settings
as multi-arms bandits and get a (likely worse) dependence on the number of nodes $\nodes$;
taking the same path for the settings considered in this part and ignoring the present structure would 
be \emph{hopeless}. As a consequence, our quest is to find the appropriate problem-dependent 
quantities also for the large structure settings of this part.

In the case of kernelized bandits, we define a notion 
of effective dimension, measuring the \emph{decay
of eigenvalues} of the covariance matrix in  kernel regression. 
Next, for polymatroid bandits, we show an algorithm
whose regret scales with the \emph{rank of the polymatroid}
(matching the lower bound in the matroid case).
Furthermore, in the black-box function optimization setting, 
we consider the \emph{near-optimality dimension}, which measures the complexity of
the optimization problem. 
Finally, in the case of infinitely many arm bandits, 
we give an algorithm optimizing simple regret
with the near-optimal guarantees, that depend on a
parameter $\infibeta$, characterizing the \emph{distribution of the
near-optimal arms}, same $\infibeta$ as in the cumulative regret version of \textcite{berry1997bandit}.

\chapterimage{chapter_head_1.pdf} 
\chapter{Kernel bandits}\label{chap:kernel} 
This chapter considers a generalized version of the setting of spectral bandits (Section~\ref{sec:spectral}).
Unlike in linear bandits \parencite{auer2002using} we avoid a possibly costly feature-engineering step
by assuming  that we have access to the similarities between actions' contexts
and that the expected reward is an \textit{arbitrary} linear function of the
contexts' images in the related 
reproducing kernel Hilbert space (RKHS).
In the following, we show how to derive \KernelUCB
by directly kernelizing the \LinUCB algorithm.
In contrast, \GPUCB is motivated from experimental design.
Our derivation is the combination of the kernel trick \parencite{shawe2004kernel}
and the kernelized version of the Mahalanobis distance \parencite{Haasdonk2010}.

\section{Kernelized \UCB}
\label{Kernelisation}

Kernel methods assume that there exists a mapping $\phi:\R^D\ra\cH$ that maps
the data to a (possibly infinite dimensional) Hilbert space in which a linear
relationship can be observed. We call $\R^D$ the \emph{primal space} and $\cH$
the associated \emph{reproducing kernel Hilbert space} (RKHS).
We use matrix notation to denote the inner product of two elements $h,h'\in\cH$, i.e., $h\transpose h'\eqdef\left<h,h'\right>_{\cH}$
and $\|h\| = \sqrt{\left<h,h\right>_{\cH}}$ to denote the RKHS norm.
From the mapping $\phi$ we have the \emph{kernel function}, defined by:
\[
k(x,x')\eqdef\phi(x)\transpose \phi(x'),\ \forall x,x'\in\R^D,
\]
and the \emph{kernel matrix} of a data set $\{x_1,\dots,x_t\}\subset\R^D$ given
by $\bK_t\eqdef\{k(x_i,x_j)\}_{i,j\leq t}$.
For our nonlinear contextual bandit model we assume the existence of a $\phi$
for which there exists a $\theta^\star\in\cH$ such that:
\[
\E(r_{a,t}\mid x_{a,t})=\phi(x_{a,t})\transpose  \theta^\star.
\]
We also let $y_t \eqdef \{r_{a_1,1},\dots,r_{a_t,t}\}\transpose$ and $X_t \eqdef  \{x_{a_1,1},\dots,x_{a_t,t}\}\transpose$.
Taking $a_t^\star\eqdef\arg\max_{a\in\cA}\{\phi(x_{a,t})\transpose  \theta^\star\}$ we can define the regret as usual.  Note that when $\phi\equiv \mathrm{Id}$, we recover the linear bandit case.

To obtain the upper confidence bounds we derive prediction and width estimators for the expected rewards. \LinUCB uses estimators built from ridge regression in the primal. Since we assume that our model is linear in the RKHS we show how to build estimators from ridge regression in $\cH$. By deriving equivalent dual forms which involve only entries of the kernel matrix we avoid working directly in the possibly infinite dimensional RKHS.

First, we take the prediction estimator to be of the form $\hmu_{a,t+1}=\phi(x_{a,t+1})\transpose  \theta_t$ where $\theta_t$ is the minimizer of the regularized least squares loss function,
\begin{align}
\cL(\theta)=\reg\|\theta\|^2+\sum_{i=1}^{t-1}\left(r_i-\phi(x_{i})\transpose  \theta\right)^2.\label{eq:leastsquares}
\end{align}
We derive a representation of this estimator involving only kernels between context vectors. We denote $\Phi_t=\left[\phi(x_1)\transpose,\dots,\phi(x_{t-1})\transpose\right]\transpose $. Note that the solution of the minimization problem $\theta_t\eqdef\min_{\theta\in\cH}\cL(\theta)$ satisfies
\[
(\Phi_t\transpose  \Phi_t+\reg I)\theta_t = \Phi_t\transpose  y_t.
\]
Rearranging this equation we obtain
\begin{equation}
\theta_t = \Phi_t\transpose  \alpha_t\label{NonDualPredictor},
\end{equation}
where $\alpha_t = \reg^{-1}(y_t-\Phi_t\theta_t) =
\reg^{-1}(y_t-\Phi_t\Phi_t\transpose
 \alpha_t)$, which implies that $\alpha_t = (\bK_t+\reg I)^{-1}y_t$.
Finally, denoting
$k_{x,t} \eqdef \Phi_t\phi(x)=\left[k(x,x_1),\dots,k(x,x_{t-1})\right]\transpose $ we
get
\begin{align}
\hmu_{a,t}= \bk_{x_{a,t},t}\transpose  (\bK_t+\reg I)^{-1}y_t.\label{KernalisedPrediction}
\end{align}
While the computation of $\theta_t$ using (\ref{NonDualPredictor}) would require evaluating $\phi(x_i)$ for every data point $x_i$, the dualized representation of the prediction (\ref{KernalisedPrediction}) allows the computation of $\hmu_{a,t}(x)$ only from objects in the kernel matrix.

Next, we construct the widths of the confidence intervals around the prediction. As for linear bandits we find appropriate widths in terms of the Mahalanobis distance of $\phi(x_{a,t})$ from the matrix~$\Phi_t$:
\begin{align}
\hs_{a,t} \eqdef  \sqrt{ \phi(x_{a,t})\transpose  (\Phi_t\transpose  \Phi_t+\reg I)^{-1} \phi(x_{a,t})  }. \label{eq:widths}
\end{align}
Once again we motivate this choice of width by noting that it is exactly the variance of the prediction estimator when the noise in the dualized data is standard normal.
In order to compute these widths we derive a dualized representation of (\ref{eq:widths}).
Our derivation is similar to the kernelization of the Mahalanobis distance for centered data by \textcite{Haasdonk2010}:
Since the matrices $(\Phi_t\transpose  \Phi_t+\reg I)$ and $(\Phi_t\Phi_t\transpose +\reg I)$ are regularized, they are strictly positive definite, and therefore
\begin{align*}
 (\Phi_t\transpose  \Phi_t+\reg I)\Phi_t\transpose  &= \Phi_t\transpose
(\Phi_t\Phi_t\transpose +\reg I), \\
 \Phi_t\transpose (\Phi_t\Phi_t\transpose +\reg I)^{-1} &=
(\Phi_t\transpose \Phi_t+\reg I)^{-1}\Phi_t\transpose .
\end{align*}
Now, we can extract the Mahalanobis distance from the last equation
\[
(\Phi_t\transpose  \Phi_t+\reg I)\phi(x) = (\Phi_t\transpose
\bk_{x,t}+\reg\phi(x)),
\]
from which we deduce that
\[
 \phi(x)=\Phi_t\transpose (\Phi_t\Phi_t\transpose +\reg
I)^{-1}\bk_{x,t}+\reg(\Phi_t\transpose \Phi_t+\reg I)^{-1} \phi(x)
\]
and express $\phi(x)\transpose \phi(x)$ as
\[\bk_{x,t}\transpose
(\Phi_t\Phi_t\transpose +\reg I)^{-1}\bk_{x,t}+\reg\phi(x)\transpose
(\Phi_t\transpose \Phi_t+\reg I)^{-1} \phi(x).\]
Rearranging, we get an expression for the width involving only inner products,
\begin{align}
\hs_{a,t} \eqdef  \reg^{-1/2}\sqrt{ k(x_{a,t},x_{a,t}) - \bk_{x_{a,t},t}\transpose (\bK_t+\reg I)^{-1}\bk_{x_{a,t},t}  }.\label{KernalisedWidth}
\end{align}
As for \LinUCB, \KernelUCB chooses the action $a_t$ at time $t$ which satisfies
\begin{align*}
a_t &\eqdef \argmax_{a \in A} \left(\bk_{x_{a,t},t}\transpose  (\bK_t+\reg I_t)^{-1}y_t
+ \frac{\eta}{\reg^{1/2}}\sqrt{ k(x_{a,t},x_{a,t}) -
\bk_{x_{a,t},t}\transpose
(\bK_t+\reg I)^{-1}\bk_{x_{a,t},t}  }  \right),
\end{align*}
where $\eta$ is a (possibly time dependent) exploration parameter of the
algorithm.
Considering $a_t$ and~$\hs_{a,t}$ we see that \GPUCB is a special case of
\KernelUCB where the regularization constant is set to the model noise.

The selection of an appropriate kernel function is problem
dependent~\parencite{shawe2004kernel}. The linear kernel corresponds to
$\phi\equiv\mathrm{Id}$ and leads to the dual representation of the \LinUCB
algorithm in the primal. A nonlinear kernel function creates a kernelized \UCB
algorithm for a nonlinear bandit.   Typical examples of nonlinear kernel
functions include: the radial basis function where $k(x_i,x_j) = \exp{(- || x_i
-x_j || ^2/2\sigma^2)}$, for $\sigma >0$ and the polynomial kernel $k(x_i, x_j)
 = \left( x_i\transpose  x_j +1 \right) ^p$.


\section{Analysis of \KernelUCB}

If we directly applied known regret bounds \parencite{auer2002using,chu2011contextual} for linear contextual bandits 
to our setting, we would obtain a bound in terms of the dimension of the RKHS, which 
is possibly infinite. 

We avoid this problem through a careful consideration of the
eigenvalues of the covariance matrix and the choice of the regularisation constant and give a bound in terms of a data
dependent quantity $\td$ which we call the \emph{effective dimension}: Let
$(\lambda_{i,t})_{i\geq 1}$ denote the eigenvalues of $C_t^{\gamma}
= \Phi_t\transpose \Phi_t+\reg I $
in decreasing order and define
\[
\td\,\eqdef\min\{j:j\reg\ln \rounds\geq\Lambda_{\rounds,j}\} \text{ where }
\Lambda_{\rounds,j}\eqdef\sum_{i>j}\lambda_{i,\rounds}-\reg.
\]
We call $\td$ the effective dimension because it
gives a proxy for the number of principal directions over which the
projection of the data in the RKHS is spread.
If the data all fall within a subspace of $\cH$ of
dimension $D'$, then $\Lambda_{\rounds,D'}=0$ and $\td\leq D'$. 

However, more generally,~$\td$ can be thought of as a measure of how quickly the eigenvalues of
$\Phi_t\transpose \Phi_t$ are decreasing. For example if the eigenvalues are
only polynomially decreasing in $i$ (i.e., $\lambda_i\leq C i^{-\alpha}$ for
some $\alpha>1$ and some constant $C>0$) then $\td\leq 1 + (C/(\reg\ln
\rounds))^{1/\alpha}$.

In order to get a better dependence of $\td$, we analyze a related algorithm, \SupKernelUCB, 
that uses the elimination technique\footnote{another option would be an approach similar to \LinearEliminator of Theorem~\ref{thm:eliminator}} of \textcite{auer2002using}. With \SupKernelUCB, however, 
the set of arms can no longer be changing.

\newpage

\begin{theorem}[Regret of \SupKernelUCB by \textcite{valko2013finite}]\label{ThmGeneral1}
Assume that $\|\phi(x_{a,t})\|\leq 1$ and $|r_{a,t}| \in [0,1]$ for all $a\in A$
and $t\geq 1$, and set $\eta=\sqrt{2\ln 2\rounds \nodes/\delta}$.
Then with probability $1-\delta$, \SupKernelUCB satisfies:
\begin{align*}
&R_\rounds\leq  \Bigg[
2+2\left(1+\sqrt{\frac{\reg}{2\ln(2\rounds \nodes(1+\ln \rounds)/\delta)}}\right)\|\theta^\star\|  + \\
 & \quad + 8\sqrt{\left(12+\frac{15}{\gamma}\right)\max\big\{\ln\left(\frac{\rounds}{\td \reg}+1\right),\ln \rounds\big\}^3}
\times \sqrt{\left(2\ln \frac{2\rounds \nodes(1+\ln \rounds)}{\delta}\right)}\Bigg]\sqrt{\td \rounds}
\end{align*}
\end{theorem}

\begin{remark}
When $\Phi\equiv\mathrm{Id}$, $\td\leq D$, the assumption that
$\|\phi(x_{a,t})\|\leq 1$ becomes the assumption that the contexts are
normalised in the primal, and we recover exactly the result of
\cite{chu2011contextual} which matches the lower bound for this setting.
\end{remark}

\begin{remark}
\label{remark:norm}
Theorem~\ref{ThmGeneral1} suggests
that if we know that $\|\theta^\star\| \leq L$, for some $L$, 
we should set $\reg$ to be of the order of $L^{-1}$
so that we obtain an $\tcO(\sqrt{L\td \rounds})$ regret.
If we do not have such knowledge,
just setting $\reg$ to a constant
(e.g.,~found by a cross-validation) will incur
$\tcO(\|\theta^\star\|\sqrt{\td \rounds})$ regret.
\end{remark}

\section{Relationship with \GPUCB}
\label{sec:gpucb}

We now relate our analysis to that of \GPUCB by
\textcite{srinivas2009gaussian}, and in particular to their Theorem~3,
which treats the agnostic case.
In this case, $\theta^\star$ is not assumed to
be sampled from a GP, but instead to
have a bounded RKHS norm $\|\theta^\star\|$.
Under this assumption,
the cumulative regret is bounded
as
\begin{equation}
\cO\Big(\big(I (y_\rounds; \theta^\star) + \|\theta^\star\|^2 \sqrt{I (y_\rounds; \theta^\star)}\big)
\sqrt{\rounds} \Big),
\label{eq:gpucb}
\end{equation}
where  $I (y_\rounds; \theta^\star)$ is the mutual information between $\theta^\star$
and the vector of (noisy) observations $y_\rounds$.
Both $I (y_\rounds; \theta^\star)$ in \eqref{eq:gpucb}
and $\td$ are data-dependent quantities.
We now relate them in order to compare the analyses.
We have that:
\begin{align*}
I (y_\rounds; f)  &= \ln|I + \sigma^{-2}\bK_\rounds| = \sum_i \ln(1 + \sigma^{-2}\lambda_{i,\rounds}) \\
&\geq
\ln\left(1+\sigma^{-2}\lambda_{\td-1,\rounds}\right)
\left(\td -1 +\frac{\sum_{i>\td-1} \lambda_{i,\rounds}}{\lambda_{\td-1,\rounds}}\right)\\
&\geq
(\td - 1)\ln\left(1+\sigma^{-2}\lambda_{\td-1,\rounds}\right)\left[1+\frac{\gamma\ln
\rounds}{\lambda_{\td-1,\rounds}}\right]\\
&\geq (\td - 1)\max_B
\min\Biggl\lbrace\ln(1+B)\gamma\sigma^{-2}\ln(\rounds),\frac{\ln(1+B)}{B} \Biggr\rbrace
\\ &\geq \Omega(\td\ln \ln \rounds)
\end{align*}
In the second equality, we used the fact that the eigenvalues of $\Phi_\rounds\transpose\Phi_\rounds$ are the
same as the eigenvalues of $\Phi_\rounds\Phi_\rounds\transpose$.
In the second inequality we used the definition of $\td$. For the second to last
inequality we considered the two cases when
$\lambda_{\td-1,\rounds}\leq B\sigma^2$ and when $\lambda_{\td-1,\rounds}\geq B\sigma^2$
for some $B$.

This shows that $\td$ is at least as good as $I (y_\rounds;
\theta^\star)$, and comparing our Theorem~\ref{ThmGeneral1} with \eqref{eq:gpucb},
our regret bound only scales as $O(\sqrt{\td})$, while the dependence
of the regret bound~\eqref{eq:gpucb} is linear in $I (y_\rounds;
\theta^\star)$. In particular, this means that for the linear kernel
we attain the lower bound for linear contextual bandits
\parencite{chu2011contextual} while \GPUCB is $\sqrt{D}$ away. This
concerns only the agnostic case of \GPUCB,
i.e.,  Theorem~3 by \textcite{srinivas2009gaussian},
which is the same setting as ours.
When~$\theta^\star$ is sampled from a GP, their result for linear case also
matches the lower bound.

\textcite{srinivas2009gaussian} also
provide an upper bound on $I (y_\rounds; \theta^\star)$,
denoted by $\gamma_\rounds$, for certain kernels.
As a consequence of the link between
$I (y_\rounds; \theta^\star)$, $\gamma_\rounds$ and $\td$,
we may also express our bounds in terms of  $\gamma_\rounds$.
Moreover, in the agnostic case again, our bounds enjoy an improved dependence on this parameter:
for example, for the widely used
RBF kernel, our bound scales with $O(\ln \rounds)^{D/2}$
in place of $O(\ln \rounds)^{D}$.

Finally, when $\|\theta^\star\|$ is unknown and we are unable to
regularize appropriately,
our regret bound only depends on
 $\|\theta^\star\|$ linearly (Remark~\ref{remark:norm}),
while the dependence in~\eqref{eq:gpucb} is
quadratic.

\section{Perspectives of bandits for stochastic processes }
In this chapter, we worked with finite (discrete) action spaces. However, 
Gaussian processes (GPs) define a distribution of (continuous) \emph{functions} 
where the smoothness properties are governed by a covariance function (kernel) $\bK$.
Therefore, a natural extension of the setting considered in this chapter is an 
optimization of a continuous function (on a bounded domain), with either 
a bounded RKHS norm (in a frequentist case) or sampled from a GP. 
A clear candidate, especially in the GP case is \ThompsonSampling.
Since the sample, in this case, is a function, the maximization is not trivial 
in general. One option is to sequentially discretize the domain of the given function, 
for example as done by \textcite{contal2016stochastic} using upper confidence bounds. This approach 
is related to general \emph{black-box function optimization} that we discuss in 
Chapter~\ref{chap:function}. Furthermore, it may be possible 
to extend the discretization techniques to other stochastic processes, 
for example to \emph{Brownian motion}. 

On a practical side,  kernel and GP bandits are based on \emph{kernel ridge regression} (KRR) \citep{scholkopf2001learning,shawe2004kernel}  which comes
with computational challenges. The kernel matrix grows and so does the per-step 
computation time, which is undesirable in any sequential setting.
This problem is not specific to bandits and appears
in online kernel regression or online PCA as well. 
A typical solution in the offline or batch case is the \emph{Nystr\"{o}m family} of algorithms which randomly selects a
subset of columns from the kernel matrix that is used to construct a
low-rank approximation. The quality of the approximate solution is strongly affected by the sampling
distribution and the number of columns selected \citep{rudi2015less}.
For KRR, \citet{alaoui2014fast}
introduce a concept of \emph{ridge leverage scores}  (RLSs) of a square
matrix, and shows that Nystr\"{o}m approximations sampled according to RLS
have strong reconstruction guarantees that translate into good guarantees for the approximate KRR solution
\parencite{alaoui2014fast,rudi2015less}.
We can apply the Nystr\"{o}m method with RLSs for the online setting \parencite{calandriello2016analysis} and in 
particular to the bandit case where the kernel matrix being constructed online may have a specific behavior:
Notice that in the cumulative regret optimization, the (well performing) algorithms would choose more and 
more near-optimal points (arms) and therefore the data from which we construct the kernel matrix are more and more \emph{correlated}.
This specific behavior could be in turn used for more adaptive and space-saving approximation of the kernel matrix.

\chapter{Polymatroid bandits}\label{chap:polymatroid} 




In this chapter, we first introduce polymatroids and illustrate them on practical problems. We use the problem 
of the \emph{minimum-cost flow}~\citep{megiddo74optimal} on a network as an illustrative example before we give the formal definition
of polymatroids and learning with them. 
\begin{example}\label{ex:flow} 
Consider a flow network with $\matL$ \emph{source} nodes and one \emph{sink} node. The network is illustrated in Figure~\ref{fig:flow network}. 
\begin{figure}[H]
  \centering
  \includegraphics[width=5.4in]{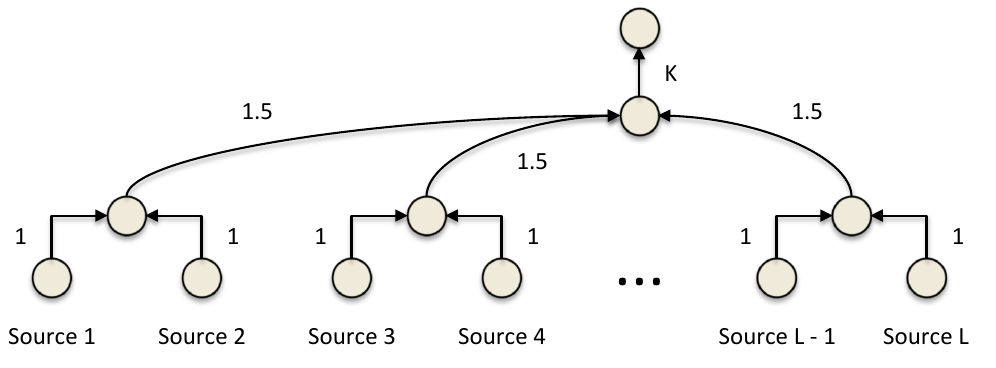}
  \caption{The flow network contains $\matL$ source nodes and the maximum flow is $\matK$. The capacity of the link is shown next to the link.}
  \label{fig:flow network}
\end{figure}
\noindent
The network is defined by three constraints. First, the maximum flow through any source node is $1$. Second, the maximum flow through any two consecutive source nodes, $e$ and $e + 1$ where $e = 2 i - 1$ for $i \in \set{1, \dots, \matL / 2}$, is $\frac{3}{2}$. Third, the maximum flow is $\matK$. We assume that $\matK$ is an integer multiple of $\frac{3}{2}$. The cost of the flow from source node $e$ is a Bernoulli random variable with mean:
\begin{align}
  \bar{\bw}(e) = \left\{
  \begin{array}{ll}
  0.5 - \Delta / 2 & e \leq \frac{4}{3} \matK \\
  0.5 + \Delta / 2 & \text{otherwise}.
  \end{array}
  \right.
  \label{eq:flow weight}
\end{align}
Our problem is parametrized by $\matK$, $\matL$, and $\Delta$. The optimal solution to the problem is to pass the maximum flow through the first $\frac{4}{3} \matK$ source nodes.

Our problem can be formulated as minimizing a \emph{modular} function on a polymatroid. The ground set $E$ are $\matL$ source nodes. The \emph{submodular} function $f$ captures the structure of the network and is defined as
\begin{align}
  f(X) = \min\set{\sum_{i = 1}^{\matL / 2} \min\set{\I\set{(2 i - 1) \in X} + \I\set{2 i \in X}, \frac{3}{2}}, \matK}.
  \label{eq:flow submodular}
\end{align}
Note that $f(X)$ can be computed in $\cO(\matL)$ time, by summing up $\matL$ indicators. The \emph{weight} of item $e$ is drawn i.i.d.\@ from a Bernoulli distribution with mean $\bar{\bw}(e)$ in \eqref{eq:flow weight}, independently of the other items.
\end{example}
With this example in mind we formalize the notion of a polymatroid.
A \emph{polymatroid} \citep{edmonds70submodular} is a polytope associated with a submodular function. More specifically, a polymatroid is a pair $M = (E, f)$. In this definition, $E = \set{1, \dots, \matL}$ is a \emph{ground set} of $\matL$ items.
In our flow problem,$E$ is the set of $\matL$ \emph{sources} of the flow network.
Furthermore $f: 2^E \to \realset^+$ is a function from the power set of $E$ to non-negative real numbers. 
The function $f$ is \emph{monotonic}, $\forall X \subseteq Y \subseteq E: f(X) \leq f(Y)$; \emph{submodular}, $\forall X, Y \subseteq E: f(X) + f(Y) \geq f (X \cup Y) + f(X \cap Y)$; and $f(\emptyset) = 0$. 
In the flow problem, $f(X)$ is the maximum flow through source nodes $X \subseteq E$.
Since $f$ is monotonic, $f(E)$ is one of its maxima. We refer to $f(E)$ as the \emph{rank} of a polymatroid and denote it by $\matK$. 
For the flow problem,~$\matK$~is the value of the maximum flow.
Without loss of generality, we assume that $f(e) \leq 1$ for all items $e \in E$. Because $f$ is submodular, we indirectly assume that $f(X + e) - f(X) \leq 1$ for all $X \subseteq E$. In the flow problem, this constrain translates to assuming that the 
value of any source in the network is upper bounded by one.
The \emph{independence polyhedron} $P_M$ associated with polymatroid~$M$ is a compact subset of $\realset^\matL$ defined as
\begin{align}
  \textstyle
  P_M = \set{\bx: \bx \in \realset^\matL, \ \bx \geq 0,
  \ \forall X \subseteq E: \sum_{e \in X} \bx(e) \leq f(X)},
  \label{eq:independence polyhedron}
\end{align}
where $\bx(e)$ is the $e$-th entry of vector $\bx$. The vector $\bx$ is \emph{independent} if $\bx \in P_M$. 
In the flow example,  $\bx(e) \leq 1$ denotes how much of the unit flow goes through the source $e$ and 
$P_M$ is the set of all possible flows respecting the constraints of a given network.
The \emph{base polyhedron} $B_M$ is a subset of $P_M$ defined as
\begin{align}
  \textstyle
  B_M = \set{\bx: \bx \in P_M, \ \sum_{e \in E} \bx(e) = \matK}.
  \label{eq:base polyhedron}
\end{align}
The vector $\bx$ is a \emph{basis} if $\bx \in B_M$. In other words, $\bx$ is independent and its entries sum up to $\matK$.
For the flow network $B_M$ is the set of all possible \emph{maximum} flows.

\section{Optimization on polymatroids}
\label{sec:optimization}
\begin{algorithm}[t]
  \caption{$\greedy$: Edmond's algorithm for the maximum-weight basis of a polymatroid.}
  \label{alg:greedy}
  \begin{algorithmic}
    \STATE {\bf Input:} Polymatroid $M = (E, f)$, weights $\bw$
    \STATE Let $e_1, \dots, e_\matL$ be an ordering of items such that:
    \STATE \quad $\bw(e_1) \geq \ldots \geq \bw (e_\matL)$
    \STATE $\bx \gets \text{All-zeros vector of length } \matL$
    \FORALL{$i = 1, \dots, \matL$}
      \STATE $\bx(e_i) \gets f(\set{e_1, \dots, e_i}) - f(\set{e_1, \dots, e_{i - 1}})$
    \ENDFOR
    \STATE {\bf Output:} Maximum-weight basis $\bx$
  \end{algorithmic}
\end{algorithm}
\noindent
A \emph{weighted polymatroid} is a polymatroid associated with a vector of weights $\bw \in (\realset^+)^\matL$. The $e$-th entry of $\bw$, $\bw(e)$, is the weight of item $e$. 
For instance $\bw(e)$ can be the cost of a unit flow going through source $e$ and for a particular flow $\bx$,
 $\langle\bw, \bx\rangle$ is the value of the flow given a weight assignment $\bw$.
A classic problem in polyhedral optimization is to find a \emph{maximum-weight basis} of a polymatroid,
\begin{align}
  \bx^\star =
  \argmax_{\bx \in B_M} \langle\bw, \bx\rangle =
  \argmax_{\bx \in P_M} \langle\bw, \bx\rangle.
  \label{eq:optimal}
\end{align}
This basis can be computed greedily (Algorithm~\ref{alg:greedy}). The greedy algorithm works as follows. First, the items $E$ are sorted in decreasing order of their weights, $\bw(e_1) \geq \ldots \geq \bw (e_\matL)$. We assume that the ties are broken by an arbitrary but fixed rule. Second, $\bx^\star$ is computed as $\bx^\star(e_i) = f(\set{e_1, \dots, e_i}) - f(\set{e_1, \dots, e_{i  - 1}})$ for all $i$. Note that the \emph{minimum-weight basis} of a polymatroid with weights $\bw$ is the maximum-weight basis of the same polymatroid with weights $\max_{e \in E} \bw(e) - \bw$,
\begin{align}
  \argmin_{\bx \in B_M} \langle\bw, \bx\rangle =
  \argmax_{\bx \in B_M} \langle\max_{e \in E} \bw(e) - \bw, \bx\rangle.
  \label{eq:minimum basis}
\end{align}
Therefore, the minimization problem is mathematically equivalent to the maximization problem \eqref{eq:optimal}, and all results in this chapter straightforwardly generalize to the minimization. For instance, the minimum-weight basis of the polymatroid corresponding to a flow network is the maximum flow with the minimum cost \citep{fujishige2005submodular}, which we refer to as the minimum-cost flow (see Example~\ref{ex:flow}).

Many existing problems can be viewed as an optimization on a polymatroid \eqref{eq:optimal}. For instance, polymatroids generalize \emph{matroids} \citep{whitney35abstract}, a notion of independence in combinatorial optimization that is closely related to computational efficiency. In particular, let $M = (E, \cI)$ be a matroid, where $E = \set{1, \dots, \matL}$ is its ground set, $\cI \subseteq 2^E$ are its independent sets, and
\begin{align}
  f(X) = \max_{Y: Y \subseteq X, Y \in \cI} \abs{Y}
\end{align}
is its \emph{rank function}. Let $\bw \in (\realset^+)^\matL$ be a vector of non-negative weights. Then the maximum-weight basis of a matroid,
\begin{align}
  A^\star = \argmax_{A \in \cI} \sum_{e \in A} \bw(e),
  \label{eq:matroid optimal}
\end{align}
can be also defined as $A^\star = \set{e: \bx^\star(e) = 1}$, where $\bx^\star$ is the maximum-weight basis of the corresponding polymatroid. The basis is $\bx^\star \in \set{0, 1}^\matL$ because the rank function is a monotonic submodular function with zero-one increments \citep{fujishige2005submodular}.
Our optimization problem can be written as a \emph{linear program} (\LP, \cite{bertsimas1997introduction}),
\begin{align}
  \max_\bx \sum_{e \in E} \bw(e) \bx(e), \quad
  \text{s.t.:} \quad \sum_{e \in X} \bx(e) \leq f(X) \quad
  \forall X \subseteq E,
  \label{eq:LP}
\end{align}
where $\bx \in (\realset^+)^\matL$ is a vector of $\matL$ optimized variables. This \LP has exponentially many constraints, one for each subset $X \subseteq E$. Therefore, it cannot be solved directly. Nevertheless, $\greedy$ can solve the problem in $O(\matL \log \matL)$ time. Therefore, our problem is a very efficient form of linear programming.

The problem of recommending diverse items can be also cast as an optimization on a polymatroid \citep{ashkan14diversified,ashkan15optimal}. Let $E$ be a set of recommendable items, $f(X)$ be the number of topics covered by items $X$, and $\bw$ be a weight vector such that $\bw(e)$ is the popularity of item $e$. Then $\bx^\star = \greedy(M, \bw)$ is a vector such that $\bx^\star(e) > 0$ if and only if item $e$ is the most popular item in at least one topic covered by item $e$. We illustrate this concept with a simple example.
\begin{example}
Let the ground set $E$ be a set of $3$ movies:
\begin{center}
  \begin{tabular}{clrl} \hline
    $e$ & \textcolor{cyan}{Movie title} & \textcolor{cyan}{Popularity} $\bw(e)$ & \textcolor{cyan}{Movie genres} \\ \hline
    1 & Inception & 0.8 & Action \\
    2 & Grown Ups 2 & 0.5 & Comedy \\
    3 & Kindergarten Cop & 0.6 & Action Comedy \\ \hline
  \end{tabular}
\end{center}
Let $f(X)$ be the number of movie genres covered by movies $X$. Then $f$ is submodular and defined as:
\begin{alignat}{8}
  f(\emptyset) & = 0, & \qquad
  f(\set{2}) & = 1, & \qquad
  f(\set{1, 2}) & = 2, & \qquad
  f(\set{2, 3}) & = 2, \\
  f(\set{1}) & = 1, & \qquad
  f(\set{3}) & = 2, & \qquad
  f(\set{1, 3}) & = 2, & \qquad
  f(\set{1, 2, 3}) & = 2. \nonumber
\end{alignat}
The maximum-weight basis of polymatroid $M = (E, f)$ is $\bx^\star = (1, 0, 1)$, and $\set{e: \bx^\star(e) > 0} = \set{1, 3}$ is the minimal set of movies that cover each movie genre by the most popular movie in that genre.
\end{example}

\section{Combinatorial optimization on polymatroids}
\label{sec:combinatorial optimization}

In this chapter, we restrict our attention to the feasible solutions,
\begin{align}
  \Theta = \set{\bx: \left(\exists \bw \in (\realset^+)^\matL: \bx = \greedy(M, \bw)\right)},
  \label{eq:feasible set}
\end{align}
that can be computed greedily for some weight vector $\bw$ and define our objective as finding
\begin{align}
  \bx^\star = \argmax_{\bx \in \Theta} \langle\bw, \bx\rangle.
  \label{eq:optimal basis}
\end{align}
The set $\Theta$ are the vertices of $B_M$ \parencite{kveton2014matroid}.
%
Our choice is motivated by three reasons. First, we study the problem of learning to act greedily. Therefore, we are only interested in the bases that can be computed greedily. Second, many optimization problems of our interest (e.g., recommendation 
of \emph{diverse items}) are combinatorial in nature and only the bases in $\Theta$ are suitable feasible solutions. For instance, in a graphic matroid, $\Theta$ is a set of spanning trees. In a linear matroid, $\Theta$ is a set of maximal sets of linearly independent vectors. The bases in $B_M \setminus \Theta$ do not have this interpretation. Another example is our recommendations problem in Section~\ref{sec:optimization}. In this problem, for any $\bx = \greedy(M, \bw)$, $\set{e: \bx(e) > 0}$ is a minimal set of items that cover each topic by the most popular item according to $\bw$. The bases in $B_M \setminus \Theta$ cannot be interpreted in this way. Finally, we note that our choice does not have any impact on the notion of optimality. In particular, let $\bx$ be optimal for some $\bw$. Then $\bx^g = \greedy(M, \bw)$ is also optimal and since $\bx^g \in \Theta$, it follows that
\begin{align}
  \max_{\bx \in B_M} \langle\bw, \bx\rangle =
  \max_{\bx \in \Theta} \langle\bw, \bx\rangle.
\end{align}

\section{Learning model}
\label{sec:model}

We formalize our learning problem as a polymatroid semi-bandit. A \emph{polymatroid semi-bandit} is a pair $(M, \cP)$, where $M$ is a polymatroid and $\cP$ is a probability distribution over the weights $\bw \in \realset^\matL$ of items $E$ in $M$. The $e$-th entry of $\bw$, $\bw(e)$, is the weight of item $e$. We assume that the weights~$\bw$ are drawn i.i.d.\@ from $\cP$ and that $\cP$ is unknown. Without loss of generality, we assume that $\cP$ is a distribution over the unit cube $[0, 1]^\matL$. Other than that, we do not assume anything about $\cP$. We denote the expected weights of the items by $\bar{\bw} = \mathbb{E}[\bw]$. By our assumptions on $P$, $\bar{\bw}(e) \geq 0$ for all items $e$.
\noindent
Each item $e$ is associated with an \emph{arm} and each feasible solution $\bx \in \Theta$ is associated with a set of arms $A = \set{e: \bx(e) > 0}$. The arms $A$ are the items with nonzero contributions in $\bx$. After the arms are \emph{pulled}, the learning agent receives a \emph{payoff} of $\langle\bw, \bx\rangle$ and \emph{observes} $\set{(e, \bw(e)): \bx(e) > 0}$, the weights of all items with nonzero contributions in $\bx$. This feedback model is known as \emph{semi-bandit} \citep{audibert2014regret}. The solution to our problem is a maximum-weight basis in expectation,
\begin{align}
  \bx^\star =
  \arg\max_{\bx \in \Theta} \EEs{\bw}{\langle\bw, \bx\rangle} =
  \arg\max_{\bx \in \Theta} \langle\bar{\bw}, \bx\rangle.
  \label{eq:optimal arm}
\end{align}
This problem is equivalent to problem \eqref{eq:optimal basis} and so can be solved greedily, $\bx^\star = \greedy(M, \bar{\bw})$.

We choose our observation model for several reasons. First, the model is a natural generalization of that in matroid bandits \citep{kveton2014matroid}. In matroid bandits, the bases are of the form $\bx \in \set{0, 1}^\matL$ and the learning agents observes the weights of all chosen items $e$, $\bx(e) = 1$. In this case, $\bx(e) = 1$ is equivalent to $\bx(e) > 0$. Second, our observation model is suitable for our motivating examples (Section~\ref{sec:optimization}). Specifically, in the minimum-cost flow problem, we assume that the learning agent observes the costs of all source nodes that contribute to the maximum flow. In the movie recommendation problem, the agent observes individual movies chosen by the user, from a set of recommended movies. Finally, our observation model allows us to derive similar regret bounds to those in matroid bandits \citep{kveton2014matroid}.

Our learning problem is \emph{episodic}. Let $(\bw_t)_{t = 1}^\rounds$ be an i.i.d.\@ sequence of weights drawn from distribution $\cP$. In episode $t$, the learning agent chooses basis $\bx_t$ based on its prior actions $\bx_1, \dots, \bx_{t - 1}$ and observations of $\bw_1, \dots, \bw_{t - 1}$; gains $\langle\bw_t, \bx_t\rangle$; and observes $\set{(e, \bw_t(e)): \bx_t(e) > 0}$, the weights of all items with nonzero contributions in $\bx_t$. The agent interacts with the environment in $\rounds$ episodes. The goal of the agent is to maximize its expected cumulative return, or equivalently to minimize its \emph{expected cumulative regret},
\begin{align}
  R_\rounds = \EEs{\bw_1, \dots, \bw_\rounds}{\sum_{t = 1}^\rounds R(\bx_t, \bw_t)},
  \label{eq:cumulative regret}
\end{align}
where $R(\bx, \bw) = \langle\bw, \bx^\star\rangle - \langle\bw, \bx\rangle$ is the regret associated with basis $\bx$ and weights $\bw$.

\section{The \OPM algorithm}
\label{sec:algorithm}

\begin{algorithm}[t]
  \caption{$\opm$: Optimistic polymatroid maximization.}
  \label{alg:ucb1}
  \begin{algorithmic}
    \STATE {\bf Input:} Polymatroid $M = (E, f)$
    \STATE Observe $\bw_0 \sim \cP$
    \COMMENT{Initialization}
    \STATE $\widehat{\bw}_1(e) \gets \bw_0(e), \forall e \in E$
    \STATE $\rounds_0(e) \gets 1, \forall e \in E$
    \FORALL{$t = 1, \dots, \rounds$}
      \STATE $U_t(e) \gets \widehat{\bw}_{\rounds_{t - 1}(e)}(e) + c_{t - 1, \rounds_{t - 1}(e)}, \forall e \in E$
      \COMMENT{Compute UCBs}
      \STATE $\bx_t \gets \greedy(M, U_t)$
      \COMMENT{Find a maximum-weight basis}
      \STATE Observe $\set{(e, \bw_t(e)): \bx_t(e) > 0}$, where $\bw_t \sim P$
      \COMMENT{Choose the basis}
      \STATE $\rounds_t(e) \gets \rounds_{t - 1}(e), \forall e \in E$
      \COMMENT{Update statistics}
      \STATE $\rounds_t(e) \gets \rounds_t(e) + 1, \forall e: \bx_t(e) > 0$
      \STATE $\displaystyle \widehat{\bw}_{\rounds_t(e)}(e) \gets
      \frac{\rounds_{t - 1}(e) \widehat{\bw}_{\rounds_{t - 1}(e)}(e) + \bw_t(e)}{\rounds_t(e)}, \forall e: \bx_t(e) > 0$
    \ENDFOR
  \end{algorithmic}
\end{algorithm}
\noindent
Our learning algorithm is designed based on the \emph{optimism in the face of uncertainty} principle \citep{auer2002finite}. In particular, it is a greedy method for finding a maximum-weight basis of a polymatroid where the expected weight $\bar{\bw}(e)$ of each item is substituted with its optimistic estimate~$U_t(e)$. We refer to our method as \emph{Optimistic Polymatroid Maximization ($\opm$)}.

The pseudocode of $\opm$ is given in Algorithm~\ref{alg:ucb1}. In each episode $t$, the algorithm works as follows. First, we compute an \emph{upper confidence bound} (UCB) on the expected weight of each item~$e$,
\begin{align}
  U_t(e) = \widehat{\bw}_{\rounds_{t - 1}(e)}(e) + c_{t - 1, \rounds_{t - 1}(e)},
  \label{eq:UCB}
\end{align}
where $\widehat{\bw}_{\rounds_{t - 1}(e)}(e)$ is our estimate of the expected weight $\bar{\bw}(e)$ in episode $t$, $c_{t - 1, \rounds_{t - 1}(e)}$ is the radius of the confidence interval around this estimate, and $\rounds_{t - 1}(e)$ denotes the number of times that item $e$ is selected in the first $t - 1$ episodes, $\bx_i(e) > 0$ for $i < t$. Second, we compute the maximum-weight basis with respect to $U_t$ using $\greedy$. Finally, we select the basis, observe the weights of all items~$e$ where $\bx_t(e) > 0$, and then update our model $\widehat{\bw}$ of the environment. The radius
\begin{align}
  c_{t, s} = \sqrt{\frac{2 \log t}{s}}
  \label{eq:confidence radius}
\end{align}
is designed such that each UCB is a high-probability upper bound on the corresponding weight~$\widehat{\bw}_s(e)$. The UCBs encourage exploration of items that have not been observed sufficiently often. As the number of past episodes increases, we get better estimates of the weights $\bar{\bw}$, all confidence intervals shrink, and $\opm$ starts exploiting most rewarding items. The $\log(t)$ term increases with time and enforces continuous exploration.

For simplicity of exposition, we assume that $\opm$ is initialized by observing each item once. In practice, this initialization step can be implemented efficiently in the first $\matL$ episodes. In particular, in episode $t \leq \matL$, $\opm$ chooses first item $t$ and then all other items, in an arbitrary order. The corresponding regret is bounded by $\matK \matL$ because $\langle\bar{\bw}, \bx\rangle \in [0, \matK]$ for any $\bar{\bw}$ (Section~\ref{sec:model}) and basis $\bx$. 

$\opm$ is a greedy method and therefore is extremely computationally efficient. In particular, suppose that the function $f$ is an oracle that can be queried in $\cO(1)$ time. Then the time complexity of $\opm$ in episode $t$ is $\cO(\matL \log \matL)$, comparable to that of sorting $\matL$ numbers. The design of $\opm$ is not very surprising and it draws on prior work \citep{kveton2014matroid,gai2012combinatorial}.

Our major contribution is that we derive a tight upper bound on the regret of $\opm$. Our analysis is a significant improvement over the one of \textcite{kveton2014matroid}, who analyze the regret of $\opm$ in the context of matroids. Roughly speaking, the analysis of \textcite{kveton2014matroid} leverages the augmentation property of a matroid. Our analysis is based on the submodularity of a polymatroid
and we state the distribution independent (gap-free) regret bound below.

\begin{theorem}[Regret of \OPM by \cite{kveton2016learning}]
\label{thm:gap-free} In any stochastic polymatroid semi-bandit, the regret of $\opm$ is bounded as:
\begin{align*}
  R_\rounds \leq 8 \sqrt{\matK \matL \rounds\log \rounds} + \frac{4}{3} \pi^2 \matL^2.
\end{align*}
\end{theorem}

\section{Discussion and perspectives of polymatroid bandits}

The bound of Theorem~\ref{thm:gap-free} is at most linear in $\matK$ and $\matL$, and sublinear in $\rounds$. In other words, it scales favorably with all quantities of interest and therefore we expect it to be practical.  Our $\cO(\sqrt{\matK \matL \rounds \log \rounds})$ upper bound matches the following lower bound  up to a factor of $\sqrt{\log \rounds}$, which is a  corollary of Theorem~\ref{thm:lowerboundauer} \parencite{auer2002nonstochastic}.

\begin{corollary}[Lower bound for matroid bandits by \cite{kveton2016learning}]
\label{prop:gap-free lower bound} For any $\matL$ and $\matK$ such that $\matL / \matK$ is an integer, and any $\rounds > 0$, the regret of any algorithm on a partition matroid bandit  is bounded from below as
\begin{align*}
  R_\rounds \geq \frac{1}{20} \min(\sqrt{\matK \matL \rounds}, \matK \rounds).
\end{align*}
\end{corollary}
Notice that the stated lower bound is for matroids. However, it is an open question whether the factor of $\matL$ is polymatroids is inherent.  It
is possible that learning in polymatroids is harder than in matroids (where the factor is $\matL-\matK$) because the order in which the learning algorithm chooses optimal items matters.

In this chapter, we studied one particular problem, the maximization of a modular function on a polymatroid, in one particular learning setting, stochastic semi-bandits. It is an open question whether the ideas in our paper generalize to other polymatroid problems, such as maximizing a modular function on the intersection of two matroids \citep{papadimitriou1998combinatorial}; and other learning variants of our problem, such as learning in the adversarial setting \citep{auer2002nonstochastic} or with the full-bandit feedback. Several recent papers studied the problem of learning how to maximize a submodular function \citep{guillory2011online,yue2011linear,gabillon2013adaptive,wen2013sequential,gabillon2014largescale}. These are only loosely related to this work because they study a different problem, which is learning how to maximize an \emph{unknown submodular function} subject to a cardinality constraint. Our learning problem is maximizing an \emph{unknown modular function} subject to a \emph{known submodular constraint}.

\chapter{Bandits for function optimization}\label{chap:function}
In this chapter, we apply bandit approaches to the problem of optimizing a function $f: \mathcal{X} \rightarrow \mathbb{R}$ given a finite budget of $\rounds$ noisy evaluations. We consider that the cost of any of these \textit{function evaluations} is high. That means we care about assessing the optimization performance in terms of the sample complexity, i.e.,\@ the number of~$\rounds$ function evaluations. This is typically the case when one needs to tune parameters for a complex system seen as a \emph{black-box}, which performance can only be evaluated by a costly simulation. One such example is the \textit{hyper-parameter tuning} where the sensitivity to perturbations
is large and the derivatives of the objective function with respect to these parameters
do not exist or are unknown.

Such setting is another instance of the sequential decision-making setting under \textit{bandit feedback}.
In this setting, the actions are the points that lie in a domain~$\mathcal{X}$.
At each step $t$, an algorithm selects an action $x_t\in\mathcal{X}$ and receives a reward $r_t$, 
which is a noisy function evaluation such that $r_t =  f(x_t) + \varepsilon_t$, 
where $\varepsilon_t$ is a bounded noise with $\EEc{\varepsilon_t}{x_t} = 0$. After $\rounds$ evaluations,
the algorithm outputs its best guess $x(\rounds)$, which can 
be different from $x_\rounds$. The performance measure we want to minimize is the value of the function at the returned point compared to the optimum, also referred to as \textit{simple regret},
\[
R_\rounds \eqdef \sup_{x\in \mathcal{X}}f(x)  - f\left(x\left(\rounds\right)\right).
\]
We assume there exists at least one point $x^\star\in \mathcal{X}$ such that $f(x^\star)=\sup_{x\in \mathcal{X}}f(x)$.
The relationship with bandit settings motivated \UCT{}~\parencite{kocsis2006bandit,coquelin2007bandit}, an empirically successful heuristic 
\citep{coulom2007efficient,gelly2006modifications,silver2016mastering} 
that hierarchically partitions domain~$\mathcal{X}$ and selects the next point $x_t\in\mathcal{X}$ using upper confidence bounds~\parencite{auer2002finite}.
The empirical success of \UCT{} on one side but the absence of performance guarantees for it on the other, incited research 
on similar but theoretically founded algorithms~\parencite{bubeck2011x,kleinberg2008multi,munos2014from,azar2014online,bull2015adaptive,grill2015black-box}.

As the global optimization of the unknown function without absolutely any assumptions would be 
a daunting needle-in-a-haystack problem, most of the algorithms assume at least a very weak assumption 
that the function does \textit{not decrease faster than a known rate} around \textit{one} of its global optima.
In other words, they assume a certain \textit{local smoothness} property of $f$.
This smoothness is often expressed in the form of a semi-metric $\ell$
that quantifies this regularity~\parencite{bubeck2011x}.
Naturally, this regularity also influences the guarantees that these
algorithms are able to furnish. Many of them define 
a \textit{near-optimality dimension} $\effd$ or a \textit{zooming dimension}.
These are $\ell$-dependent quantities used to bound the simple regret $R_\rounds$ or a related notion called \emph{cumulative regret}.
Table~\ref{table:hoos} lists some of the algorithms for the setting with both \emph{known} and \emph{unknown} smoothness of $f$;
and for both \emph{stochastic} and \emph{deterministic} (where $\eps_t=0$ for all $t$) function evaluations.
In the rest of the chapter, we focus on the stochastic case.

\begin{table}[h]
\centering
 \begin{tabular}[r]{|r|cc|} \hline
& \textbf{\textcolor{cyan}{deterministic}} & \textbf{\textcolor{cyan}{stochastic}} \\ \hline
\specialcell{\textbf{known} \\smoothness} & \DOO{} & \Zooming{}, \HOO{}, \HCT{} \\
\hline
\specialcell{\textbf{unknown} \\ smoothness} & \Direct{}, \SOO{} &
\textbf{\StoSOO{}}, \TZ{}, \ATB{}, \SHOO{}  \\
\hline
\end{tabular}
\caption{Hierarchical optimistic optimization algorithms}
\label{table:hoos}
\end{table}

\section{Near-optimality dimension independent of a semi-metric}\label{sec:nearopt}
In our recent work \parencite{grill2015black-box} we gave a notion of such near-optimality dimension $\effd$ that does not directly relate the smoothness property of $f$ to a specific metric $\ell$ but \emph{directly} to the \emph{hierarchical partitioning} $\mathcal{P}=\{\mathcal{P}_{h,i}\}$, a\emph{ tree-based representation} of the space used by the algorithm.
Indeed, an interesting fundamental question is to determine a good characterization of the difficulty of the optimization for an algorithm that uses a given hierarchical partitioning of space $\mathcal{X}$ as its input. 
The kind of hierarchical partitioning $\{\mathcal{P}_{h,i}\}$ we consider is similar to the ones introduced in prior work: for any depth $h\geq 0$ in the tree representation, the set of \emph{cells} $\{\mathcal{P}_{h,i}\}_{1\leq i\leq I_h}$ form a partition of$\mathcal{X}$, where $I_h$ is the number of cells at depth $h$. At depth 0, the root of the tree, there is a single cell $\mathcal{P}_{0,1}=\mathcal{X}$. A cell $\mathcal{P}_{h,i}$ of depth $h$ is split into several children subcells $\{\mathcal{P}_{h+1,j}\}_{j}$ of depth $h+1$. We refer to the standard partitioning (Figure~\ref{fig:partitioning}) as to one where each cell is split into regular same-sized subcells~\parencite{preux2014bandits}.

\begin{figure}[H]
 \begin{center}
  \includegraphics[height=4cm]{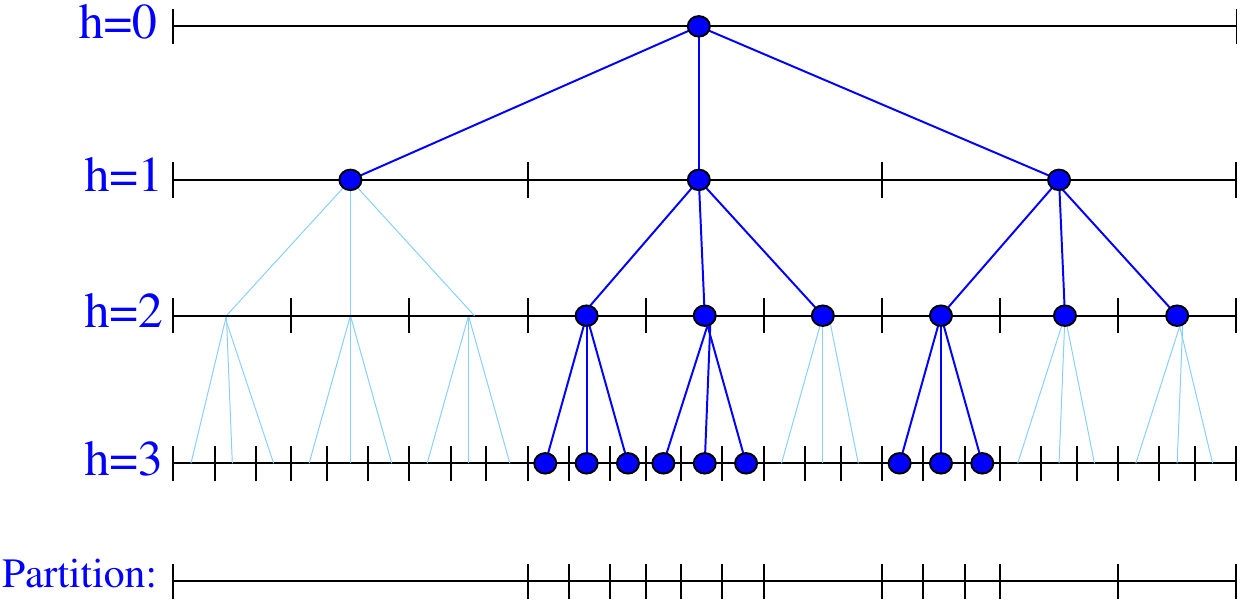}
  \hspace{1cm}
  \includegraphics[height=4cm]{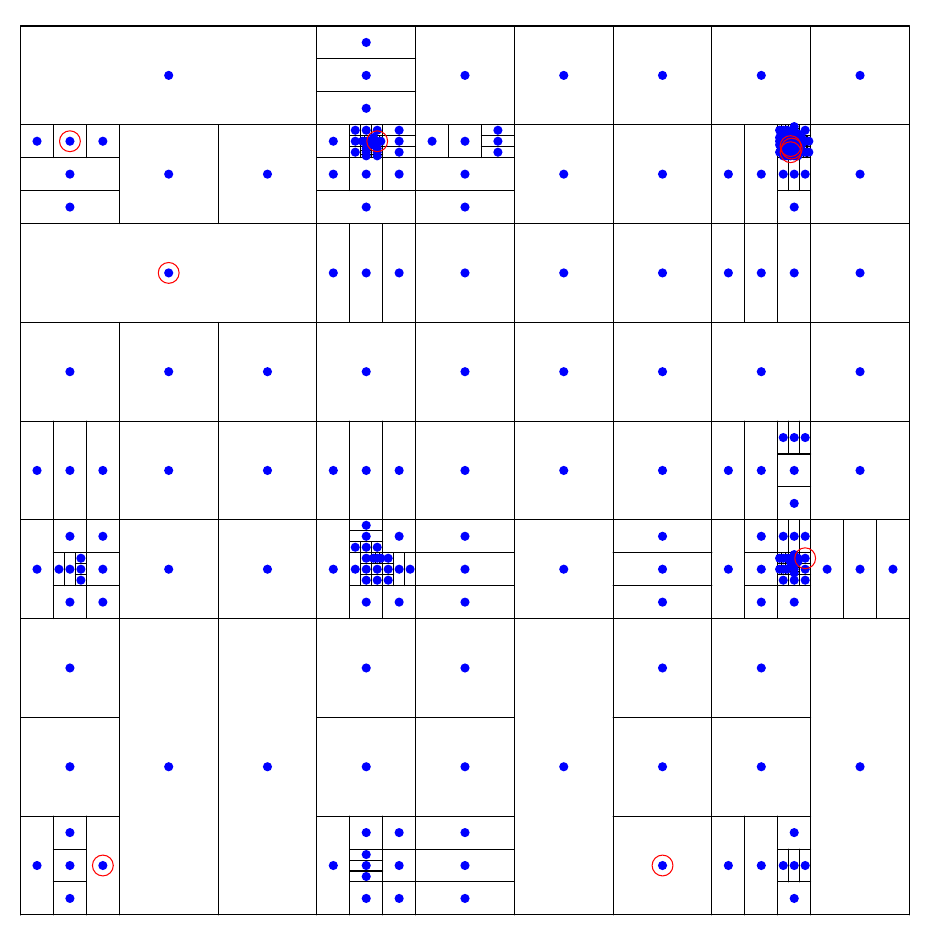}
  \caption{Standard partitioning in one dimension (\textbf{left}) and two dimensions (\textbf{right}).}
 \label{fig:partitioning}
 \end{center}
\end{figure}

\noindent
An important insight (\cite{grill2015black-box}, Section 2)
is that a near-optimality dimension $\effd$ that is independent from the partitioning used by an algorithm 
(as defined in prior work~\cite{bubeck2011x,kleinberg2008multi,azar2014online}) \textit{does not embody the optimization difficulty perfectly}.\/ 
This is easy to see, as for any $f$ we could 
define a  partitioning, perfectly suited for $f$. 
An example is a partitioning, that at the root splits 
$\mathcal{X}$ into $\{x^\star\}$ and $\mathcal{X}\setminus x^\star$,
which makes the optimization trivial, whatever $\effd$ is.
This insight was already observed by~\citet{slivkins2011multi-armed} and~\citet{bull2015adaptive},
whose \textit{zooming dimension} depends both 
on the function and the partitioning. 

Therefore, we defined \parencite{grill2015black-box}  a notion of near-optimality dimension $\effd$ which measures the complexity of the optimization problem \emph{directly in terms of the partitioning} used by an algorithm. First, we make the following local smoothness assumption about the function, expressed in terms of the partitioning  and \emph{not any metric}: For a given partitioning $\mathcal{P}$, we assume that there exist $\nu>0$ and $\rho\in(0,1)$, s.t.\@,
\begin{align*}\label{ass1}
\forall h\geq 0, \forall x\in\mathcal{P}_{h,i^\star_h},\quad  f(x)\ge f\left(x^\star\right)-\nu\rho^h,
\end{align*}
where $\left(h,i^\star_h\right)$ is the (unique) cell of depth $h$ containing $x^\star$. 
Then, we define the near-optimality dimension $\effd(\nu,\rho)$ as
\[\effd(\nu, \rho) \eqdef \inf \left\{d'\in\mathbb{R}^+:\exists C>0, \forall h\geq 0, \mathcal{N}_h(2\nu\rho^h) \le C\rho^{-d'h}\right\},\]
where for all $\varepsilon > 0$, $\mathcal{N}_h(\varepsilon)$ is the number of cells $\mathcal{P}_{h,i}$ of depth $h$ s.t. $\sup_{x\in\mathcal{P}_{h,i}} f(x)\ge f\left(x^\star\right)-\varepsilon$. 
Intuitively, functions with smaller $\effd$ are easier to optimize and we denote $(\nu,\rho)$, for which  $\effd(\nu,\rho)$ is the smallest, 
as $(\nu_\star,\rho_\star)$. Obviously, $\effd(\nu,\rho)$ depends on $\mathcal{P}$ and $f$,  but \emph{does not depend} on any choice of a specific metric.  This definition of~$\effd$\footnote{we use the simplified notation $\effd$ instead of $\effd(\nu,\rho)$ for clarity when no confusion is possible} encompasses the optimization complexity \emph{better} and it is not an artifact of our analysis since many algorithms, such as \HOO{}~\parencite{bubeck2011x},  \Zooming{}~\parencite{slivkins2011multi-armed}, \StoSOO~ \parencite{valko2013stochastic}, or \HCT{}~\parencite{azar2014online}, can be shown to scale with this notion of~$\effd$.
An example of a function with nonzero~$\effd$ is in Figure~\ref{fig:regretrho}.
%

\begin{figure}[H]
\begin{center}
\includegraphics[width=0.45\columnwidth]{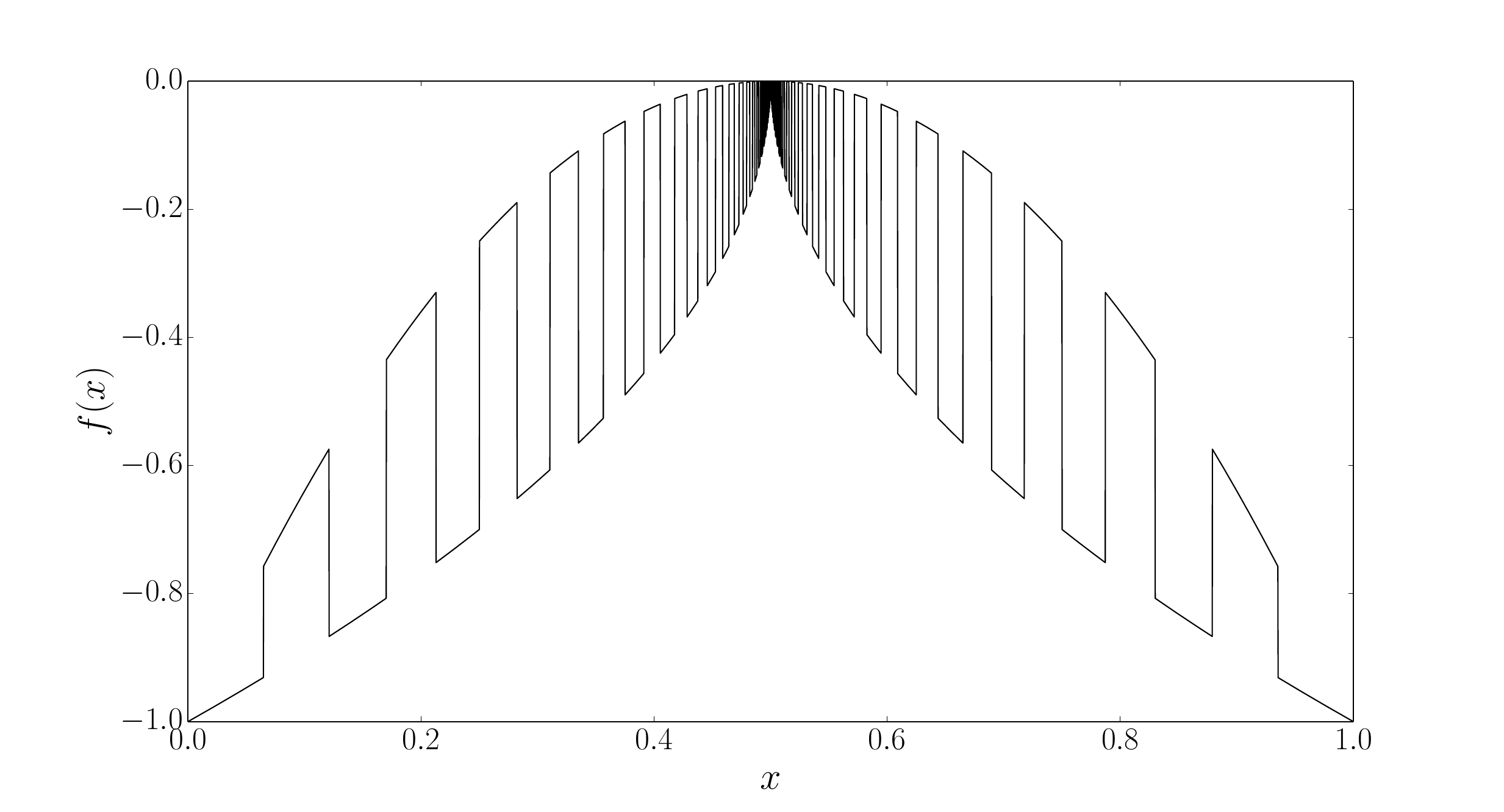}
\includegraphics[width=0.45\columnwidth]{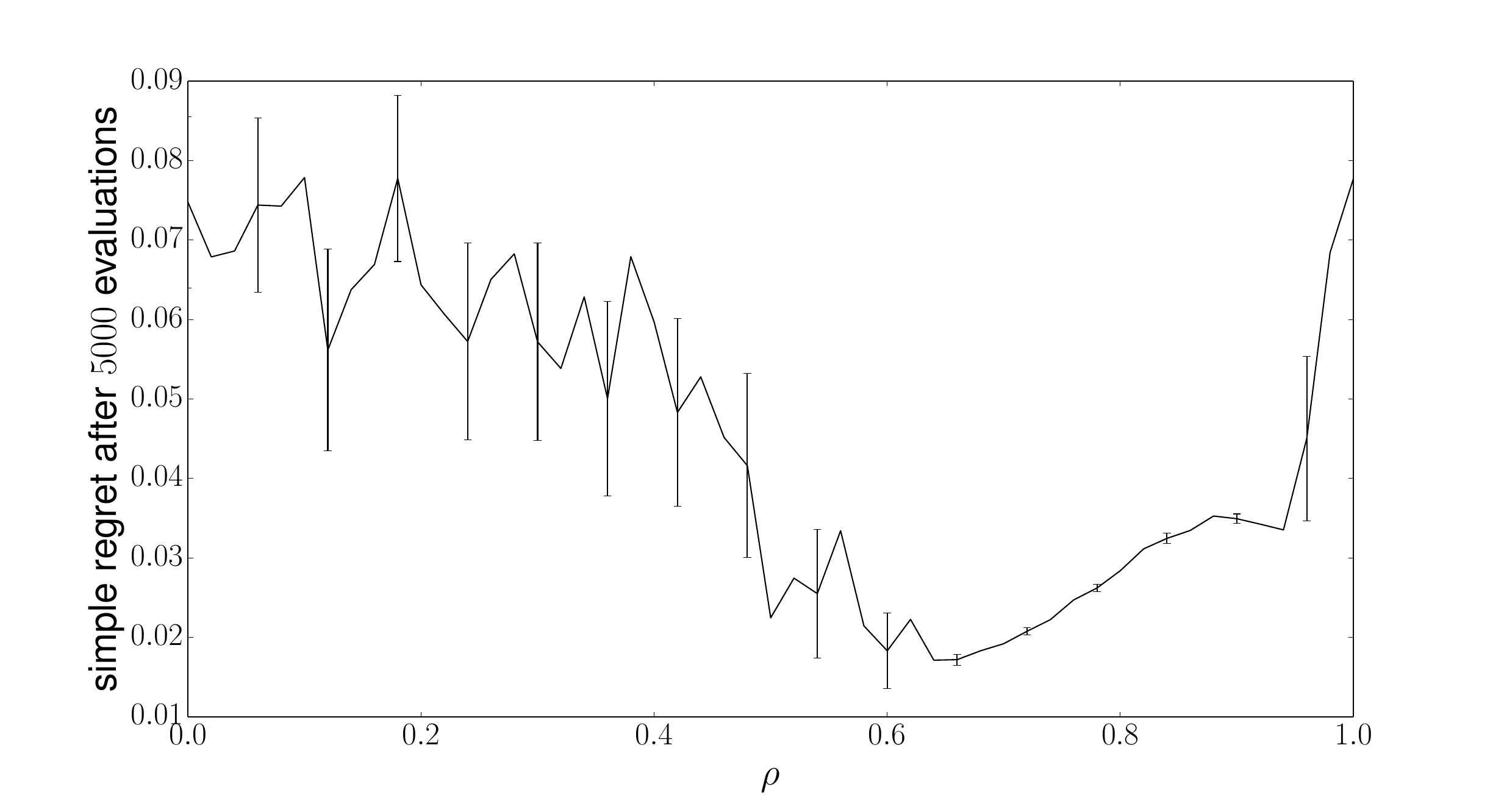}
\end{center}
\caption{ 
Difficult function
$f:x\rightarrow s\left(\log_2|x-0.5|\right)\cdot(\sqrt{|x-0.5|} - {(x-0.5)}^2)-\sqrt{|x-0.5|}$ where, $s(x) = 1$ if the fractional part of $x$, that is,  $x - \lfloor x \rfloor$,  is in $[0,0.5]$ and $s(x) = 0$,  if it is in $(0.5,1)$.
\emph{Left:} Oscillation between two envelopes of different smoothness leading to a nonzero $\effd$ for a standard partitioning. 
\emph{Right:} Regret of \HOO{} after $5000$ evaluations for different values of $\rho$. }
\label{fig:regretrho}
\end{figure}

\section{Hierarchical optimistic optimization: \HOO} 
\label{ss:hoo}
One of the first known algorithms applying bandit approach to function optimization is \HOO{}~\parencite{bubeck2011x}, 
which assumed the knowledge of the function smoothness.
\HOO{} follows an optimistic strategy close to \UCT{}~\parencite{kocsis2006bandit}, but unlike \UCT{}, it uses proper confidence bounds to provide theoretical guarantees. \HOO{} refines a partition of the space based on a hierarchical partitioning, where at each step, a yet unexplored cell (a leaf of the corresponding tree) is selected, and the function is evaluated at a point within this cell.
The selected path (from the root to the leaf) is the one that maximizes the minimum value $U_{h,i}(t)$ among all cells of each depth, where the value~$U_{h,i}(t)$ of any cell $\mathcal{P}_{h,i}$ is defined as
\[U_{h,i}(t) = \widehat{\mu}_{h,i}(t) + \sqrt{\frac{2\ln(t)}{N_{h,i}(t)}}+\nu\rho^h,\]
where $t$ is the number of evaluations done so far, $\widehat{\mu}_{h,i}(t)$ is the empirical average of all evaluations done within  $\mathcal{P}_{h,i}$, and $N_{h,i}(t)$ is the number of them. 
The second term in the definition of $U_{h,i}(t)$ is a Chernoff-Hoeffding type confidence interval, measuring the estimation error induced by the noise. The third term, $\nu \rho^h$ with $\rho\in(0,1)$ is, by assumption, a bound on the difference $f(x^\star)-f(x)$ for any $x\in\mathcal{P}_{h,i^\star_{h}}$, a cell containing $x^\star$. It is this bound, where \HOO{} relies on the knowledge of the smoothness, because the algorithm requires the values of $\nu$ and $\rho$. 
As a consequence of the analysis of \HOO \parencite{bubeck2011x,bubeck2011pure} using 
the assumption from Section~\ref{sec:nearopt} the simple regret of \HOO can be bounded as follows.
\begin{theorem}[Simple regret of \HOO by \cite{bubeck2011pure}]\label{thm:hooregret}
Let~$R_\rounds$ be the simple regret of \HOO{} at step~$\rounds$. 
Let $\effd(\nu_\star,\rho_\star)$ be the near-optimality dimension verifying the assumption from Section~\ref{sec:nearopt} there exists $\kappa$ such that for all $\rounds$, then for any $\effd' \ge \effd(\nu_\star,\rho_\star) $
\[ \mathbb{E}[R_\rounds] \leq \kappa \left(\left(\ln \rounds\right)/\rounds\right)^{1/(\effd' + 2)}.\]
\end{theorem}
\noindent
 \HOO{} was later followed by \HCT{}~\parencite{azar2014online} that needs to assume a slightly stronger condition on the cell
and has a better dependency on the smoothness.
However, as \HOO{}, also  \HCT{} and other algorithms  assume that the smoothness of the 
optimized function is \textit{known}. This is the 
case of known \emph{semi-metric}~\parencite{bubeck2011x,azar2014online} and
\emph{pseudo-metric}~\parencite{kleinberg2008multi}.
This assumption limits the application of these algorithms and opened a very compelling question of whether this knowledge is necessary. 
We provide some answers in the following.

\section{Unknown function smoothness: \StoSOO} 
We now describe \StoSOO, an algorithm for stochastic function evaluations
that does not require the knowledge of the smoothness.
\StoSOO operates in the
traversals
of the tree $\T$, starting from the root down to the
current depth, that is upper bounded by $h_{\max}$, a parameter of the
algorithm.
During each traversal, \StoSOO{} selects
a set of promising nodes, at most one per depth $h$. These nodes are then
either \textit{evaluated} or \textit{expanded}.

Evaluating a node at time $t$
means sampling the
function in the representative point $x_{h,i}$ of the cell $\cX_{h,i}$ and observing the
evaluation $r_t$.
Expanding a node $\node{h}{i}$, means splitting its corresponding
cell into its $K$ subcells corresponding to the children:
\[\{\node{h+1}{i_1},\node{h+1}{i_2},\dots,\node{h+1}{i_K}\}.\]
We denote by $\calL$ the set of leaves in $\T$, i.e., the nodes with no
children.
At any time, only the leaves are eligible for an evaluation or expansion
and we never expand the leaves beyond  depth $h_{\max}$.
If the function $f$ were deterministic, such as in
\SOO~\parencite{munos2011optimistic}, we would expand (simultaneously) any leaf
$\node{h}{i}$
whose value $f(x_{h,i})$ is the largest among all leaves of the same or a lower
depth, because all such nodes may contain $x^\star$.
Unfortunately, we do not receive
$f(x_{h,i})$,
but only a noisy estimate $r_t$.
Therefore, the main algorithmic idea of \StoSOO{} is
to evaluate the leaves several times in order to build a confident
estimate of $f(x_{h,i})$. For this purpose,
let us define $\widehat\mu_{h,i}(t)=\frac{1}{\rounds_{h,i}(t)}\sum_{s=1}^t r_s
 \1\{x_s\in
\X_{h,i}\}$ the empirical average of rewards obtained at state $x_{h,i}$ at time
$t$, where $\rounds_{h,i}(t)$ is the number of
 times that $\node{h}{i}$ has been sampled up to time $t$.

\StoSOO{} builds an accurate estimate of $f(x_{h,i})$ before $\node{h}{i}$ is
expanded.
To achieve this, we define an upper confidence bound (or a $b$-value) for
each node $\node{h}{i}$ as:
\begin{align}\label{eq:b-val.SSOO}
b_{h,i}(t)\eqdef \widehat\mu_{h,i}(t) + \sqrt{\frac{\log(\rounds k/\delta)}{2
\rounds_{h,i}(t)}},
\end{align}
where $\delta$ is the confidence parameter.
In the case of $\rounds_{h,i}(t) = 0$, we let $b_{h,i}(t)= \infty$.
 We refer to $\sqrt{\log(\rounds k/\delta)/2 \rounds_{h,i}(t)}$
 as to the \emph{width} of the estimate.
Now instead of selecting the promising nodes
according to their values $f(x_{h,i})$, we  select them according to their
$b$-values $b_{h,i}$.
The analysis of \StoSOO \parencite{valko2013stochastic} reveals that the simple regret 
is linked to the depth of the tree after $\rounds$ iterations.
This depends on the number of the evaluations per node $k$
before the node is expanded. However, we were only able to 
deduce the value of $k$ for the partitioning with exponentially decreasing diameters 
and for the case of near-optimality dimension zero:

\begin{corollary}[Simple regret of \StoSOO for $\effd=0$ by \cite{valko2013stochastic}]\label{thm:col1}
For the choice $k=\rounds/\log^3(\rounds)$  and $\delta =
1/\sqrt{\rounds}$, we have:
\[\E[R_\rounds]=O\left(\frac{\log^2 \rounds}{\sqrt{\rounds}}\right).\]
\end{corollary}
This result shows that, surprisingly, \StoSOO{} achieves the same rate $\tcO(\rounds^{-1/2})$, up to a logarithmic factor, as the \HOO algorithm run with the best possible metric, although \StoSOO{} does not require the knowledge of~it.
While it is not clear how to \emph{adaptively} set $k$ for $\effd>0$, in the next section we show 
how to approach the case $\effd>0$ differently, by running several \HOO algorithms \emph{in parallel}.

\section{Parallel optimistic optimization: \POO} 
The \SHOO{} algorithm  \parencite{grill2015black-box} is an algorithm aiming at optimizing functions with unknown smoothness with $\effd\ge0$.
\SHOO{}  uses, as a subroutine, an optimization algorithm that \textit{requires the knowledge} of the function smoothness. We use \HOO{}~\parencite{bubeck2011x} 
as the base algorithm, but other algorithms, such as \HCT{}~\parencite{azar2014online}, 
could be used as well. \SHOO{} runs several \HOO{} instances in parallel, hence the 
name \textit{\textbf{p}arallel \textbf{o}ptimistic \textbf{o}ptimization}. The number of base  \HOO{}
instances and other parameters are adapted to the budget of evaluations and are automatically decided on the fly. 

Each instance of \HOO{} requires two real numbers $\nu$ and $\rho$. Running \HOO{} parametrized with ($\rho, \nu)$ that are far from the optimal one $(\nu_\star, \rho_\star)$\footnote{the parameters $(\nu,\rho)$ satisfying the assumption from Section~\ref{sec:nearopt} for which $\effd(\nu,\rho)$ is the smallest} would cause \HOO{} to underperform. Surprisingly, our analysis of this \textit{suboptimality gap} reveals that it does not decrease too fast as we stray away from $(\nu_\star, \rho_\star)$. This motivates the following observation. If we \textit{simultaneously} run a slew of \HOO{}s  with different $(\nu,\rho)$s, one of them is going to perform decently well. 

In fact, we show that to achieve a good performance, we only require $\ln \rounds$ \HOO{} instances, where~$\rounds$ is the current number of function evaluations. Notice, that we do not require to know the total number of rounds in advance which hints that we can hope for a \textit{naturally anytime} algorithm.

The strategy of \SHOO{} is quite simple: It consists of running $N$ instances of \HOO{} in parallel, that are all launched with different $(\nu, \rho)$s.  At the end of the whole process,
\SHOO{}  selects the instance~$s^\star$ which performed the best and returns one of the points selected by this instance, chosen uniformly at random. Note that just using a doubling trick in \HOO{} with increasing values of $\rho$ and $\nu$ is not enough to guarantee a good performance. Indeed, it is important to keep track of all \HOO{} instances. Otherwise, the regret rate would suffer way too much from using the value of $\rho$ that is too far from the optimal one. 


Since \SHOO{} is anytime,  the number of instances $N(\rounds)$ is time-dependent and does not need to be known in advance.
In fact, $N(\rounds)$ is increased alongside the execution of the algorithm. 
More precisely, we want to ensure that 
\[N(\rounds) \ge \tfrac{1}{2}D_{\max}\ln\left( \rounds / \ln \rounds\right), \quad
\text{\ where\ } \quad  D_{\max} \eqdef (\ln K)/\ln\left( 1/\rho_{\max}\right)\cdot\]

\noindent
To keep the set of different $(\nu, \rho)$s well distributed,
the number of  \HOO{}s is not increased one by one but instead is doubled when needed.
Moreover, we also require that \HOO{}s run in parallel, perform the same number of function evaluations. Consequently, when we start running new instances, we first ensure to make these instances on par with already existing ones in terms of the number of evaluations. 

Finally, as our analysis reveals, a good choice of parameters $(\rho_i)$ is not a uniform grid on~$[0,1]$. 
Instead, as suggested by our analysis, we require that $1/\ln(1/\rho_i)$ is a uniform grid on~$[0,1/(\ln1/\rho_{\max})]$. As a consequence, we add \HOO{} instances in batches 
such that  $\rho_i = {\rho_{\max}}^{N/i}$.

\SHOO{} does not require the knowledge of a $(\nu, \rho)$ verifying the assumption from Section~\ref{sec:nearopt} and\footnote{note that several possible values of those parameters are possible for the same function} yet we prove that it achieves a performance close\footnote{up to a logarithmic term $\sqrt{\ln \rounds}$ in the simple regret} to the one obtained by \HOO{} using the best parameters $(\nu_\star,\rho_\star)$. This result solves the open question from the previous section, whether the stochastic optimization of~$f$ with unknown parameters $(\nu, \rho)$  when $\effd>0$ for the standard partitioning is possible.

 \begin{theorem}[Simple regret of \POO by \cite{grill2015black-box}]\label{thm:poo}
Let~$R_\rounds$ be the simple regret of \SHOO{} at step~$\rounds$. 
For any~$(\nu,\rho)$ verifying the assumption from Section~\ref{sec:nearopt} such that $\nu \le \nu_{\max}$ and $\rho \le \rho_{\max}$ there exists $\kappa$ such that for all $\rounds$
\[\mathbb{E}[R_\rounds] \leq \kappa\cdot\left(\left(\ln^2\rounds \right)/\rounds\right)^{1/(\effd(\nu,\rho) + 2)}\]
Moreover, $\kappa = \alpha \cdot D_{\max}{\left(\nu_{\max}/\nu_\star\right)}^{D_{\max}}$, where 
$\alpha$ is a constant independent of $\rho_\text{max}$ and $\nu_\text{max}$. 
\end{theorem}
The \SHOO{}'s performance should be compared to the simple regret of \HOO{} run with the best parameters~$\nu_\star$ and~$\rho_\star$  (Theorem~\ref{thm:hooregret}).
Thus \SHOO{}'s performance is only a factor of $\cO ( \left( \ln \rounds\right)^{1/(\effd(\nu_\star,\rho_\star) + 2)})$ away from the optimally fitted \HOO{}.
Furthermore, the regret bound for \SHOO{} is slightly better than the regret bound for \StoSOO{}~(Corollary~\ref{thm:col1}) in the  case when $\effd(\nu,\rho)=0$ for the same partitioning, i.e., $\mathbb{E}[R_\rounds] = \cO\left(\ln \rounds/\sqrt{\rounds}\right).$ This way \POO generalizes the bound of \HOO for any value of~$\effd\geq 0$.

Note that we only give a simple regret bound for \SHOO{} whereas \HOO{} ensures a bound on both the cumulative and simple regret.\footnote{in fact, the bound on the simple regret is a direct consequence of the cumulative regret bound~\parencite{bubeck2011pure}} Notice that since \SHOO{} runs several \HOO{}s  with nonoptimal values of the $(\nu,\rho)$ parameters, this algorithm explores much more than the optimally fitted \HOO{}, which dramatically impacts the cumulative regret. As a consequence, our result applies to the simple regret only. 

\section{Applicability and perspectives of bandit function optimization}
In this section, we comment on practical issues when using the methods from this chapter in practice.
\subsubsection{Scaling with dimension}
The approaches discussed in this chapter strive to optimize the function 
with \emph{minimal assumptions}. In particular, we showed that it is enough to assume
only a local smoothness property around one of the optima and we are able to provide simple regret guarantees.
This generality has a \emph{cost}, in particular for scaling with the ambient dimension $D$.
This scaling is exponential and \emph{in general unavoidable}: intuitively, if we split $D$ dimensional 
hypercube (of the domain of $f$) along each dimension, the optimum can be in general in any of the sub-hyperrectangles 
and we need to search each of them. This means that these methods are practically relevant 
only for a small $D$. 
\subsubsection{Hyperparameter optimization}
However, these methods can prove very useful for very difficult functions,
for which we know very little (\emph{black-box} setting). One good example 
is hyperparameter optimization, where the \emph{number of parameters is small} 
and the \emph{functions are complex.} As an example, \StoSOO was already 
used in a \textcite{kaggle2013} competition \emph{Cause-effect pairs}\footnote{\url{https://www.kaggle.com/c/cause-effect-pairs}}
in March 2013. The team using \StoSOO \parencite{samothrakis2013training} arrived 3rd (out of 266 teams) and received a prize.
Another application is in parameter optimization of
simulators which are very costly to run, when the (provable) sample complexity is important.

\subsubsection{Extremely difficult functions}
Since \HOO, \SOO, \StoSOO, or \POO require almost no assumptions on $f$, they can 
 compete with methods that also avoid various smoothness assumptions or existence of derivatives. 
 The most common choices for extremely difficult functions are various
\emph{genetic and evolutionary algorithms}. To test the bandit approach to function optimization 
with them, we participated at their annual \emph{CEC'2014 competition} on single-objective real-parameter numerical optimization test suite
\parencite{preux2014bandits}, which showed that on some very difficult functions, 
\SOO can be competitive while providing performance guarantees.

\subsubsection{Extension to other settings }
The methods used in this chapter that optimize functions with unknown smoothness
could be also used in other settings where we can expect smooth rewards to be present, 
but where we are unable to quantify this smoothness. One instance is Monte-Carlo planning 
in MDPs \parencite{szorenyi2014optimistic, grill2016blazing}, where the discount factor naturally induces 
smoothness among the rewards in distant rounds.

\chapter{Infinitely many armed bandits}\label{chap:infi}\index{infinitely many armed bandits}

In this chapter, we consider an extension of multi-arm setting to infinitely many actions, where \emph{no topology} 
(or metric) between the arms is known, the \textit{infinitely many armed bandits}~\parencite{berry1997bandit,wang2008algorithms,bonald2013two-target}. Inevitably, the sheer amount of possible actions makes it impossible to try each of them even once. Such a setting is practically relevant for cases where one faces a finite, but an extremely large number of actions. This setting was first formalized by~\citet{berry1997bandit} as follows. At each time $t$, the learner can either sample an arm (a distribution) that has been already observed in the past, or sample a new arm, whose mean $\mu$ is sampled from the \textit{mean reservoir distribution} $\cL$. 

An example where efficient strategies for minimizing the simple regret of an infinitely many armed bandits are relevant is the search of a good \textit{biomarker} in biology, a single \textit{feature} that performs best on average~\parencite{hauskrecht2006fundamentals}. There can be too many possibilities that we cannot afford to even try each of them in a reasonable time. Our setting is then relevant for this special case of \textit{single feature selection}. 
In this chapter, we provide the results for the simple regret of an infinitely many armed bandits, a problem that was not considered before.
This setting has recently found an application in \emph{hyperparameter optimization} \parencite{li2016efficient}.

The additional challenges of the infinitely many armed bandits with respect to the multi-armed bandits come from two sources. 
First, we need to find a good arm among the sampled ones. Second, we need to sample 
(at least once) enough arms in order to have (at least once) a reasonably good one. These two difficulties ask for a tradeoff which we call the \textit{arm selection tradeoff}. It is different from the known \textit{exploration/exploitation tradeoff} and more linked to model selection principles: On one hand, we want to sample only from a small subsample of arms so that we can decide, with enough accuracy, which one is the best one among them. On the other hand, we want to sample as many arms as possible in order to have a higher chance to sample a good arm at least once. This tradeoff makes the problem of infinitely many armed bandits significantly different from the classic bandit problem.

\citet{berry1997bandit} provide asymptotic, minimax-optimal (up to a $\log \rounds$ factor) bounds for the \textit{average cumulative regret}, defined as the difference between  $\rounds$ times the highest possible value $\bar \mu^\star$ of the mean reservoir distribution and the mean of the sum of all samples that the learner collects. A follow-up on this result was the work of~\citet{wang2008algorithms}, providing algorithms with finite-time regret bounds and the work of~\citet{bonald2013two-target}, giving an algorithm that is optimal with exact constants in a strictly more specific setting. In all of this prior work, the authors show that it is the \textit{shape} of the arm reservoir distribution what characterizes the \textit{minimax-optimal rate} of the average cumulative regret. Specifically, \citet{berry1997bandit} and~\citet{wang2008algorithms} assume that the mean reservoir distribution is such that, for a small $\varepsilon>0$, locally around the best arm $\bar \mu^\star$, we have that
\begin{align}\label{eq:refreg}
\mathbb P_{\mu \sim \cL}\left(\bar \mu^\star - \mu \geq \varepsilon\right)\approx \varepsilon^\infibeta,
\end{align}
that is, they assume that the mean reservoir distribution is $\infibeta$-regularly varying in $\bar \mu^\star$. 
When this assumption is satisfied with a known $\infibeta$, their algorithms achieve an expected cumulative regret of order
\begin{equation}\label{eq:cumreg}
\EE{R_\rounds}\!=\!\cO\left(\max\left(\rounds^{\frac{\infibeta}{\infibeta+1}} \polylog \rounds, \sqrt{\rounds} \polylog \rounds\right) \right).
\end{equation}
The limiting factor in the general setting is a $1/\sqrt{\rounds}$ rate for estimating the mean of any of the arms with $\rounds$ samples. This gives the rate~\eqref{eq:cumreg} of~$\sqrt{\rounds}$. It can be refined if the distributions of the arms, that are sampled from the mean reservoir distribution, 
are Bernoulli of mean~$\mu$ and $\bar \mu^\star = 1$  or in the same spirit, if the distributions of the arms are defined on $[0,1]$ and $\bar \mu^\star = 1$~as
\begin{equation}\label{eq:cumreg2}
\EE{R_\rounds}  = \cO\left(\rounds^{\frac{\infibeta}{\infibeta+1}} \polylog \rounds\right).
\end{equation}
\citet{bonald2013two-target} refine the result~\eqref{eq:cumreg2} even more  by removing the $\polylog \rounds$ factor and proving upper and lower bounds that \textit{exactly match}, even in terms of constants, for a specific subcase of a uniform mean reservoir distribution. Notice that the rate~\eqref{eq:cumreg2} is faster than the more general rate~\eqref{eq:cumreg}. This comes from the fact that they  assume that the variances of the arms decay with their quality, making finding a good arm easier. For both rates (\ref{eq:cumreg} and~\ref{eq:cumreg2}), $\infibeta$~is the \textit{key parameter} for solving the arm selection tradeoff: with smaller $\infibeta$ it is more likely that the mean reservoir distribution outputs a high value, and therefore, we need fewer arms for the optimal arm selection tradeoff.

Previous algorithms for this setting were designed for minimizing the cumulative regret of the learner which optimizes the cumulative sum of the rewards. In this chapter, we consider the problem of minimizing the \textit{simple regret}. 

\section{Learning setting}

Let $\tilde {\mathcal L}$ be a distribution of distributions. We call $\tilde{\mathcal L}$ the \emph{arm reservoir distribution}, i.e.,~the distribution of arms. Let $\mathcal L$ be the distribution of the means of the distributions output by $\tilde {\mathcal L}$, i.e.,~the \textit{mean reservoir distribution}.
Let $\mathbb A_t$ denote the changing set of $K_t$ arms at time $t$.


At each time $t+1$, the learner can either choose an arm $k_{t+1}$ among the set of the $K_t$ arms $\mathbb A_t = \{\nu_1, \ldots, \nu_{K_t}\}$ that it has already observed (in this case, $K_{t+1} = K_t$ and $\mathbb A_{t+1} = \mathbb A_t$), or choose to get a sample of a new arm that is generated according to $\tilde {\mathcal L}$ (in this case, $K_{t+1} = K_t+1$ and $\mathbb A_{t+1} = \mathbb A_t \cup \{\nu_{K_t +1}\}$ where $\nu_{K_t +1} \sim \tilde {\mathcal L}$). Let $\mu_i$ be the mean of arm $i$, i.e.,~the mean of distribution $\nu_i$ for $i \leq K_t$. We assume that $\mu_i$ always exists.

In this setting, the learner observes  a sample at each time. At the end of the horizon, which happens at a given time $\rounds$, the learner has to output an arm $\widehat k \leq K_\rounds$, and its performance is assessed by the simple regret
\[r_\rounds = \bar \mu^\star - \mu_{\widehat k},\]
where $\bar \mu^\star = \arginf_{m} \left(\mathbb P_{\mu \sim \cL} (\mu \leq m) = 1\right)$ is the right end point of the domain.

\paragraph{Assumption on the samples}
The domain of the arm reservoir distribution $\tilde {\mathcal L}$ are distributions of arm samples. We assume that these distributions $\nu$ are bounded.

\begin{assumption}[Bounded distributions in the domain of $\tilde {\mathcal L}$]\label{ass:sample}
Let $\nu$ be a distribution in the domain of $\tilde {\mathcal L}$. Then $\nu$ is a bounded distribution. Specifically, there exists a universal constant $C>0$ such that the domain of $\nu$ is contained in $[-C,C]$.
\end{assumption}
This implies that the expectations of all distributions generated by $\tilde {\mathcal L}$ exist, are finite, and bounded by $C$. In particular, this implies that
\[\bar \mu^\star = \arginf_{m} \left(\mathbb P_{\mu \sim \mathcal L} (\mu \leq m) = 1\right)< +\infty,\]
which implies that the regret is well defined and that the domain of $\mathcal L$ is bounded by $2C$. Note that all the results that we prove hold also for sub-Gaussian distributions $\nu$ and bounded $\mathcal L$. Furthermore, it would be possible to relax the sub-Gaussianity using different estimators recently developed for heavy-tailed distributions~\citep{catoni2012challenging}.

\paragraph{Assumption on the arm reservoir distribution}


We now assume that the mean reservoir distribution $\mathcal L$ has a certain regularity in its right end point, which is a standard assumption for infinitely many armed bandits. Note that this implies that the distribution of the means of the arms is in the domain of attraction of a Weibull distribution, and that it is related to assuming that the distribution is $\infibeta$ regularly varying in its end point $\bar \mu^\star$.
\begin{assumption}[$\infibeta$ regularity in $\bar \mu^\star$]\label{ass:reg}
Let $\infibeta>0$. There exist $\tilde E,\tilde E'>0$, and $0< \tilde B<1$ such that for any $0\leq \varepsilon \leq \tilde B$,
\[\tilde E' \varepsilon^{\infibeta} \geq \mathbb P_{\mu \sim \mathcal L}\left(\mu > \bar \mu^\star - \varepsilon\right)\geq \tilde E \varepsilon^{\infibeta}.\]
\end{assumption}%
\noindent This assumption is the same as the classic one~\eqref{eq:refreg}. 
Standard bounded distributions satisfy Assumption~\ref{ass:reg} for a specific $\infibeta$, e.g.,~all the $\infibeta$ distributions, in particular the uniform distribution.

%
We first present the information-theoretic lower bounds for the infinitely many armed bandits with simple regret as the objective. 
We then present our algorithm and its analysis proving the upper bounds that match the lower bounds --- in some cases, depending on $\infibeta$, up to a $\polylog \rounds$ factor. This makes our algorithm (almost) \emph{minimax} optimal. Finally, we provide three important extensions.

\section{Lower bounds}

Theorem~\ref{thm:lb} exhibits the \emph{information theoretic complexity} of our problem.
Comparing these results with the rates for the cumulative regret problem~\eqref{eq:cumreg} from the prior work, one can notice
that there are \emph{two regimes} for the cumulative regret results. One regime is 
characterized by a rate of $\sqrt{\rounds}$ for $\infibeta \leq 1$, 
and the other characterized by a $\rounds^{\infibeta/(1+\infibeta)}$ rate for $\infibeta \geq 1$. Both of these regimes are related to the arm selection tradeoff. The first regime corresponds to \textit{easy} problems where the mean reservoir distribution puts a high mass close to $\bar \mu^\star$, which favors sampling a good arm with high mean from the reservoir. In this regime, the $\sqrt{\rounds}$ rate comes from the parametric $1/\sqrt{\rounds}$ rate for estimating the mean of any arm with $\rounds$ samples. The second regime corresponds to \textit{more difficult} problems where the reservoir is unlikely to output a distribution with a mean close to $\bar \mu^\star$ and where one has to sample many arms from the reservoir. In this case, the~$\sqrt{\rounds}$ rate is not reachable anymore because there are too many arms to choose from subsamples of arms containing good arms.
The same dynamics exists also for the \textit{simple regret}, where 
there are again two regimes, one characterized by a 
$\rounds^{-1/2}$ rate for $\infibeta \leq 2$, and the other characterized 
by a~$\rounds^{-1/\infibeta}$ rate for $\infibeta \geq 2$. 
Provided that these bounds are tight (which is the case, up to a $\polylog \rounds$, Section~\ref{ss:uppersiri}), one can see that there is an 
interesting  difference between the cumulative regret problem 
and the simple regret one. Indeed, the change of regime 
is here for $\infibeta = 2$ and not for $\infibeta = 1$, i.e.,~the 
parametric rate of~$\rounds^{-1/2}$ is valid for larger values of 
$\infibeta$ for the simple regret. This comes from the fact that 
for the simple regret objective, there is no exploitation 
phase and \emph{everything is about exploring}. Therefore, an 
optimal strategy can spend more time exploring the set of 
arms and reach the parametric rate also in situations where 
the cumulative regret does not correspond to the parametric 
rate. 

\begin{theorem}[Simple regret lower bounds for infinitely many arms bandits by \cite{carpentier2015simple}]\label{thm:lb}

Let us write $\mathcal S_{\infibeta}$ for the set of distributions of arms distributions $\tilde {\mathcal L}$ that satisfy Assumptions~\ref{ass:sample} and~\ref{ass:reg} for the parameters~$\infibeta, \tilde E, \tilde E', C$. Assume that $\rounds$ is larger than a constant that depends on $\infibeta, \tilde E, \tilde E',\tilde B, C$. Depending on the value of~$\infibeta$, we have the following results, for any algorithm $\mathcal{A}$, where $v$ is a small enough constant.
\begin{itemize}
\item Case $\infibeta<2$: With probability larger than $1/3$,
\begin{align*}
\inf_{\mathcal{A}} \sup_{\tilde {\mathcal L} \in \mathcal S_{\infibeta}} r_\rounds &\geq v \rounds^{-1/2}.
\end{align*}

\item Case $\infibeta \geq 2$: With probability larger than $1/3$,
\begin{align*}
\inf_{\mathcal{A}}\sup_{\tilde {\mathcal L} \in \mathcal S_{\infibeta}} r_\rounds \geq v \rounds^{-1/\infibeta}.
\end{align*}

\end{itemize}

\end{theorem}

\section{\SiRI and its upper bounds}
\label{ss:uppersiri}

In this section, we present our algorithm, the Simple Regret for Infinitely many arms (\SiRI).
\paragraph{The \SiRI algorithm}
Let
$b = \min(\infibeta,2),$
and let
\[\bar \rounds_{\infibeta} = \lceil A(\rounds) \rounds^{b/2} \rceil,\]
where
 \[
  A(\rounds)=\begin{cases}
    A, & \text{if $\infibeta<2$}\\
		A/\log(\rounds)^2, & \text{if $\infibeta=2$}\\
		A/\log(\rounds), & \text{if $\infibeta>2$}    
  \end{cases}
\]
where $A$ is a small constant whose precise value will depend on our analysis.
Let $\log_2$ be the logarithm in base 2. Let us define
\[\bar t_{\infibeta} = \lfloor \log_2(\bar \rounds_{\infibeta}) \rfloor.\]
Let $\rounds_{k,t}$ be the number of pulls of arm $k \leq K_t$, and $X_{k,u}$ for the $u$-th sample of $\nu_k$. 
The empirical mean of the samples of arm $k$ is defined as
\[\widehat \mu_{k,t} = \frac{1}{\rounds_{k,t}} \sum_{u=1}^{\rounds_{k,t}} X_{k,u}.\]
With this notation, we provide \SiRI in Algorithm~\ref{alg:siri}.
\begin{algorithm}[t]
\begin{algorithmic}
\STATE {\bf Parameters:} $\infibeta, C, \delta$
\STATE \textbf{Initial pull of arms from the reservoir:} 
\STATE Choose $\bar \rounds_{\infibeta}$ arms from the reservoir $\tilde {\mathcal L}$ .
\STATE Pull each of $\bar \rounds_{\infibeta}$ arms once.
\STATE $t \gets \bar \rounds_{\infibeta}$

\STATE \textbf{Choice between these arms:}

\WHILE{$t \leq \rounds$}
\STATE For any $k \leq \bar \rounds_{\infibeta}$: 
\begin{align}
B_{k,t} \gets  \widehat \mu_{k,t} &+  2\sqrt{\frac{C}{\rounds_{k,t}}\log\big(2^{2\bar t_{\infibeta}/b}/(\rounds_{k,t}\delta)\big)} + 
\frac{2C}{\rounds_{k,t}}\log\left(2^{2\bar t_{\infibeta}/b}/(\rounds_{k,t}\delta)\right)\label{eq:UCBSiRI}
\end{align}
\STATE Pull $\rounds_{k,t}$ times the arm $k_t$ that maximizes $B_{k,t}$ and receive $\rounds_{k,t}$ samples from it.
\STATE $t\leftarrow t+ \rounds_{k,t}$
\ENDWHILE
\STATE \textbf{Output:} Return the most pulled arm $\widehat k$.
\end{algorithmic}
\caption{\SiRI 
\textit{SImple Regret for Infinitely many armed bandits}
\label{alg:siri}}
\end{algorithm}

\paragraph{Discussion}
\SiRI is a UCB-based algorithm, where the leading confidence term is of order
\[ \sqrt{\frac{\log\left(\rounds/(\delta \rounds_{k,t})\right)}{\rounds_{k,t}}}\cdot \]
Similar to the \MOSS algorithm~\parencite{audibert2009minimax}, we divide the $\log(\cdot)$ term by $\rounds_{k,t}$, in order to avoid additional logarithmic factors in the bound. But a simpler algorithm with a confidence term as in a classic \UCB algorithm for cumulative regret,
\[\sqrt{\frac{\log(\rounds/\delta)}{\rounds_{k,t}}}\text{,}\]
would provide almost optimal regret, up to a $\log \rounds$, i.e.,~with a slightly worse regret than what we get. It is quite interesting that with such a confidence term, \SiRI  is optimal for minimizing the \textit{simple} regret for infinitely many armed bandits, since \MOSS, as well as the classic \UCB algorithm, targets the cumulative regret. The main difference between our strategy and the cumulative strategies~\parencite{berry1997bandit,wang2008algorithms,bonald2013two-target} is in the number of arms sampled from the arm reservoir:  For the simple regret, we need to sample more arms. Although the algorithms are related, their analyses are quite different: Our proof \parencite{carpentier2015simple}  is \textit{event-based} whereas the proof for the cumulative regret targets \textit{directly the expectations}.

It is also interesting to compare \SiRI with existing algorithms targeting the simple regret for finitely many arms, as the ones by~\citet{audibert2010best}. \SiRI can be related to their \UCBE with a specific confidence term  and a specific choice of the number of arms selected. Consequently, the two algorithms are related but the regret bounds obtained for \UCBE are not informative when there are infinitely many arms. Indeed, the theoretical performance of \UCBE is decreasing with the sum of the inverse of the gaps squared, which is infinite when there are infinitely many arms. In order to obtain a useful bound, in this case, we need to consider a more refined analysis which is the one that leads to Theorem~\ref{thm:ub}.

\newpage
\paragraph{Main result}
We now state the main result which characterizes \SiRI's simple regret according to $\infibeta$.

\begin{theorem}[Upper bounds of \SiRI by \cite{carpentier2015simple}]\label{thm:ub}
Let $\delta>0$. Assume all Assumptions~\ref{ass:sample} and~\ref{ass:reg} of the model and that $\rounds$ is larger than a large constant that depends on $\infibeta, \tilde E, \tilde E', \tilde B, C$. Depending on the value of $\infibeta$, we have the following results, where $E$ is a large enough constant.
\begin{itemize}
\item Case $\infibeta<2$: With probability larger than $1-\delta$,
\begin{align*}
\hspace{-1.2cm} r_\rounds &\leq E \rounds^{-1/2}  \log(1/\delta) (\log(\log(1/\delta)))^{96} \sim \rounds^{-1/2}.
\end{align*}

\item Case $\infibeta> 2$: With probability larger than $1-\delta$,
\begin{align*}
\hspace{-1.2cm}  r_\rounds &\leq E (\rounds\log(\rounds))^{-1/\infibeta}  (\log(\log(\log(\rounds)/\delta)))^{96}  \log(\log(\rounds)/\delta) \sim (\rounds\log \rounds)^{-1/\infibeta} \polyloglog \rounds.
\end{align*}

\item Case $\infibeta = 2$: With probability larger than $1-\delta$,
\begin{align*}
\hspace{-1.2cm}  r_\rounds &\leq E \log(\rounds) \rounds^{-1/2}  (\log(\log(\log(\rounds)/\delta)))^{96}
\log(\log(\rounds)/\delta) \sim \rounds^{-1/2} \log \rounds  \polyloglog \rounds.
\end{align*}
\end{itemize}
\end{theorem}

\subsubsection{Upper bounds discussion}
The bound we obtain is minimax optimal for $\infibeta<2$ \textit{without additional $\log \rounds$ factors}. We emphasize it since the previous results on infinitely many armed bandits give results which are optimal up to a $\polylog \rounds$ factor for the cumulative regret,  except the one by~\citet{bonald2013two-target} which considers a very specific and fully parametric setting. 
For $\infibeta\geq 2$, our result is optimal up to a $\polylog \rounds$ factor. We conjecture that the lower bound of Theorem~\ref{thm:lb} for $\infibeta \geq 2$ can be improved to $(\log(\rounds)/\rounds)^{1/\infibeta}$ and that \SiRI is actually optimal up to a $\polyloglog(\rounds)$ factor for $\infibeta > 2$.\\

\section{Extensions of \SiRI}
We provided also three important extensions \parencite{carpentier2015simple}.
The first extension concerns the case where the distributions of the arms are defined on $[0,1]$ and where~$\bar \mu^\star = 1$. In this case, replacing the Hoeffding bound in the confidence term of our algorithm by a Bernstein bound, bounds the simple regret as
\begin{align*}
\hspace{-0.7cm} r_\rounds = \cO\left(\max(\tfrac{1}{\rounds} \polylog \rounds, (\rounds \log \rounds)^{-\frac{1}{\infibeta}} \polyloglog \rounds \right).
\end{align*}
The second extension treats \textit{unknown} $\infibeta$. We prove \parencite{carpentier2015simple}  that it is possible to estimate $\infibeta$ with enough precision, so that its knowledge is not necessary for implementing the algorithm. This can also be applied to the prior work~\parencite{berry1997bandit,wang2008algorithms} where $\infibeta$ is also necessary for implementation and optimal bounds.
Finally, in the third extension, we make the algorithm anytime using known tools \parencite{carpentier2015simple}.
\vspace{1.8cm}
\begin{figure}[H]
\hspace{-2.5cm}  \includegraphics[width=1.35\columnwidth]{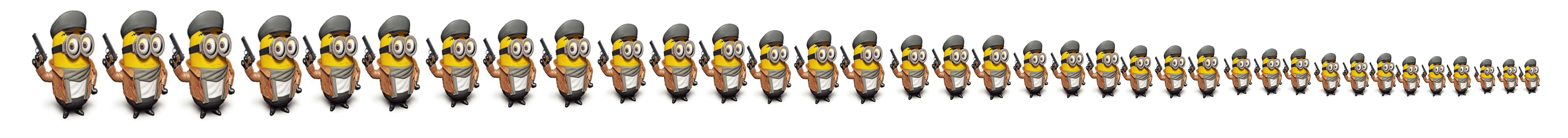}
\end{figure}

\chapter*{References}
\markboth{References}{}
\addcontentsline{toc}{chapter}{\textcolor{ocre}{References}}
\printbibliography[heading=bibempty]




\end{document}